\documentclass[a4paper,fleqn]{cas-dc}

\usepackage[authoryear]{natbib}

\def\tsc#1{\csdef{#1}{\textsc{\lowercase{#1}}\xspace}}
\tsc{WGM}
\tsc{QE}

\newenvironment{PBrev}{\color{black}}{}

\usepackage{booktabs}
\usepackage{csvsimple}
\usepackage[dvipsnames]{xcolor}
\definecolor{lightorange}{RGB}{255,230,204} 
\usepackage{fancyvrb}

\AtBeginEnvironment{equation}{\footnotesize}
\AtBeginEnvironment{gather}{\footnotesize}
\AtBeginEnvironment{align}{\footnotesize}
\AtBeginEnvironment{multline}{\footnotesize}
\csname AtBeginEnvironment\endcsname{equation*}{\footnotesize}
\csname AtBeginEnvironment\endcsname{gather*}{\footnotesize}
\csname AtBeginEnvironment\endcsname{align*}{\footnotesize}
\csname AtBeginEnvironment\endcsname{multline*}{\footnotesize}

\definecolor{elsblue}{RGB}{0,128,172}
\definecolor{elsred}{RGB}{200,30,40}
\colorlet{hscolor}{elsblue}   
\hypersetup{linkcolor=elsred,    
            citecolor=elsblue,   
            urlcolor=elsblue, filecolor=elsblue, menucolor=elsblue}
\ExplSyntaxOn
\RenewDocumentCommand \firstname {} { \textcolor{black}{ \seq_use:Nn \l_stm_au_seq { ~ } } }
\ExplSyntaxOff

\begin{document}
\let\WriteBookmarks\relax
\def\floatpagepagefraction{1}
\def\textpagefraction{.001}

\shorttitle{Report Supervision}    

\shortauthors{P.R.A.S. Bassi et~al.}  

\title [mode = title]{Report Supervision}  

\author[1]{Pedro R. A. S. Bassi}
\author[1]{Wenxuan Li}
\author[2]{Jakob Wasserthal}
\author[1,3]{Jieneng Chen}
\author[1]{Xinze Zhou}
\author[4]{Zheren Zhu}
\author[1]{Chuntung Zhuang}
\author[5]{Sergio Decherchi}
\author[5,6,7]{Andrea Cavalli}
\author[4]{Kang Wang}
\author[4]{Yang Yang}
\author[1]{Alan Yuille}
\author[1,8]{Zongwei Zhou}\cormark[1]
\cortext[cor1]{Corresponding author: Zongwei Zhou (\href{mailto:zzhou82@jh.edu}{zzhou82@jh.edu})}

\address[1]{Department of Computer Science, Johns Hopkins University, Baltimore, MD, USA}
\address[2]{Clinic of Radiology and Nuclear Medicine, University Hospital Basel, Basel, Switzerland}
\address[3]{Department of Computer Science, Stanford University, Stanford, CA, USA}
\address[4]{Department of Radiology and Biomedical Imaging, University of California, San Francisco, San Francisco, CA, USA}
\address[5]{Computational and Chemical Biology, Italian Institute of Technology, Genoa, Italy}
\address[6]{Department of Pharmacy and Biotechnology, University of Bologna, Bologna, Italy}
\address[7]{Centre Europ\'een de Calcul Atomique et Mol\'eculaire (CECAM), \'Ecole Polytechnique F\'ed\'erale de Lausanne, Lausanne, Switzerland}
\address[8]{Department of Oncology, Johns Hopkins Medicine, Baltimore, MD, USA}

\begin{abstract}
Segmentation models can surpass radiologists, classification models, and vision-language models in tumor detection. Importantly, segmentation models outline tumors, allowing radiologists to better verify and trust the AI output. Their main limitation is the scarcity of tumor masks: creating one 3D tumor mask takes up to 30 minutes, so most public CT datasets contain only a few hundred masks, and even the largest private datasets contain only a couple of thousand. Tumor masks are not produced in clinical routine, but radiology reports are. Public datasets contain tens of thousands of CT-Report pairs, and hospitals contain hundreds of thousands. These reports describe tumors in detail, providing large-scale, informative training data. Here, we introduce \textbf{Report Supervision (R-Super)}, a training framework that uses reports to directly supervise and improve tumor segmentation. R-Super introduces new loss functions that teach segmentation models to segment tumors that match report descriptions of tumor count, sizes, and locations. Reports are only used for training. We evaluated R-Super on kidney and pancreatic tumor segmentation, exploring diverse training data sizes, up to 41,418 CT-Report plus 3,488 pancreatic tumor CT-Mask pairs. On external validation, R-Super increased tumor detection F1-Score and segmentation DSC by up to +15\% with respect to mask-only training. It also surpassed alternative methods such as CLIP and multi-task learning. Leveraging numerous readily available reports to supplement scarce masks, R-Super strongly improves AI performance when very few training masks are available (e.g., 50), and when many masks are available (e.g., 3,488), unlocking scale in tumor segmentation. Project: \href{https://github.com/MrGiovanni/R-Super}{https://github.com/MrGiovanni/R-Super}.
\end{abstract}





\maketitle
\makeatletter\gdef\@pdfauthor{Pedro R. A. S. Bassi et al.}\makeatother  



\section{Introduction}

Tumor detection in CT remains difficult: over 50\% of small pancreatic tumors are missed at first read, even by experienced radiologists \citep{hoogenboom2022prevalence}. {\PBrev Kidney tumor detection is also challenging: studies have shown that, on non-contrast CT, radiologists missed 22\% of small kidney tumors, while on contrast-enhanced CT (non-delayed), 11--23\% of tumors were missed \citep{gobara2019t1a,zeman1996helical}.} AI segmentation can support early detection because segmentation models reveal where tumors are, allowing radiologists to better understand and verify the AI output. Additionally, segmentation models often surpass other AI models in tumor detection accuracy, including classification models \citep{hooper2023case} and vision-language models \citep{bassi2025radgpt}. Finally, segmentation models can surpass human performance in tumor detection, since segmentation models can perceive subtle image abnormalities that may be imperceptible to humans \citep{cao2023large,li2026early}.

Despite their large potential, tumor segmentation models face a long-standing limitation: the scarcity of tumor masks. Tumor segmentation masks are outlines of tumors in medical images, needed to train tumor segmentation models. Creating masks is expensive and time-consuming. Creating a single mask in a 3D scan can take a radiologist 30 minutes \citep{zhou2025efficient}. Creating a private dataset with 3,488 pancreatic tumor masks (JHH dataset) required 8 radiologists, 5 years, and millions of dollars \citep{xia2022felix}. Public datasets are even more limited. The majority of the public CT datasets have fewer than 100 tumor masks, few surpass 1,000 tumor masks, and none exceeds 2,000 masks per tumor type \citep{li2025pants,antonelli2021medical,landman2017multiatlas,li2024abdomenatlas,chen2025scaling}. Tumor masks are scarce because radiologists do not create them as part of their job, but they write radiology reports. Hospitals therefore hold hundreds of thousands of reports, and the public Merlin dataset alone contains over 25,000 CT-Report pairs~\citep{blankemeier2024merlin}. Crucially, reports describe exactly what segmentation models need to learn: tumor count, location, and size.

Reports are abundant and detailed, so they may be a key to unlocking scale in tumor segmentation, as text supervision has unlocked scale in general computer vision and natural language processing \citep{radford2021learning}. Therefore, we ask: \textit{can reports supplement masks and scale tumor segmentation datasets, improving tumor segmentation AI?}


Previous work has used radiology reports to improve tumor segmentation. Studies extracted classification labels from reports (e.g., pancreatic tumor present vs. absent). They used image-label pairs, along with image-mask pairs, to train multi-task learning (MTL) models for classification and segmentation \citep{zhang20213d}. Studies have also used reports to create foundational vision-language AI models, by pre-training them with CT-Report pairs and contrastive losses (e.g., CLIP) \citep{blankemeier2024merlin}. These models can be fine-tuned for tumor segmentation using tumor masks. However, MTL- and CLIP-based models share limitations: they do not use reports to directly improve tumor segmentation. Instead, reports optimize auxiliary tasks (classification or report-image contrastive alignment). The benefit of these auxiliary tasks to segmentation is indirect and, in our experiments, small or even negative (Tab.~\ref{tab:all_results}).


Here, we introduce \textbf{R}eport \textbf{Super}vision (R-Super), a training framework that uses radiology reports to \textit{directly optimize} tumor segmentation. R-Super trains segmentation models to segment tumors that match the tumor descriptions in reports. To this end, we introduce two loss functions, the Volume Loss and Ball Loss. These loss functions penalize differences between segmented tumors and reported tumor count, sizes, and locations (organ or organ sub-segments). This information is extracted from radiology reports by a large language model (LLM) before segmentation training. R-Super can train any segmentation architecture and uses reports only at training time, not at inference. Unlike previous methods, R-Super uses reports to directly optimize tumor segmentation, enabling better tumor detection and segmentation. {\PBrev R-Super trains with both CT-Report pairs and CT-Mask pairs, together. For CT-Report pairs, R-Super trains with the Volume Loss and Ball Loss. For CT-Mask pairs, R-Super trains with standard segmentation losses (Dice and cross-entropy).} Therefore, with R-Super we can scale CT-Mask datasets with many CT-Report pairs, improving tumor detection and segmentation performance.

To train R-Super, we built two new private CT-Report datasets, sourced from the University of California San Francisco (UCSF) Hospital and nearby institutions: UCSF-Train (6,718 CT-Report pairs, kidney and pancreatic tumors) and UCSF-Huge (41,418 CT-Report pairs, 28,295 pancreatic tumor images). To test R-Super and to contribute data to the field, we built a new public CT-Report dataset (684 CT-Report pairs, 445 with pancreatic tumors).

We performed three main experiments. \textit{First}, to demonstrate that reports improve tumor segmentation when few or many masks are available, we trained R-Super on UCSF-Train CT-Report pairs plus few (50), medium (344), and many (1.7K) CT-Mask pairs. \textit{Second}, to demonstrate that reports can scale a state-of-the-art public segmentation dataset, improving performance, we added 1,848 public pancreatic tumor CT-Report pairs from Merlin \citep{blankemeier2024merlin} to PanTS \citep{li2025pants} (926 masks), the largest public pancreatic mask dataset. \textit{Third}, to demonstrate that reports can also scale a massive private mask dataset, improving performance, we added 41,418 CT-Report pairs from UCSF-Huge to JHH \citep{xia2022felix} (6,212 CT-Mask pairs, 3,488 with pancreatic tumors), the largest private pancreatic mask dataset, annotated by 8 radiologists over 5 years. In all experiments, R-Super leveraged reports to improve tumor detection (sensitivity, specificity) and segmentation (DSC, NSD), with external validation on hospitals unseen during training. Furthermore, R-Super consistently surpassed alternative training methods (CLIP \citep{radford2021learning}, MTL \citep{zhang20213d}, Models Genesis \citep{zhou2021models}, standard mask-based segmentation \citep{gao2022data}, nnU-Net \citep{isensee2021nnu}, and report-guided pseudo-labels \citep{bosma2023semisupervised}), some of which can also leverage reports. Our main contributions are:

\begin{enumerate}
\item R-Super introduces loss functions that optimize segmented tumors to be consistent with radiology reports. To the best of our knowledge, the R-Super losses are the first to use reports to directly optimize tumor segmentation in CT.
\item R-Super uses reports to improve tumor detection and segmentation whether few or many masks are available for training (up to +15\% F1-Score and DSC over training with masks alone on external validation, Tab.~\ref{tab:results_by_size}).
\item R-Super uses reports to improve performance even on top of state-of-the-art mask datasets, both public (PanTS, Sec.~\ref{sec:pants}) and private (JHH, Sec.~\ref{sec:jhh_ucsf}).
\item {\PBrev By supplementing contrast-enhanced CT-Mask datasets with non-contrast CT-Report pairs, R-Super improves the detection of pancreatic tumors on non-contrast CT scans (Sec.~\ref{sec:jhh_ucsf}).}
\item We release a new public CT-Report dataset, with 684 CT-Report pairs, focusing on pancreatic tumors (Sec.~\ref{sec:jhh_ucsf}).
\end{enumerate}

\section{Related Work}

We compare R-Super against six alternative training methods for tumor segmentation: standard mask-supervised segmentation (MedFormer \citep{gao2022data}, nnU-Net \citep{isensee2021nnu}), self-supervised pre-training (Models Genesis \citep{zhou2021models}), CLIP pre-training with segmentation fine-tuning \citep{radford2021learning}, multi-task learning \citep{chen2019lesion,zhang20213d}, and report-guided pseudo-labels \citep{bosma2023semisupervised}. Most of these alternative methods either ignore reports or use them to supervise an auxiliary training task, whereas R-Super uses reports to directly optimize the tumor segmentation output. We develop these comparisons below. We then explain how this paper extends our previous work.

\subsection{Alternative Training Methods} 

\textbf{CLIP.} A common way to use reports for tumor segmentation today is contrastive language--image pre-training (CLIP). Most medical vision-language models (VLMs) are pre-trained with variants of the CLIP loss \citep{blankemeier2024merlin,hamamci2024ct2rep,sellergren2025medgemma}. These losses pre-train the AI model by giving it a set of images and a set of texts and asking it to find the correct image-text pairs across the two sets. After pre-training, the models can be fine-tuned for tumor segmentation using masks. The features learned by the model during pre-training may help it learn segmentation, but this benefit is indirect and, in our experiments, small or even negative (Tab.~\ref{tab:all_results}). A low CLIP loss during pre-training does not guarantee high-quality segmentation. In contrast, the Report Supervision losses supervise the segmentation output itself, penalizing the model whenever the segmented tumors disagree with the tumor descriptions in the report. A low R-Super loss therefore corresponds, by construction, to segmentations that match the reports.

\textbf{Multi-task Learning.} Multi-task learning (MTL) frameworks \citep{chen2019lesion,zhang20213d} proposed that classification can improve segmentation. In MTL, both segmentation and classification are learned together, by the same model. Features that the model learns for classification can also help segmentation. However, there is no guarantee that one task will help the other. In some cases, this mutual help was marginal or absent \citep{bassi2024improving}. Like CLIP, MTL uses reports to indirectly improve segmentation, so a model can achieve strong classification and poor segmentation. In contrast, R-Super directly uses reports to supervise segmentation.

{\PBrev \textbf{Weakly Supervised Learning and Promptable Segmentation Models.} Medical segmentation methods have reduced segmentation mask requirements using alternative, weaker sources of supervision, such as bounding boxes and scribbles. DeepCut learns segmentation from bounding boxes \citep{deepcut}, while scribble-supervised methods learn from sparse lines drawn within target structures \citep{can_scrible}. Weaker annotations, such as drag-and-drop marks, can also localize tumors in volumetric data at a much lower cost than masks \citep{chou2024acquiring}. More recently, promptable foundation models such as SAM and MedSAM convert points or bounding boxes into segmentation masks \citep{ma2024segment}. These approaches can reduce annotation effort relative to per-voxel segmentation masks, but their spatial annotations or prompts must still be deliberately created and are not routinely available in hospital archives. In contrast, radiology reports are routinely produced and available at scale in hospitals. R-Super uses reports to supervise tumor segmentation without requiring new annotations. Some weak-supervision methods use their annotations (e.g., bounding boxes) only during training, whereas promptable models such as SAM and MedSAM require a point or bounding box at inference to identify the object to segment (e.g., a tumor). This inference-time requirement is particularly limiting for tumor detection: SAM assumes that a radiologist or another AI system has already detected and roughly localized the tumor and is therefore not directly designed to assist in tumor detection.}

\textbf{Self-supervised Learning.} Self-supervised training methods can learn from CT scans alone, using neither masks nor reports. Masked autoencoders (e.g., Models Genesis \citep{zhou2019models,zhou2021models,haghighi2020learning}) receive images with parts masked out and learn to reconstruct the unmasked images, thus learning structural and geometric relationships in medical images. {\PBrev Other self-supervised methods use different pre-training objectives. Rubik’s Cube+ trains a model to recover the order and orientation of shuffled 3D image cubes and identify masked cubes \citep{zhu2020rubik}. \citet{Chaitanya_Contrastive} use global and local contrastive learning. Swin UNETR combines masked-volume inpainting, rotation prediction, and contrastive learning \citep{tang2022self}. Here, we adopt Models Genesis as a self-supervised baseline because it is a well-established 3D medical self-supervised method that has been independently validated on multiple downstream tasks \citep{wald2024revisiting}, making it a strong representative of image-only self-supervised pre-training. After pre-training, self-supervised models} can be fine-tuned for tumor segmentation using masks. However, this approach ignores reports, which provide detailed tumor information. In contrast, R-Super is designed to exploit this detailed information, specifically the tumor count, sizes, and locations.

\textbf{Report-guided Pseudo-labels.} The report-guided pseudo-label method \citep{bosma2023semisupervised} trains a segmentation model on images with tumor masks, and uses this model (teacher) to create pseudo-masks for images without masks. Then, these pseudo-masks are filtered according to reports: only the top-$n$ most confident tumors are kept in each mask, where $n$ is the number of tumors in the report. The filtered pseudo-masks (plus the original masks) are used to train a second segmentation model (student). This filtering scheme reduces false positives in the pseudo-masks, but does not address false negatives, i.e., tumors that the teacher model missed. Thus, masks with false negatives are discarded from the student's training set \citep{bosma2023semisupervised}. However, removing these cases removes precisely the hard-to-detect tumors that would be most informative to learn from. In contrast, R-Super can learn from these hard-to-detect cases, as we designed the Volume Loss to strongly penalize false negatives (Sec.~\ref{sec:volume_loss}). Report-guided pseudo-labels also do not use important report details, such as tumor sizes and locations. In contrast, R-Super uses these details in its loss functions (Sec.~\ref{sec:ball_loss}).

\subsection{Our Previous Work}

Our MICCAI conference paper \citep{bassi2025learning} first introduced the Report Supervision framework, proposing the Volume Loss and the Ball Loss. This paper extends the preliminary work substantially with the following improvements:

\begin{enumerate}
    \item \textit{New public dataset:} 684 CT-Report pairs, including 445 pancreatic tumor cases.
    \item \textit{New experiment with public data:} we show that adding CT-Report pairs, from the \textit{public} Merlin dataset, to the largest \emph{public} pancreatic tumor segmentation dataset (PanTS \citep{li2025pants}) improves tumor detection and segmentation.
    \item \textit{New large-scale experiment:}  we show that adding 41,418 CT-Report pairs to the largest \emph{private} pancreatic tumor mask dataset (6,212 CT-Mask pairs, created by 8 radiologists in 5 years) further improves AI performance.
    \item {\PBrev \textit{Experiments on non-contrast CT:} we demonstrate that adding non-contrast CT-Report pairs to existing contrast-enhanced CT-Mask datasets allows R-Super to detect pancreatic cancer in non-contrast CT.}
    \item \textit{Formalized methodology:} we provide a more complete formulation of the loss functions, with additional equations and a clearer description of the framework.
\end{enumerate}

Follow-up work has since extended R-Super to more tumor types and larger report collections \citep{bassi2025scaling}, and to supervision from follow-up (future) reports \citep{bassi2026rtsuper}.

\section{R-Super}
\label{sec:rsuper}

\begin{figure*}[t]
  \centering
   \includegraphics[width=1\linewidth]{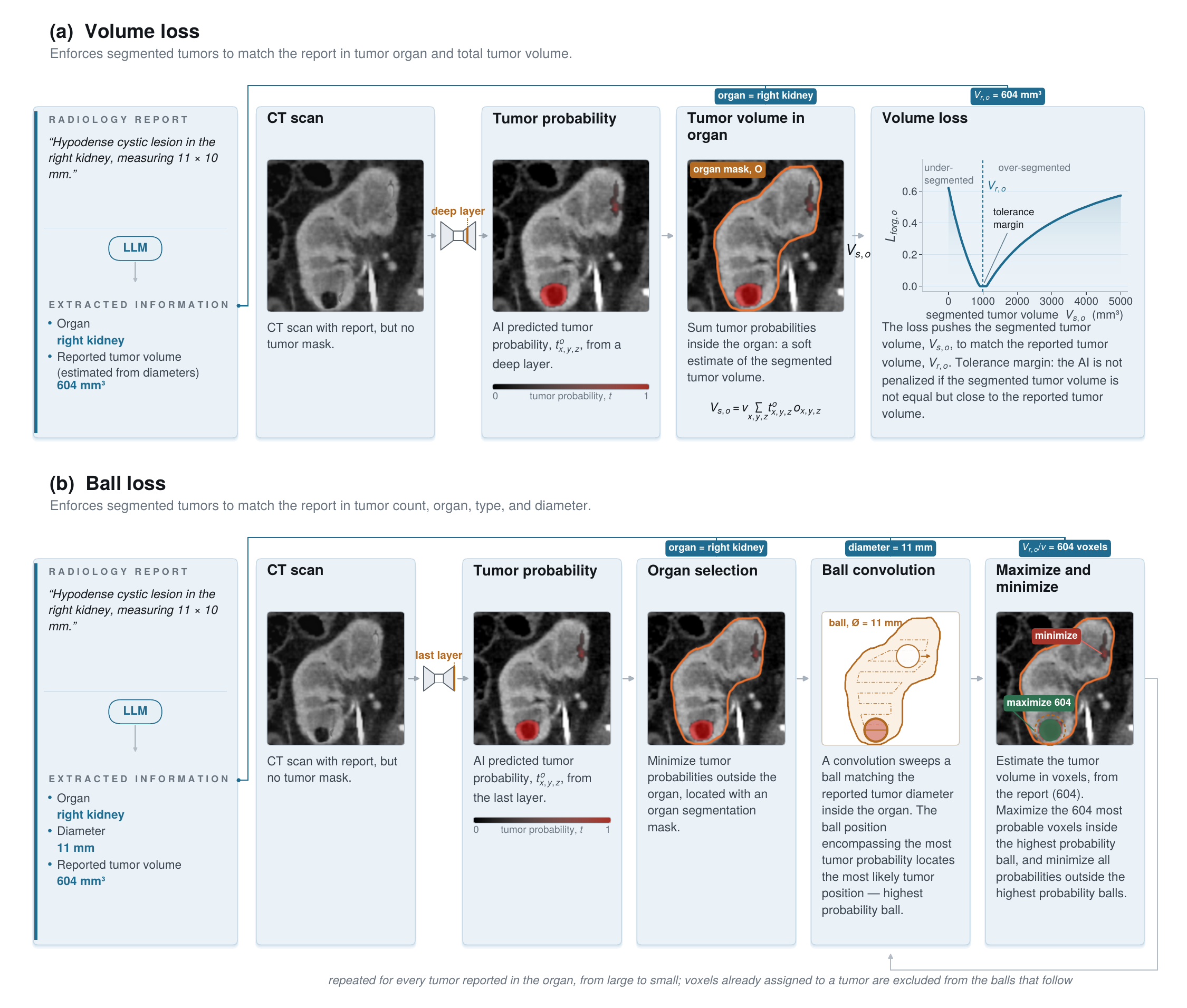}
    \caption{\textbf{R-Super trains tumor segmentation models using information from radiology reports, with Volume Loss and Ball Loss guiding predictions to match reported tumor size, location, and number. (a)} \textit{The Volume Loss (deep supervision) aligns segmented tumors with volumes and locations from reports.} It consists of three steps.
    \textbf{(I)} A large language model (LLM) extracts tumor location (organ/sub-segment) and diameters from the report. Using ellipsoid approximations, we estimate the total tumor volume in each organ $o$ as $V_{r,o}$. 
    \textbf{(II)} From a deep layer of the segmenter, we compute tumor probabilities via a 1$\times$1$\times$1 convolution and sigmoid/softmax. We sum the probabilities inside the organ (using pre-saved organ masks) to estimate the segmented tumor volume, $V_{s,o}$. 
    \textbf{(III)} The Volume Loss penalizes discrepancies between $V_{s,o}$ and $V_{r,o}$. As shown in the right panel (for $V_{r,o} = 1000$ mm$^3$), the loss is zero when $V_{s,o}$ is close to $V_{r,o}$, allowing for some tolerance due to uncertainty in report-based volume estimates. \textbf{(b)} \textit{The Ball Loss enforces segmented tumors to match detailed tumor information from reports---tumor count, locations, diameters, and volumes.} It consists of four steps.
    \textbf{(I)} A large language model (LLM) extracts this tumor information from the report. 
    \textbf{(II)} A Ball Convolution—a simple convolution with a fixed spherical kernel matching the reported tumor diameter—slides over the segmenter output to find the highest-probability tumor region. 
    \textbf{(III)} Within this region, the loss maximizes probabilities in the top-$N_i$ most probable voxels, where $N_i$ is the tumor volume estimated from the report (in voxels, reduced by 20\%). These voxels can form any shape that stays within the ball.
    \textbf{(IV)} The process (II--III) is repeated for all reported tumors, largest to smallest. Already-assigned voxels are ignored in later iterations to prevent overlap. 
    Finally, the loss minimizes tumor probabilities in unassigned voxels (background). It focuses more on central tumor voxels and avoids penalizing uncertain tumor borders.}\label{fig:method}
\end{figure*}

R-Super is a training framework that enforces segmented tumors to match the tumor descriptions in radiology reports. It is summarized in Fig.~\ref{fig:method}. First, an LLM extracts the descriptions of tumor diameters, locations (e.g., pancreatic head), and counts from radiology reports (Sec.~\ref{sec:llm}). Second, this information is used as ground truth for two loss functions, the Volume Loss and the Ball Loss, which train the tumor segmentation model (of any architecture; we used MedFormer \citep{gao2022data}). The \textit{Volume Loss} (Sec.~\ref{sec:volume_loss}) penalizes differences between the tumor volume estimated from the model's segmentation output and the volume estimated from the report. The \textit{Ball Loss} (Sec.~\ref{sec:ball_loss}) penalizes differences in tumor location (e.g., pancreatic head, body, or tail), count, and diameter. 

The Ball Loss is stricter than the Volume Loss, because it requires segmented tumors to match multiple tumor characteristics from the reports. In contrast, the Volume Loss is easier to optimize. Therefore, we apply the Volume Loss as deep supervision (e.g., applying it at the second layer of the MedFormer decoder), and we apply the Ball Loss at the final layer of the segmentation model. This design enables easy optimization while ensuring that the final segmentation output matches the detailed tumor descriptions in radiology reports. We train R-Super with CT-Mask pairs plus numerous CT-Report pairs. For CT-Report pairs, we use the Ball Loss and Volume Loss; for CT-Mask pairs, we use standard segmentation losses, Dice and cross-entropy.

The Volume Loss and Ball Loss are guided not only by reports but also by organ and organ sub-segment masks. These masks localize in the image the diseased organs mentioned in the report, allowing our losses to teach the tumor segmentation model to segment tumors only within these diseased organs. We generate the organ/sub-segment masks before training the tumor segmentation model. To this end, we use an nnU-Net \citep{isensee2021nnu} trained on AbdomenAtlas 3.0 (9,262 CT scans) \citep{bassi2025radgpt} to segment organs such as the pancreas and its sub-segments (pancreas head/body/tail). {\PBrev The kidney also has sub-segments, including the renal cortex, medulla, and pelvis. However, our nnU-Net does not segment these sub-segments separately. Therefore, for the kidneys, we use organ masks (which separate the right and left kidneys).} Organ masks can reliably guide our tumor segmentation losses because organ segmentation is an easier task, and public AI models segment organs with much higher accuracy than tumors \citep{antonelli2021medical,liu2024universal,qu2023annotating}. For example, nnU-Net trained on AbdomenAtlas achieves 82.9\% DSC for pancreas segmentation on the Touchstone benchmark \citep{bassi2024touchstone}. Still, to compensate for inaccuracies in organ/sub-segment masks and to compensate for tumors that grow beyond organ boundaries, we dilate the organ/sub-segment masks (by 2 cm). {\PBrev Even in ablation studies with organ masks generated by an nnU-Net trained on only 50 CT scans, R-Super maintained high tumor detection and segmentation performance (Tab.~\ref{tab:ablations}). Although organ masks improve the R-Super performance by making the Volume Loss and Ball Loss more precise, these losses remain effective even without organ masks (Tab.~\ref{tab:ablations}).}

\subsection{LLM for Extracting Tumor Information}
\label{sec:llm}

We use a large language model (LLM) to extract tumor information from reports, because LLMs can understand both context and semantics. This understanding is critical because reports vary widely across hospitals, locations, time periods, and radiologist/pathologist preferences. These variations affect both style and wording: some reports are highly structured, while others are narrative, and the same finding can be expressed with different terms (e.g., a “malignant mass” or a “lesion consistent with the patient’s known pancreatic cancer”). Semantics allows the LLM to recognize that these different expressions refer to the same underlying finding. Context is also essential, as tumor measurements may refer to current tumors or to tumors in prior exams. By combining semantic and contextual understanding, and by following clinical guidelines specified in their prompts, LLMs can accurately extract tumor information from reports.

As the LLM, we use Llama 3.1 70B AWQ \citep{dubey2024llama}. Its prompts explain how to interpret reports, and they were developed through iterative error analysis and refinement, in a collaboration between computer scientists and radiologists. The prompt is available in Appendix~\ref{app:prompts}. To reduce computational cost, the LLM is used only once, before training the segmenter. Our prompts instruct the LLM to return structured outputs, where the extracted tumor information fills a template. This template is then automatically parsed into a table, which is used as ground truth for the Volume Loss and Ball Loss. In addition to structured outputs, the LLM returns justifications for its answers and supporting quotes from the reports. We use in-context learning, providing the LLM with examples of reports and their desired outputs. By not fine-tuning the LLM, we avoid overfitting it to the style of reports in our datasets, enabling easy generalization across hospitals, and keeping computational costs low. A radiologist evaluated this LLM on 447 reports (182 with tumors) and found that it achieves 96\% accuracy in extracting tumor information. {\PBrev The tumor detection and segmentation performance of R-Super shows minimal degradation with LLM error rates of up to 10\% (Tab.~\ref{tab:ablations}).}

\subsection{Volume Loss}
\label{sec:volume_loss}

The Volume Loss minimizes the dissimilarity between the volume of segmented tumors ($V_{s,o}$) and the tumor volume estimated from radiology reports ($V_{r,o}$). Specifically, $V_{r,o}$ is the combined volume of all tumors reported in the organ or sub-segment $o$ (e.g., pancreatic head), and $V_{s,o}$ is the combined volume of all tumors segmented in organ or sub-segment $o$. By minimizing the dissimilarity between $V_{s,o}$ and $V_{r,o}$, the Volume Loss teaches the segmentation model to match the volume of tumors mentioned in the report, to avoid false positives (segmenting a tumor when the report mentions no tumor), and to avoid false negatives (segmenting no tumor when the report mentions tumors). The Volume Loss also penalizes tumors segmented outside of the organ/sub-segment $o$. 

\textbf{Main Equation.} Eq.~\ref{eq:vol_final} defines the Volume Loss for an organ $o$ (or organ sub-segment; we will say organ for simplicity). When the report mentions tumors in organ $o$, the Volume Loss has two terms: $L_{\text{forg},o}(V_{s,o},V_{r,o})$, which penalizes the dissimilarity between segmented and reported tumor volumes, and $L_{\text{bkg},o}(\mathbf{T}^{o})$, which penalizes tumors segmented outside organ $o$. When no tumors are reported in organ $o$, the loss penalizes any segmented tumor, using a per-voxel cross-entropy with target zero, $\text{CE}(\mathbf{T}^{o},\mathbf{0})$.
Here, $\mathbf{T}^{o}$ is the segmentation model's output channel for tumors in organ $o$ (e.g., the pancreatic tumor channel), made of per-voxel estimated tumor probabilities. When applying the Volume Loss as deep supervision, $\mathbf{T}^{o}$ is produced by a $1{\times}1{\times}1$ convolution with sigmoid activation, applied to an intermediate layer of the segmentation model's decoder, then up-sampled to the size of the segmentation model's input. The total Volume Loss is the sum of $L_{\text{vol},o}$ over all organs of interest $\mathcal{O}$ (Eq.~\ref{eq:vol_all_organs}). Next, we break down the calculation of all terms in the Volume Loss.


\begin{gather}
\label{eq:vol_final}
L_{\text{vol},o}=
\begin{cases}
\begin{aligned}
&L_{\text{forg},o}(V_{s,o},V_{r,o}) \\
&\quad + L_{\text{bkg},o}(\mathbf{T}^{o})
\end{aligned} & \text{if tumors in } o \\[1ex]
\text{CE}(\mathbf{T}^{o},\mathbf{0}) & \text{otherwise}
\end{cases}
\\
\label{eq:vol_all_organs}
L_{\text{vol}} = \sum_{o\in\mathcal{O}}L_{\text{vol},o}
\end{gather}

\textbf{Estimating tumor volumes from radiology reports ($V_{r,o}$).} Most radiology reports do not report tumor volumes, but they often report tumor diameters. One, two, or three perpendicular diameters are reported per tumor ($d_1$, $d_2$, $d_3$). For instance, the WHO standard \citep{miller1981reporting} uses two diameters, and RECIST (Response Evaluation Criteria in Solid Tumors) uses one \citep{eisenhauer2009new}. When the report provides a single diameter for a tumor, we estimate its volume as a ball's volume, $d_1^3\pi/6$. When three diameters are provided, we use the ellipsoid volume, $d_1d_2d_3\pi/6$. When two diameters are provided, we estimate the third as the average of the other two, and use the ellipsoid volume formula again, $d_1d_2((d_1+d_2)/2)\pi/6$. We calculate the volume for each tumor reported in organ $o$, and sum them, obtaining the total reported tumor volume, $V_{r,o}$.

\textbf{Estimating tumor volumes from segmentation ($V_{s,o}$).} To estimate the total volume of all tumors segmented inside organ $o$, $V_{s,o}$, we could simply count the tumor voxels segmented inside organ $o$ and multiply the result by the volume of one voxel, $v$. This is similar to element-wise multiplying the tumor segmentation output ($\mathbf{T}^{o}=[t_{x,y,z}^{o}]$) with the organ segmentation mask ($\mathbf{O}=[o_{x,y,z}]$), summing over the spatial dimensions ($x,y,z$), and multiplying the result by the voxel volume $v$ (Eq.~\ref{eq:vol_calculation}). Here, $\mathbf{O}$ is a binary mask, equal to 1 inside organ $o$ and 0 outside. Multiplying by $\mathbf{O}$ ensures that $V_{s,o}$ counts only tumors segmented inside organ $o$. If $\mathbf{T}^{o}$ were binary, Eq.~\ref{eq:vol_calculation} would be exactly equivalent to counting tumor voxels. Since $\mathbf{T}^{o}$ contains tumor probabilities, Eq.~\ref{eq:vol_calculation} provides a soft estimate of tumor volume. Importantly, this estimate is differentiable, and a correct tumor segmentation would yield $V_{s,o} = V_{r,o}$.

\begin{gather}
    \label{eq:vol_calculation}
    V_{s,o} = v \sum_{x,y,z} t_{x,y,z}^{o} o_{x,y,z}
\end{gather}

\textbf{Tumor volume dissimilarity ($L_{\text{forg},o}(V_{s,o},V_{r,o})$).} To enforce the volume of segmented tumors to match the tumor volume estimated from the radiology reports, we minimize the dissimilarity between $V_{s,o}$ and $V_{r,o}$. Specifically, we minimize the dissimilarity function $L_{\text{forg},o}(V_{s,o},V_{r,o})$ defined in Eq.~\ref{eq:loss_formula}. In the equation, $E$ is a small constant (set to 500 mm$^{3}$) that provides numerical stability for small $V_{r,o}$, and $\tau \in (0,1)$ is a tolerance margin (we set $\tau=0.1$). Three key properties of $L_{\text{forg},o}(V_{s,o},V_{r,o})$ are observable in its plot in Fig.~\ref{fig:method}. \textit{First}, $L_{\text{forg},o}(V_{s,o},V_{r,o})$ has a steep but finite gradient at $V_{s,o}=0$ (if $V_{r,o}\neq0$). Thus, the Volume Loss strongly penalizes false negatives (missed tumors) while remaining numerically stable. \textit{Second}, the gradient is less steep when $V_{s,o}>V_{r,o}$, because strong gradients in this region could enforce false negatives, by pushing the segmentation model towards $V_{s,o}=0$. \textit{Third}, the Volume Loss has a tolerance margin: $L_{\text{forg},o}(V_{s,o},V_{r,o})$ and its gradient are zero when $V_{s,o}$ is very close to $V_{r,o}$ (we use a tolerance, $\tau$, of 10\%). This margin compensates for inaccuracies in $V_{r,o}$, which arise from human error in measuring tumor diameters in reports, and from the approximation errors when we convert those diameters into volumes. 
Therefore, the segmentation model is not penalized if the tumors it segments have a volume that is close but not equal to $V_{r,o}$.

\begin{gather}
\label{eq:loss_formula}
\begin{aligned}
L_{\text{forg},o}(V_{s,o},V_{r,o}) = \max\{L_{\text{forg},o}'(V_{s,o},V_{r,o})\\-L_{\text{forg},o}'((1-\tau) V_{r,o},V_{r,o}),0\} \end{aligned}\\
L_{\text{forg},o}'(V_{s,o},V_{r,o}) = \frac{|V_{s,o} - V_{r,o}|}{V_{s,o} + V_{r,o} + E}
\end{gather}

{\emergencystretch=2em\relax
\textbf{Avoiding tumors in wrong locations ($L_{\text{bkg},o}(\mathbf{T}^{o})$).} The minimization of $L_{\text{forg},o}(V_{s,o},V_{r,o})$ teaches the segmentation model to segment tumors inside the organ $o$ where the report mentions tumors. However, we must also penalize tumors segmented outside organ $o$ (e.g., pancreatic tumors segmented outside of the pancreas). To this end, the Volume Loss includes the term $L_{\text{bkg},o}(\mathbf{T}^{o})$, defined in Eq.~\ref{eq:background_volume}. This term uses voxel-wise cross-entropy to penalize tumors segmented outside the organ $o$, located by its organ segmentation mask $\mathbf{O}$. Specifically, we element-wise multiply $(1-\mathbf{O})\odot\mathbf{T}^{o}$, capturing tumors segmented outside organ $o$. Afterwards, we penalize these tumors with a cross-entropy against a zero target. Organ masks were dilated (by 2 cm) to compensate for tumors that grow beyond organ borders.\par}

\begin{gather}
    \label{eq:background_volume}
    L_{\text{bkg},o}(\mathbf{T}^{o}) = \mathrm{CE}((1-\mathbf{O})\odot\mathbf{T}^{o},\mathbf{0})
\end{gather}

\textbf{Reports missing tumor count or diameters.} The estimation of $V_{r,o}$ requires the report to provide diameters for every tumor in organ $o$. When a report mentions tumors in $o$ but omits the number of tumors or their diameters, we fall back to a relaxed form of the loss: it requires the segmentation model to segment \textit{at least one} small tumor inside organ $o$, but does not penalize segmenting a larger tumor or multiple tumors. The full reformulation is given in Appendix~\ref{app:loss_details}.

\textbf{Input cropping.} Due to memory constraints, tumor segmentation models are often trained on patches rather than full 3D images. This creates a problem for the Volume Loss (and for the Ball Loss): if organ $o$ is only partially contained in a training patch, we cannot know whether a tumor reported in $o$ is inside the patch or outside. The Volume Loss and the Ball Loss cannot be applied in this case. To avoid this, we use the organ masks to guide cropping, ensuring that each training patch fully contains at least one organ $o$. The cropped organ $o$ is randomly selected, with a high probability (80\%) of cropping on organs with tumors. If the patch partially contains a second organ with tumors, we skip the Volume Loss and the Ball Loss for this second organ.

\subsection{Ball Loss}
\label{sec:ball_loss}

The Ball Loss encourages the segmentation model to segment tumors that match radiology reports in tumor count, location (organ/organ sub-segment), and diameters. The Ball Loss localizes, in the output of the tumor segmentation model, the most likely location for each tumor mentioned in the radiology report. Then, each of these located tumors is optimized to match the corresponding tumor diameter in the report. 
The Ball Loss can be defined by a sequence of procedures, illustrated in Fig.~\ref{fig:method} and described below.

\textbf{Locate the organ $o$ and penalize tumors segmented outside of it.} As in the Volume Loss, we first locate the organ (or organ sub-segment) $o$ in the output of the tumor segmentation model, using the pre-saved organ segmentation mask $\mathbf{O}$. In case the report mentions no tumor in organ $o$, we use cross-entropy with zero target to minimize the entire tumor segmentation output for that organ, $L_{\text{ball},o} = \text{CE}(\mathbf{T}^{o},\mathbf{0})$. Otherwise, we multiply the tumor segmentation output ($\mathbf{T}^{o}$, per-voxel probabilities for tumors in organ $o$) by the organ segmentation mask for organ $o$, $\mathbf{O}\odot\mathbf{T}^{o}$. This step removes tumors segmented outside organ $o$ before the next step, when we locate the tumors reported in organ $o$.

\textbf{{\PBrev Find} the most likely {\PBrev location} for a tumor $i$ described in the report (Ball Convolution).} We begin by searching for the largest tumor that the report describes in organ $o$, tumor $i=1$ with diameter $d_i$ (the largest diameter reported for tumor $i$). To {\PBrev find} the most likely {\PBrev location}, we use a convolution that we call \textit{Ball Convolution}. It is a standard 3D convolution (stride 1, odd-sized kernel, zero padding of $(k-1)/2$ voxels for a kernel of size $k$, so that the output keeps the input size), but its kernel is not learnable. Its kernel $\mathbf{k}_i$ is a binary ball with diameter matching the tumor diameter $d_i$ in the report (Eq.~\ref{eq:ball_kernel}). The convolution operation moves this ball over the output of the tumor segmentation model, $\mathbf{T}^{o}$ (which we previously multiplied by the organ mask, $\mathbf{O}$), producing $\mathbf{B}_i$, as shown in Eq.~\ref{eq:ball_conv}. At each location, the ball sums the per-voxel tumor probabilities inside it. Therefore, when the ball is at the most likely location for a tumor of diameter $d_i$ (the reported tumor diameter), the output of the Ball Convolution is the highest. That is, the location of the maximum output of the Ball Convolution, $\mathbf{c}_i=[c_x,c_y,c_z]$ (Eq.~\ref{eq:ball_max}), is the most likely center for a {\PBrev ball encompassing} the tumor of the reported diameter $d_i$. This center must be inside the organ $o$, due to the multiplication of the tumor segmentation output $\mathbf{T}^{o}$ and the organ mask $\mathbf{O}$ in the last step. After finding $\mathbf{c}_i$, we define the \textit{highest-probability ball}, $\mathbf{H}_i$, as a ball of diameter $(1+\mu)d_i$ and center $\mathbf{c}_i$ (Eq.~\ref{eq:highest_prob_ball}). We expand the highest-probability ball by $\mu=0.2$ (20\%) to account for inaccuracies in the reported diameter $d_i$. {\PBrev Even if a tumor has an irregular shape, it is expected to fit inside a ball of diameter $(1+\mu)d_i$. That is because $d_i$ is defined as the maximum distance between two tumor points in any axial plane, according to most radiology reporting guidelines (e.g., WHO and RECIST). Our $(1+\mu)d_i$ expansion accommodates diameter measurement inaccuracies and larger out-of-plane tumor extents.}

\begin{gather}
\label{eq:ball_kernel}
\mathbf{k}_{i}(u,v,w)=
\begin{cases}
1, & \text{if } \sqrt{u^{2}+v^{2}+w^{2}} \le \dfrac{d_i}{2} \\
0, & \text{otherwise}
\end{cases}
\\
\label{eq:ball_conv}
\mathbf{B}_i = (\mathbf{O}\odot \mathbf{T}^{o}) * \mathbf{k}_{i}
\\
\label{eq:ball_max}
\mathbf{c}_i=[c_x,c_y,c_z]
=
\underset{(x,y,z)}{\mathrm{argmax}}\,\mathbf{B}_i(x,y,z) \\
\label{eq:highest_prob_ball}
\mathbf{H}_i(x,y,z)=
\begin{cases}
1, & \text{if } \| (x,y,z) - \mathbf{c}_i \| \le \dfrac{d_i(1+\mu)}{2} \\
0, & \text{otherwise}
\end{cases}
\end{gather}

\textbf{Locate the $N_i$ most likely voxels for a tumor $i$ described in the report.} The Ball Loss does not maximize the tumor probability at all voxels inside the highest-probability ball $\mathbf{H}_i$, because we do not want to enforce the segmentation model to segment perfectly spherical tumors. Instead, the Ball Loss maximizes the predicted tumor probability at the top-$N_i$ voxels inside the highest-probability ball ($\widetilde{\mathbf{T}}^{o}_i$, Eq.~\ref{eq:t_tilde}). These top-$N_i$ voxels can assume any shape that \textit{fits inside} the highest-probability ball. We obtain $N_i$ by converting the volume of tumor $i$, estimated from the report (Sec.~\ref{sec:volume_loss}), to a voxel count, and reducing it by a small margin (20\%). From these top-$N_i$ voxels we build a binary mask $\mathbf{M}_i$, with value 1 at the top-$N_i$ voxels and 0 elsewhere (Eq.~\ref{eq:mask}, where $\operatorname{top}_{[a,b]}(\cdot)$ denotes the voxels ranked $a$ to $b$ by predicted tumor probability). We then build a second mask $\mathbf{M}'_i$, with value 1 at the voxels ranked $N_i+1$ to $\nu N_i$ (we set $\nu=1.2$) by tumor probability inside the highest-probability ball (Eq.~\ref{eq:margin}). $\mathbf{M}_i$ thus marks the most probable tumor voxels for tumor $i$, and $\mathbf{M}'_i$ marks a tolerance margin around $\mathbf{M}_i$. This tolerance margin accounts for uncertain tumor borders and tumor measurement error in the reports.

\begin{gather}
\label{eq:t_tilde}
\widetilde{\mathbf{T}}^{o}_i = \mathbf{T}^{o} \odot \mathbf{H}_i\\
\label{eq:mask}
\mathbf{M}_i(x,y,z) =
\begin{cases}
1, & \text{if } (x,y,z) \in \operatorname{top}_{[1,\,N_i]}(\widetilde{\mathbf{T}}^{o}_i) \\
0, & \text{otherwise}
\end{cases} \\
\label{eq:margin}
\mathbf{M}'_i(x,y,z) =
\begin{cases}
1, & \text{if } (x,y,z) \in \operatorname{top}_{[N_i+1,\,\nu N_i]}(\widetilde{\mathbf{T}}^{o}_i) \\
0, & \text{otherwise}
\end{cases}
\end{gather}

\textbf{Maximize and minimize tumor probabilities.} In case the report mentions a single tumor $i$ in organ $o$, the Ball Loss maximizes the segmentation model's output $\mathbf{T}^{o}$ at the $N_i$ most likely tumor voxels (mask $\mathbf{M}_i$), leaves the surrounding margin (mask $\mathbf{M}'_i$) unpenalized, and minimizes $\mathbf{T}^{o}$ everywhere else. We achieve all three behaviors with a single cross-entropy plus Dice loss, using $\mathbf{M}_i$ as the target and $\mathbf{T}^{o}$ as the prediction (Eq.~\ref{eq:ce_dice}, assuming $\mathbf{M}=\mathbf{M}_i$ and $\mathbf{M}'=\mathbf{M}'_i$ in case of a single tumor in organ $o$). To avoid penalizing the margin voxels $\mathbf{M}'_i$, we multiply $\mathbf{T}^{o}$ by $(\mathbf{1}-\mathbf{M}'_i)$ before computing the loss, zeroing the gradient at those voxels (Eq.~\ref{eq:ce_dice}). The resulting loss maximizes tumor probabilities at the most likely tumor voxels given the report and the segmentation model's output, accounting for uncertainty.

\begin{gather}
\label{eq:ce_dice}
\begin{aligned}
L'_{\text{ball},o}=\mathrm{CE}((1-\mathbf{M}')\odot\mathbf{T}^{o},\mathbf{M})
    +\\
    \mathrm{Dice}((1-\mathbf{M}')\odot\mathbf{T}^{o},\mathbf{M})\end{aligned}\\
L_{\text{ball},o}
=
\begin{cases}
    L'_{\text{ball},o}, & \text{if tumors in } o \\
\text{CE}(\mathbf{T}^{o},\mathbf{0}), & \text{otherwise}
\end{cases}\\
\label{eq:ball_all_organs}
L_{\text{ball}} = \sum_{o\in\mathcal{O}}L_{\text{ball},o}
\end{gather}

\textbf{Reports with multiple tumors.} When a report mentions multiple tumors in organ $o$, we locate them sequentially from largest to smallest, applying the procedure above for each tumor $i$: the Ball Convolution, then the creation of the masks $\mathbf{M}_i$ and $\mathbf{M}'_i$. To avoid mask overlaps between tumors, after locating tumor $i$, we zero its most probable voxels and uncertainty border ($\mathbf{M}_i \cup \mathbf{M}'_i$) in a copy of the segmentation output, $\mathbf{T}^{o}\leftarrow\mathbf{T}^{o}\odot(\mathbf{1}-(\mathbf{M}_i \cup \mathbf{M}'_i))$, which is used only to locate the remaining tumors. Thus, the next Ball Convolution (to locate tumor $i+1$) cannot place its center inside the voxels already assigned to tumor $i$. Once all reported tumors in $o$ are located, we take the union of their per-tumor masks ($\mathbf{M}_i$ and $\mathbf{M}_i'$) to form the combined masks $\mathbf{M}$ and $\mathbf{M}'$ (Eq.~\ref{eq:union_masks}, where $K^o$ is the number of tumors reported in organ $o$). Then, these combined masks ($\mathbf{M}$ and $\mathbf{M}'$) guide the cross-entropy and Dice losses in Eq.~\ref{eq:ce_dice}. When tumors are reported across multiple organs, the final Ball Loss is the sum of the per-organ losses (Eq.~\ref{eq:ball_all_organs}).

\begin{equation}
\mathbf{M} = \bigcup_{i=1}^{K^o} \mathbf{M}_i,
\qquad
\mathbf{M}' = \bigcup_{i=1}^{K^o} \mathbf{M}'_i
\label{eq:union_masks}
\end{equation}

\textbf{Reports missing tumor count or diameters.} The presented formulation of the Ball Loss needs the report to provide diameters for all tumors in the organ $o$. When a report misses tumor diameters or counts, we use a high-tolerance variant of the Ball Loss (Appendix~\ref{app:loss_details}). It encourages the segmentation model to segment at least one small tumor in $o$. However, it does not penalize the segmentation model for segmenting larger tumors or multiple tumors. 

As final details, to further compensate for uncertainty at tumor borders, our implementation of cross-entropy (Eq.~\ref{eq:ce_dice}) is weighted, giving higher weights to voxels where the segmentation model predicts higher tumor probabilities. The Ball Convolution kernel also has a small Gaussian decay, with values larger near the ball center, slightly smaller near the ball boundary, and zero outside the ball. Because predicted tumor probabilities tend to peak at tumor centers, this decay improves the alignment between the ball center and the tumor center. The Gaussian decay and the weighted cross-entropy are formally defined in Appendix~\ref{app:loss_details}.


\begin{figure}
    \centering
    \includegraphics[width=1\linewidth]{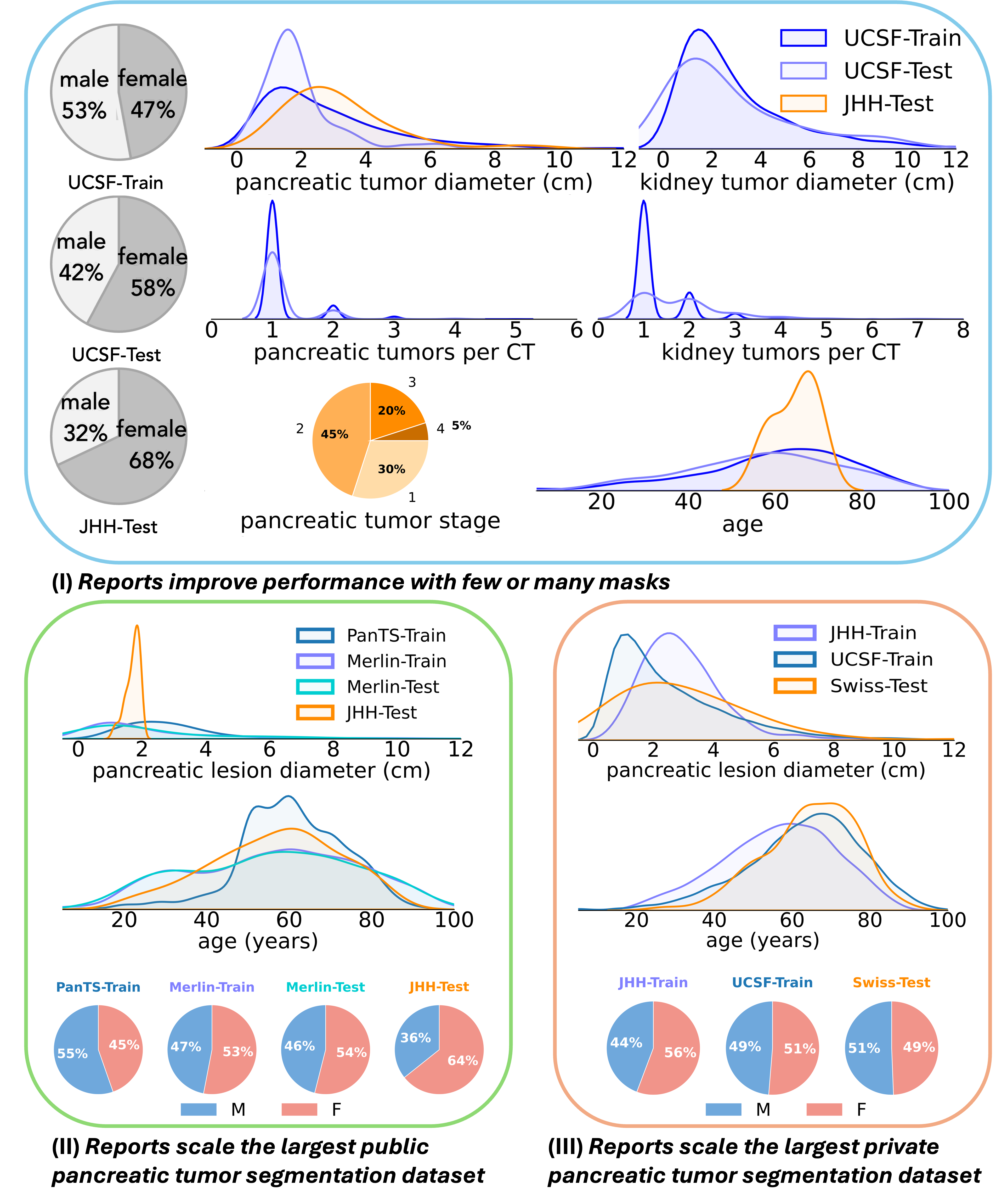}
    \caption{\textbf{Dataset overview (Secs.~\ref{sec:miccai_results} to \ref{sec:jhh_ucsf}).} Summary of all datasets, including CT-Mask and CT-Report cohorts from multiple institutions. In (I), (II) and (III), we show the number of tumor scans, tumor types (pancreatic and kidney), and distributions of tumor size, tumor count per scan, and patient demographics (age and sex) across training and test sets (UCSF, JHH, and public datasets). Radiology reports enable large-scale dataset construction \citep{li2025scalemai}, which R-Super leverages for segmentation training.}
    \label{fig:dataset_stats}
\end{figure}

\section{Results}

We present experiments at three scales of training data, using both public and private datasets. Sec.~\ref{sec:miccai_results} trains on 6,718 CT-Report pairs from a private dataset, combined with 50 to 1,674 public tumor masks. Results are provided on both pancreatic and kidney tumors. This section also includes comprehensive comparisons against six alternative methods; ablation studies follow in Sec.~\ref{sec:ablations}. Sec.~\ref{sec:pants} uses only public training data, demonstrating that R-Super can scale the largest public pancreatic tumor mask dataset (PanTS, 926 CT-Mask pairs) by adding CT-Report pairs from Merlin (1,848 CT-Report pairs). Sec.~\ref{sec:jhh_ucsf} reaches the largest scale in this paper, with over 47,000 training CT scans. 

\begin{figure}
    \centering
    \includegraphics[width=1\linewidth]{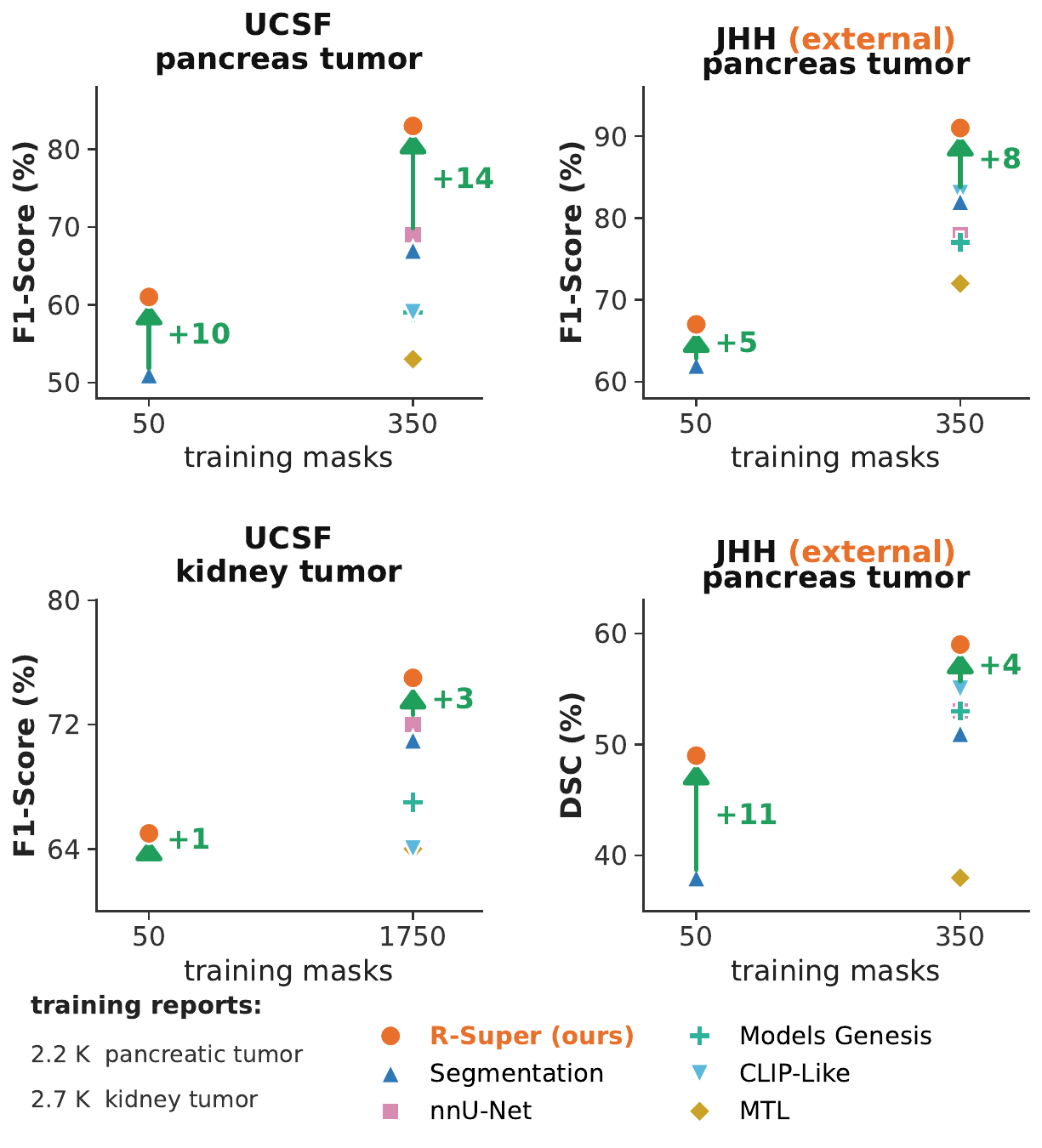}
    \caption{\textbf{By learning from reports, R-Super substantially outperforms five state-of-the-art methods on both internal (UCSF-Test) and external (JHH-Test) validation} (all six compared methods are in Tab.~\ref{tab:all_results}). Green arrows show the gain of R-Super over the best competing method at each mask budget, up to +14\% in tumor detection F1-Score and +11\% in DSC; over standard segmentation (no reports) alone, the gain reaches +16\% in F1-Score (Tab.~\ref{tab:all_results}). Each panel corresponds to one tumor type and one test set, and its x-axis shows the number of training \textit{masks} (50 from AbdomenAtlas-Subset; 344 or 1.7K from AbdomenAtlas 3.0; the axis labels 350 and 1750 are rounded). R-Super and the other report-based methods shown (CLIP, MTL) supplement these masks with 2.2K pancreatic and 2.7K kidney tumor \textit{reports} from UCSF-Train. Both test sets are challenging: JHH-Test is out-of-distribution relative to training, and UCSF-Test spans a wide range of CT resolutions and contrast phases, including non-contrast scans. {\PBrev DSC is only reported for JHH-Test because UCSF-Test does not include ground-truth tumor masks, needed for calculating DSC.}}
    \label{fig:all_plots}
\end{figure}



\textbf{Training procedure.} 
R-Super is a training methodology that can be applied to any segmentation architecture. In our experiments, we adopted MedFormer \citep{gao2022data}, a hybrid transformer-CNN architecture that has topped prior benchmarks \citep{bassi2024touchstone}, and we kept its default training hyper-parameters (Appendix~\ref{supp:training}). Because the competing approaches (Models Genesis, MTL, CLIP, report-guided pseudo-labels, and standard mask-based segmentation) are likewise training strategies rather than architectures, we paired each of them with the same MedFormer backbone and training hyper-parameters, so that performance differences reflect the training strategy rather than implementation choices. The MTL baseline additionally required a classification head, which we attached to the output of MedFormer's encoder. The only exception was nnU-Net \citep{isensee2021nnu}, which we ran with its self-configuring ResEncL architecture and automatic hyper-parameters. 

{\PBrev \textbf{Little training overhead and no inference overhead.} Relative to a standard segmentation model using the same segmentation architecture (MedFormer), the R-Super training strategy increases training time by only 30\% and does not change inference time. R-Super was trained for approximately 105 hours on one NVIDIA H100 GPU, with a peak GPU memory usage of 46 GB. Memory usage can be reduced to a minimum of 23 GB per GPU by training on two GPUs with a batch size of one per GPU. Each report is processed by an LLM only once, before training; the LLM Llama 3.1 70B AWQ processes approximately 250 reports per hour, using vLLM on one NVIDIA RTX 6000 GPU. At inference, R-Super processes one CT scan in approximately 1.2 minutes on one NVIDIA H100 GPU, using only 10 GB of GPU memory. This modest memory requirement enables deployment on less powerful, more widely available GPUs, supporting use in resource-constrained clinical settings. Due to its small training overhead and no test overhead, R-Super’s training and inference times and memory usage are determined primarily by the underlying segmentation architecture, which can be arbitrarily chosen.}

\subsection{Reports Improve Performance with Few or Many Masks}
\label{sec:miccai_results}

In our first experiment, we train R-Super with CT-Report pairs plus a varying number of masks, from 50 to 1,674. We assess its performance on internal validation (on the same hospital that provided the CT-Report pairs) and external validation (a completely unseen hospital). The objective is to demonstrate that including many CT-Report pairs in the training set is useful whether few or many masks are available.


\textbf{Datasets.}
\textbf{(I) AbdomenAtlas 3.0} \citep{bassi2025radgpt} is a large public CT dataset with 9,262 CT-Mask pairs collected from 88 hospitals in 19 countries, including 344 pancreatic tumor CTs and 1,674 kidney tumor CTs.
\textbf{(II) AbdomenAtlas-Subset} is a subset of AbdomenAtlas 3.0 {\PBrev with 200 CT-Mask pairs, including} 50 pancreatic tumor CTs, 50 kidney tumor CTs, and 100 normal CTs.
\textbf{(III) UCSF-Train} is a large dataset (private) from UCSF and its affiliated institutions in California, USA. It has 6,718 CT-Report pairs: 2,229 pancreatic tumor CTs, 2,738 kidney tumor CTs, and 1,751 normal CTs. {\PBrev UCSF-Train does not include tumor masks.}
\textbf{(IV) UCSF-Test} is a test set drawn from the same hospital as UCSF-Train, including 169 kidney tumor CTs, 139 pancreatic tumor CTs, and 100 normal CTs. It was created by randomly splitting a dataset into training and testing. Therefore, UCSF-Test has a data distribution similar to that of UCSF-Train. {\PBrev UCSF-Test does not include tumor masks.} Tumor labels and sizes for UCSF-Test come from its radiology reports. 
\textbf{(V) JHH-Test} \citep{xia2022felix,park2020annotated} is an external test set from an unseen hospital, the Johns Hopkins Hospital (Maryland, USA), with 50 pancreatic tumor CTs and 50 normal CTs. No data from this hospital was seen by any of the AI models in training, making this dataset out-of-distribution. {\PBrev JHH-Test includes radiologist-drawn tumor masks.} More details on the datasets are available in Fig.~\ref{fig:dataset_stats}.

\begin{table*}[!t]
\centering
\scriptsize
\setlength{\tabcolsep}{5.4pt}
\caption{\textbf{R-Super outperforms state-of-the-art training methods on tumor detection and segmentation.} On pancreatic tumors, R-Super achieves the highest DSC, NSD, F1-Score, and AUC across all baselines. On kidney tumors, it matches mask-based segmentation when few masks are available (50) and surpasses it when many are available (1.7K). DSC and NSD are computed on scans with tumors and reported only in datasets where masks exist (UCSF-Test has none). Sensitivity (Se) and Specificity (Sp) correspond to the threshold that maximizes F1-Score. Statistical tests were performed to compare F1-Score and AUC (paired permutation test for F1-Score and DeLong's test for AUC). Orange highlights statistically significant gains over standard segmentation ($p<0.05$). {\PBrev Standard deviations and confidence intervals are available in Tab.~\ref{tab:cif_all_results} (Appendix~\ref{app:variability}).}}
\begin{tabular}{l*{19}{c}}
\toprule
 & \multicolumn{12}{c}{\footnotesize pancreas tumor} & \multicolumn{6}{c}{\footnotesize kidney tumor} \\
\cmidrule(lr){2-13}\cmidrule(lr){14-19}
 & \multicolumn{8}{c}{\scriptsize JHH-Test} & \multicolumn{4}{c}{\scriptsize UCSF-Test} & \multicolumn{6}{c}{\scriptsize UCSF-Test} \\
\cmidrule(lr){2-9}\cmidrule(lr){10-13}\cmidrule(lr){14-19}
\scriptsize train paradigm & mask & rep. & DSC & NSD & F1 & AUC & Se & Sp & F1 & AUC & Se & Sp & mask & rep. & F1 & AUC & Se & Sp \\
\midrule
\multicolumn{19}{l}{\textit{few training masks (50)}} \\
\href{https://arxiv.org/abs/2203.00131}{standard segmentation} & 50 & 0 & 38 & 41 & 62 & 63 & 62 & 62 & 51 & 63 & 47 & 77 & 50 & 0 & 64 & 68 & 68 & 63 \\
\textbf{R-Super (ours)} & 50 & 2.2K & 49 & 52 & 67 & 75 & 68 & 64 & 61 & 70 & 73 & 59 & 50 & 2.7K & 65 & 66 & 73 & 55 \\
\midrule
\multicolumn{19}{l}{\textit{medium / many training masks (344 / 1.7K)}} \\
\href{https://www.nature.com/articles/s41586-026-10181-8}{CLIP-Like} & 344 & 2.2K & 55 & 58 & 83 & 86 & 90 & 70 & 59 & 74 & 59 & 75 & 1.7K & 2.7K & 64 & 71 & 75 & 48 \\
\href{https://ieeexplore.ieee.org/document/8759483/}{Multi-task learning} & 344 & 2.2K & 38 & 43 & 72 & 80 & 78 & 60 & 53 & 62 & 65 & 50 & 1.7K & 2.7K & 64 & 71 & 68 & 63 \\
\href{https://pubs.rsna.org/doi/full/10.1148/ryai.230031}{Report-G Pseudo-labels} & 344 & 2.2K & 56 & 62 & 80 & 79 & 80 & 78 & 69 & 83 & 65 & 86 & 1.7K & 2.7K & 71 & 74 & 80 & 60 \\
\href{https://doi.org/10.1016/j.media.2020.101840}{Models Genesis} & 344 & 0 & 53 & 57 & 77 & 81 & 78 & 74 & 59 & 72 & 66 & 63 & 1.7K & 0 & 67 & 73 & 77 & 54 \\
\href{https://www.nature.com/articles/s41592-020-01008-z}{nnU-Net} & 344 & 0 & 53 & 57 & 78 & 75 & 76 & 82 & 69 & 79 & 74 & 74 & 1.7K & 0 & 72 & 74 & 87 & 53 \\
\href{https://arxiv.org/abs/2203.00131}{standard segmentation} & 344 & 0 & 51 & 59 & 82 & 84 & 78 & 88 & 67 & 78 & 62 & 85 & 1.7K & 0 & 71 & 73 & 65 & 68 \\
\textbf{R-Super (ours)} & 344 & 2.2K & \textbf{59} & \textbf{69} & \cellcolor{lightorange}\textbf{91} & \cellcolor{lightorange}\textbf{92} & 94 & 88 & \cellcolor{lightorange}\textbf{83} & \cellcolor{lightorange}\textbf{90} & 83 & 89 & 1.7K & 2.7K & \cellcolor{lightorange}\textbf{75} & \cellcolor{lightorange}\textbf{78} & 80 & 70 \\
\bottomrule
\end{tabular}
\label{tab:all_results}
\end{table*}


\textbf{Training and Evaluation.} We trained two configurations of R-Super. The first trained on UCSF-Train (many CT-Report pairs) plus AbdomenAtlas-Subset (few CT-Mask pairs). The second version trained on UCSF-Train plus AbdomenAtlas 3.0 (more CT-Mask pairs). We trained first on the CT-Mask pairs, followed by fine-tuning on CT-Mask plus CT-Report pairs. Evaluation was performed on UCSF-Test (internal) and JHH-Test (external).

\begin{figure}
    \centering
    \includegraphics[width=1\linewidth]{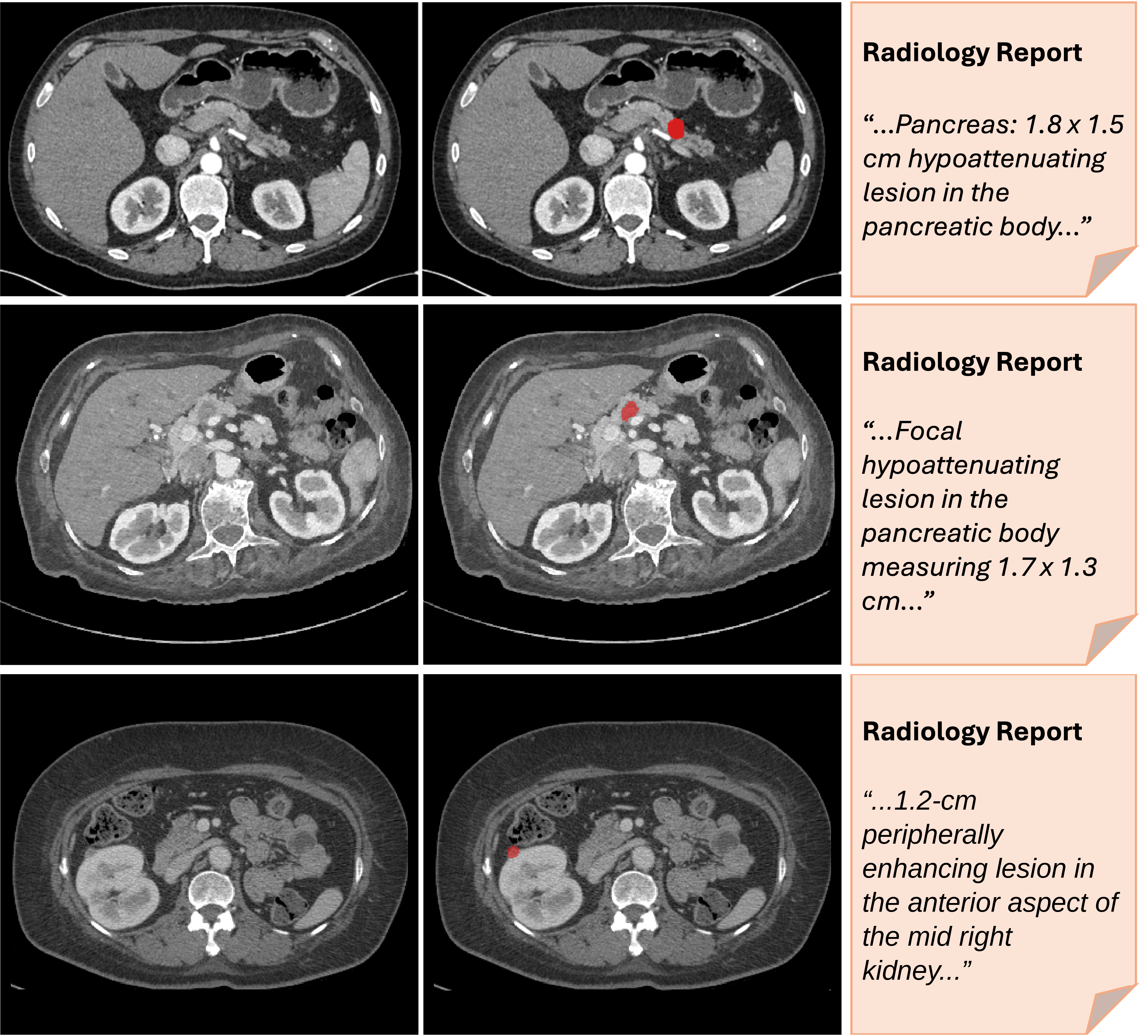}
    \caption{\textbf{Tumors segmented by R-Super.} The examples show two cases of small pancreatic adenocarcinoma {\PBrev and one case of small renal carcinoma (bottom)} correctly segmented by R-Super. Radiology reports are shown, but they are used only in training, not at inference.}
    \label{fig:qualitative_results}
\end{figure}

\textbf{R-Super improves tumor segmentation with few or many training masks.} Fig.~\ref{fig:all_plots} compares R-Super against standard segmentation (trained without reports); detailed results are provided in Tab.~\ref{tab:all_results}, and examples of outputs in Fig.~\ref{fig:qualitative_results}. Adding 2.2K pancreatic and 2.7K kidney tumor CT-Report pairs to training yielded consistent gains in F1-Score across every mask regime: up to +10\% with few masks (50, AbdomenAtlas-Subset), up to +16\% with a moderate number of masks (344, AbdomenAtlas 3.0), and +4\% with many masks (1.7K, AbdomenAtlas 3.0). Similar gains are observed in DSC: +11\% with 50 masks, and +8\% with 344 masks. We only report DSC for pancreatic tumors, because only JHH-Test includes masks. In summary, by leveraging reports, R-Super can improve both tumor detection and segmentation whether few or many training masks are available.

\textbf{R-Super generalizes to unseen hospitals.} Fig.~\ref{fig:all_plots} shows that R-Super substantially outperforms standard segmentation on UCSF-Test (drawn from the same hospital as UCSF-Train) and on JHH-Test, a hospital never seen during training (out-of-distribution). Performance on JHH-Test was actually higher than on UCSF-Test, likely reflecting data differences: JHH-Test has high resolution (0.5 mm) and arterial-phase contrast, whereas UCSF-Test mixes a range of slice thicknesses and includes both contrast and non-contrast scans. On both test sets, R-Super surpasses six state-of-the-art methods, three of which also use reports during training (CLIP-Like, MTL, report-guided pseudo-labels). The advantage is consistent across F1-Score, AUC, DSC, and NSD (Tab.~\ref{tab:all_results}).

\textbf{R-Super improves small tumor detection and segmentation.} With 344 or 1.7K training masks, R-Super improved tumor detection and segmentation across the full size spectrum, including the small tumors (diameter $\leq$ 2 cm) that matter most for early cancer detection \citep{li2026early}. Detailed results are presented in Tab.~\ref{tab:results_by_size}. On small pancreatic tumors in the external JHH-Test, R-Super reached an F1-Score of 80 and AUC of 89, exceeding standard segmentation by +15 F1 and +6 AUC.

\begin{table*}[t]
\centering
\scriptsize
\setlength{\tabcolsep}{5.4pt}
\caption{\textbf{R-Super outperforms state-of-the-art methods on both small (diameter $\leq$ 2 cm) and large tumors.} For pancreatic tumors (344 training masks), R-Super achieves the highest DSC, NSD, and F1-Score across all baselines in both small ($\leq$ 2 cm) and large ($>$ 2 cm) tumors, and the highest AUC except for small tumors in JHH-Test (CLIP-Like 90 vs. 89). For kidney tumors (1.7K training masks), R-Super leads in F1-Score and AUC on small tumors and in F1-Score on large tumors. DSC and NSD are computed on scans with tumors and reported only where masks exist (UCSF-Test has none). Sensitivity (Se) and Specificity (Sp) correspond to the threshold that maximizes F1-Score. Statistical tests were performed to compare F1-Score and AUC (paired permutation test for F1-Score and DeLong's test for AUC). Orange highlights statistically significant gains over standard segmentation ($p<0.05$). Test set sizes: JHH-Test contains 15 small and 35 large pancreatic tumors; UCSF-Test contains 90 small and 28 large pancreatic tumors (21 CTs with tumors of unknown size were excluded from this table) and 77 small and 65 large kidney tumors (27 excluded). Tumor sizes were extracted from reports in UCSF-Test and from masks in JHH-Test. {\PBrev Standard deviations and confidence intervals are available in Tab.~\ref{tab:cif_results_by_size} (Appendix~\ref{app:variability}).}}
\begin{tabular}{l*{19}{c}}
\toprule
 & \multicolumn{12}{c}{\footnotesize pancreas tumor} & \multicolumn{6}{c}{\footnotesize kidney tumor} \\
\cmidrule(lr){2-13}\cmidrule(lr){14-19}
 & \multicolumn{8}{c}{\scriptsize JHH-Test} & \multicolumn{4}{c}{\scriptsize UCSF-Test} & \multicolumn{6}{c}{\scriptsize UCSF-Test} \\
\cmidrule(lr){2-9}\cmidrule(lr){10-13}\cmidrule(lr){14-19}
\scriptsize train paradigm & mask & rep. & DSC & NSD & F1 & AUC & Se & Sp & F1 & AUC & Se & Sp & mask & rep. & F1 & AUC & Se & Sp \\
\midrule
\multicolumn{19}{c}{\textbf{small tumors (diameter $\leq$ 2 cm)}} \\
\midrule
\addlinespace[2pt]
\multicolumn{19}{l}{\textit{few training masks (50)}} \\
\href{https://arxiv.org/abs/2203.00131}{standard segmentation} & 50 & 0 & 7 & 17 & 38 & 61 & 53 & 62 & 41 & 59 & 40 & 77 & 50 & 0 & 43 & 59 & 53 & 63 \\
\textbf{R-Super (ours)} & 50 & 2.2K & 15 & 25 & 35 & 68 & 47 & 64 & 54 & 72 & 76 & 59 & 50 & 2.7K & 47 & 61 & 65 & 55 \\
\midrule
\multicolumn{19}{l}{\textit{medium / many training masks (344 / 1.7K)}} \\
\href{https://www.nature.com/articles/s41586-026-10181-8}{CLIP-Like} & 344 & 2.2K & 11 & 19 & 68 & \textbf{90} & 100 & 71 & 50 & 74 & 54 & 75 & 1.7K & 2.7K & 40 & 65 & 57 & 48 \\
\href{https://ieeexplore.ieee.org/document/8759483/}{Multi-task learning} & 344 & 2.2K & 15 & 26 & 54 & 83 & 87 & 60 & 42 & 61 & 60 & 50 & 1.7K & 2.7K & 46 & 65 & 57 & 63 \\
\href{https://pubs.rsna.org/doi/full/10.1148/ryai.230031}{Report-G Pseudo-labels} & 344 & 2.2K & 19 & 32 & 61 & 77 & 73 & 80 & 63 & 82 & 62 & 86 & 1.7K & 2.7K & 50 & 71 & 66 & 60 \\
\href{https://doi.org/10.1016/j.media.2020.101840}{Models Genesis} & 344 & 0 & 10 & 20 & 62 & 85 & 80 & 76 & 48 & 70 & 60 & 63 & 1.7K & 0 & 43 & 65 & 60 & 54 \\
\href{https://www.nature.com/articles/s41592-020-01008-z}{nnU-Net} & 344 & 0 & 7 & 17 & 55 & 74 & 60 & 82 & 60 & 78 & 70 & 74 & 1.7K & 0 & 53 & 71 & 79 & 53 \\
\href{https://arxiv.org/abs/2203.00131}{standard segmentation} & 344 & 0 & 17 & 36 & 65 & 83 & 67 & 88 & 59 & 77 & 58 & 85 & 1.7K & 0 & 39 & 69 & 44 & 68 \\
\textbf{R-Super (ours)} & 344 & 2.2K & \textbf{25} & \textbf{48} & \textbf{80} & 89 & 93 & 88 & \cellcolor{lightorange}\textbf{75} & \cellcolor{lightorange}\textbf{90} & 77 & 89 & 1.7K & 2.7K & \textbf{56} & \cellcolor{lightorange}\textbf{78} & 69 & 69 \\
\midrule[1pt]
\multicolumn{19}{c}{\textbf{large tumors (diameter > 2 cm)}} \\
\midrule
\addlinespace[2pt]
\multicolumn{19}{l}{\textit{few training masks (50)}} \\
\href{https://arxiv.org/abs/2203.00131}{standard segmentation} & 50 & 0 & 3 & 5 & 59 & 63 & 65 & 62 & 35 & 78 & 61 & 77 & 50 & 0 & 56 & 75 & 82 & 63 \\
\textbf{R-Super (ours)} & 50 & 2.2K & 8 & 13 & 67 & 78 & 76 & 64 & 31 & 75 & 79 & 59 & 50 & 2.7K & 54 & 72 & 86 & 55 \\
\midrule
\multicolumn{19}{l}{\textit{medium / many training masks (344 / 1.7K)}} \\
\href{https://www.nature.com/articles/s41586-026-10181-8}{CLIP-Like} & 344 & 2.2K & 13 & 19 & 75 & 84 & 85 & 71 & 43 & 84 & 82 & 75 & 1.7K & 2.7K & 54 & 77 & 95 & 48 \\
\href{https://ieeexplore.ieee.org/document/8759483/}{Multi-task learning} & 344 & 2.2K & 18 & 28 & 63 & 78 & 74 & 60 & 28 & 70 & 82 & 50 & 1.7K & 2.7K & 54 & 77 & 78 & 63 \\
\href{https://pubs.rsna.org/doi/full/10.1148/ryai.230031}{Report-G Pseudo-labels} & 344 & 2.2K & 19 & 31 & 78 & 79 & 82 & 80 & 55 & 90 & 82 & 86 & 1.7K & 2.7K & 62 & 75 & 97 & 60 \\
\href{https://doi.org/10.1016/j.media.2020.101840}{Models Genesis} & 344 & 0 & 13 & 21 & 72 & 79 & 76 & 76 & 36 & 83 & 86 & 63 & 1.7K & 0 & 58 & \cellcolor{lightorange}\textbf{80} & 97 & 54 \\
\href{https://www.nature.com/articles/s41592-020-01008-z}{nnU-Net} & 344 & 0 & 11 & 19 & 79 & 75 & 82 & 82 & 46 & 89 & 93 & 74 & 1.7K & 0 & 57 & 76 & 97 & 53 \\
\href{https://arxiv.org/abs/2203.00131}{standard segmentation} & 344 & 0 & 18 & 33 & 82 & 84 & 82 & 88 & 56 & 88 & 86 & 85 & 1.7K & 0 & 63 & 75 & 91 & 68 \\
\textbf{R-Super (ours)} & 344 & 2.2K & \textbf{33} & \textbf{51} & \textbf{89} & \cellcolor{lightorange}\textbf{93} & 94 & 88 & \cellcolor{lightorange}\textbf{68} & \cellcolor{lightorange}\textbf{94} & 96 & 89 & 1.7K & 2.7K & \cellcolor{lightorange}\textbf{67} & 75 & 97 & 69 \\
\bottomrule
\end{tabular}
\label{tab:results_by_size}
\end{table*}

\subsection{Reports Scale the Largest Public Pancreatic Tumor Segmentation Dataset}
\label{sec:pants}

\begin{table*}[!h]
\centering
\scriptsize
\setlength{\tabcolsep}{5.4pt}
\caption{\textbf{R-Super sets a new state-of-the-art on the largest public pancreatic tumor dataset.} Combining CT-Mask pairs from PanTS (the largest public pancreatic tumor CT-Mask dataset) with CT-Report pairs from Merlin (the largest public abdominal CT-Report dataset), R-Super outperforms a segmentation model trained on CT-Mask pairs alone (PanTS) by +7\% sensitivity on small pancreatic tumors. R-Super also surpasses all alternative models on DSC and NSD, and on F1-Score and AUC in JHH-Small and Merlin-Test (in JHH-Large, all methods saturate at F1-Score 97--99 and AUC 99--100). All models were trained on the same 926 pancreatic tumor CT-Mask pairs from PanTS. Multi-task learning and R-Super additionally trained on Merlin-Train (1{,}848 pancreatic tumor and 1{,}848 no-tumor CT-Report pairs from Merlin). DSC and NSD are computed on scans with tumors and reported only where masks exist (Merlin-Test has none). Sensitivity (Se), Specificity (Sp), and F1-Score correspond to the threshold that maximizes F1-Score for each model. Statistical tests were performed to compare R-Super and standard segmentation for F1-Score and AUC (paired permutation test for F1-Score and DeLong's test for AUC); orange highlights statistically significant gains ($p<0.05$). {\PBrev Standard deviations and confidence intervals are available in Tab.~\ref{tab:cif_pants} (Appendix~\ref{app:variability}).}}

\begin{tabular}{l*{18}{c}}
\toprule
  & \multicolumn{2}{c}{\footnotesize training} & \multicolumn{6}{c}{\footnotesize JHH-Small} & \multicolumn{6}{c}{\footnotesize JHH-Large} & \multicolumn{4}{c}{\footnotesize Merlin-Test} \\
\cmidrule(lr){2-3}\cmidrule(lr){4-9}\cmidrule(lr){10-15}\cmidrule(lr){16-19}
\scriptsize train paradigm & mask & rep. & DSC & NSD & F1 & AUC & Se & Sp & DSC & NSD & F1 & AUC & Se & Sp & F1 & AUC & Se & Sp \\
\midrule
\href{https://www.nature.com/articles/s41592-020-01008-z}{nnU-Net} & 926 & 0 & 33 & 58 & 83 & 91 & 80 & 87 & 49 & 72 & 88 & 93 & 84 & 92 & 69 & 76 & 68 & 73 \\
\href{https://ieeexplore.ieee.org/document/8759483/}{Multi-task learning} & 926 & 1{,}848 & 33 & 57 & 92 & 95 & 92 & 92 & 50 & 73 & 97 & 99 & 98 & 96 & 69 & 75 & 63 & 82 \\
\href{https://arxiv.org/abs/2203.00131}{standard segmentation} & 926 & 0 & 31 & 55 & 91 & 94 & 86 & 98 & 50 & 73 & \textbf{99} & \textbf{100} & 98 & 100 & 67 & 74 & 60 & 82 \\
\textbf{R-Super (ours)} & 926 & 1{,}848 & \textbf{34} & \textbf{59} & \cellcolor{lightorange}{\textbf{95}} & \textbf{97} & 93 & 98 & \textbf{52} & \textbf{74} & 98 & \textbf{100} & 100 & 96 & \cellcolor{lightorange}{\textbf{78}} & \cellcolor{lightorange}{\textbf{83}} & 77 & 79 \\
\bottomrule
\end{tabular}
\label{tab:pants}
\end{table*}



In this experiment, we demonstrate that reports can scale the largest public pancreatic tumor dataset, PanTS \citep{li2025pants}. Reports were sourced from the public Merlin dataset \citep{blankemeier2024merlin}. Therefore, this experiment trains on only public data, ensuring reproducibility. Here, we compare R-Super against standard mask-only segmentation, nnU-Net, and multi-task learning.

\textbf{Datasets.} \textbf{(I) PanTS} \citep{li2025pants} is the largest public CT-Mask dataset for pancreatic tumors, with 9,901 CT-Mask pairs from 143 medical institutions in 17 countries, including 1,077 with pancreatic tumors, of which 926 are used for training here. 

\textbf{Training and Evaluation.} We trained R-Super first on the CT-Mask pairs from PanTS, followed by fine-tuning on CT-Mask (PanTS) plus CT-Report pairs (Merlin-Train). Evaluation was performed on Merlin-Test (internal), JHH-Large (external), and JHH-Small (external). JHH-Small is highly challenging, as small tumors are difficult to detect.

\textbf{R-Super scales the largest public pancreatic tumor segmentation dataset.} Results for the 3 test sets are shown in Tab.~\ref{tab:pants}. By leveraging reports, R-Super achieved +11\% higher F1-Score and +9\% higher AUC over training on masks only (standard segmentation), testing on Merlin-Test (internal). Even on external testing on small tumors (JHH-Small), R-Super surpassed standard segmentation by +4\% F1-Score, +3\% AUC, and +3\% DSC. As in Sec.~\ref{sec:miccai_results}, R-Super also surpassed the alternative training methods (Tab.~\ref{tab:pants}). The results in this section demonstrate that even the largest collections of CT-Mask pairs can be effectively scaled with CT-Report pairs, improving inter-hospital generalization and improving the detection and segmentation of small tumors. 



\subsection{Reports Scale the Largest Private Pancreatic Tumor Segmentation Dataset}
\label{sec:jhh_ucsf}

\begin{table}[!h]
\centering
\scriptsize
\setlength{\tabcolsep}{5.4pt}
\caption{\textbf{R-Super surpasses mask-based training when large amounts of both reports and masks are available.} On Swiss-Test, an external dataset which we publicly release, R-Super outperforms standard mask-based segmentation by +8\% sensitivity at the same specificity, for pancreatic tumor detection. {\PBrev On UCSF-Test-Non-Contrast, an internal test set of non-contrast CT scans, R-Super outperforms the standard segmentation model by +19\% sensitivity at the same specificity.} DSC and NSD are absent because there are no ground-truth tumor masks for testing. All models were trained on the same 3,488 pancreatic tumor CT-Mask pairs from JHH {\PBrev (contrast-enhanced)}; multi-task learning and R-Super additionally trained on 28,295 tumor CT-Report pairs from UCSF-Huge (41,418 CT-Report pairs{\PBrev , 4,809 non-contrast}). Sensitivity (Se), Specificity (Sp), and F1-Score correspond to the threshold that maximizes F1-Score for each model. Statistical tests compare R-Super and standard segmentation (paired permutation test for F1-Score, DeLong's test for AUC); orange highlights statistically significant gains ($p<0.05$). {\PBrev Standard deviations and confidence intervals are available in Tab.~\ref{tab:cif_jhh_ucsf} (Appendix~\ref{app:variability}).}}
\begin{tabular}{l*{6}{c}}
\toprule
  & \multicolumn{2}{c}{\footnotesize training} & \multicolumn{4}{c}{\footnotesize pancreas tumor} \\
\cmidrule(lr){2-3}\cmidrule(lr){4-7}
\scriptsize train paradigm & mask & rep. & F1 & AUC & Se & Sp \\
\midrule
\multicolumn{7}{c}{\footnotesize\textit{{\PBrev Swiss-Test}}} \\
\midrule
\href{https://www.nature.com/articles/s41592-020-01008-z}{nnU-Net} & 3,488 & 0 & 78 & 74 & 74 & 71 \\
\href{https://ieeexplore.ieee.org/document/8759483/}{Multi-task learning} & 3,488 & 28{,}295 & 74 & 76 & 64 & 85 \\
\href{https://arxiv.org/abs/2203.00131}{standard segmentation} & 3,488 & 0 & 80 & 82 & 72 & 86 \\
\textbf{R-Super (ours)} & 3,488 & 28{,}295 & \cellcolor{lightorange}{\textbf{85}} & \cellcolor{lightorange}\textbf{85} & {80} & 86 \\
\midrule
\multicolumn{7}{c}{\footnotesize\textit{{\PBrev UCSF-Test-Non-Contrast}}} \\
\midrule
\href{https://www.nature.com/articles/s41592-020-01008-z}{nnU-Net} & {\PBrev 3,488} & {\PBrev 0} & {\PBrev 52} & {\PBrev 68} & {\PBrev 67} & {\PBrev 73} \\
\href{https://ieeexplore.ieee.org/document/8759483/}{Multi-task learning} & {\PBrev 3,488} & {\PBrev 28{,}295} & {\PBrev 19} & {\PBrev 54} & {\PBrev 10} & {\PBrev 99} \\
\href{https://arxiv.org/abs/2203.00131}{standard segmentation} & {\PBrev 3,488} & {\PBrev 0} & {\PBrev 62} & {\PBrev 74} & {\PBrev 50} & {\PBrev 96} \\
\textbf{R-Super (ours)} & {\PBrev 3,488} & {\PBrev 28{,}295} & \cellcolor{lightorange}{{\PBrev \textbf{75}}} & \cellcolor{lightorange}{{\PBrev \textbf{92}}} & {{\PBrev 69}} & {\PBrev 96} \\
\bottomrule
\end{tabular}
\label{tab:jhh_ucsf}
\end{table}



In this experiment, we explore the largest scale in this paper, both in terms of masks and reports: 6,212 CT-Mask pairs and 41,418 CT-Report pairs. To reach this scale, we use private training data. However, we make our external test set public, contributing a new CT-Report dataset (Swiss-Test). Again, we compare R-Super against standard mask-only segmentation, nnU-Net, and multi-task learning.

\textbf{Datasets.} \textbf{JHH} \citep{xia2022felix} is, to the best of our knowledge, the largest private CT-Mask dataset for pancreatic tumor segmentation. It includes 6,212 CT-Mask pairs; 3,488 have pancreatic tumors, and 1,093 of those are small ($\leq$ 2 cm). \textbf{UCSF-Huge} includes 41,418 CT-Report pairs; 28,295 have pancreatic tumors, and 11,413 of those are small. \textbf{Swiss-Test} is a new dataset, sourced from the University Hospital Basel (Basel, Switzerland). This dataset includes 684 CT-Report pairs, of which 445 include pancreatic tumors. Here, we leverage it as an external test set. We publicly release this dataset. {\PBrev \textbf{UCSF-Test-Non-Contrast} includes 207 non-contrast CT-Report pairs, including 48 pancreatic tumor CT-Report pairs, 15 with only small tumors ($\leq$ 2 cm).}  More details on the datasets are available in Fig.~\ref{fig:dataset_stats}.

\textbf{R-Super scales the largest private pancreatic tumor segmentation dataset.} Tab.~\ref{tab:jhh_ucsf} shows the performance of R-Super on the Swiss-Test external dataset. R-Super surpassed standard mask-based segmentation, achieving +8\% higher pancreatic tumor detection sensitivity at matched specificity (AUC 85 vs. 82). This result demonstrates that reports can improve tumor detection even when a large CT-Mask dataset exists. Notably, the CT-Report pairs in UCSF-Huge include all pancreatic tumor cases registered in the UCSF Hospital and affiliated Californian institutions in the last 28 years. Downloading and preparing all these CT-Report pairs took around one month, at no monetary cost. In contrast, creating the 3,488 pancreatic tumor masks in JHH took 5 years, with the support of 8 radiologists, and millions of dollars in funding. This demonstrates that reports are an accessible path to scale tumor segmentation datasets and improve AI performance.

{\PBrev \textbf{R-Super detects pancreatic tumors on non-contrast CT scans.} R-Super surpassed the standard segmentation model by +19\% in sensitivity at matched specificity (AUC 92 vs. 74) for detecting pancreatic tumors on non-contrast CT scans (Tab.~\ref{tab:jhh_ucsf}). Seeing pancreatic tumors in CT scans taken without intravenous contrast is considered extremely challenging for radiologists \citep{cao2023large}. Consequently, drawing pancreatic tumor masks on non-contrast CT scans is also extremely difficult, making standard mask-supervised segmentation challenging. Previous closed-source work addressed this challenge by asking radiologists to draw a large number of masks on contrast-enhanced CT scans and registering these masks to paired non-contrast CT scans from the same patients \citep{cao2023large}. In contrast, we did not require drawing any masks for the patients with non-contrast CT. To train R-Super, we only needed to supplement an existing dataset of contrast-enhanced CT-Mask pairs (JHH) with a large-scale CT-Report dataset (UCSF-Huge), which includes 4,809 non-contrast CT-Report pairs, 3,106 with pancreatic tumors. Report supervision therefore allowed us to extend pancreatic tumor segmentation models and datasets from contrast-enhanced to non-contrast CT scans without additional masks. Performance may improve further with more non-contrast CT-Report pairs for training.}









\subsection{{\PBrev Ablation Studies}}
\label{sec:ablations}

\begin{table*}[t]
\centering
\scriptsize
\setlength{\tabcolsep}{5.2pt}
\caption{{\PBrev \textbf{Ablation studies of R-Super.} \textit{Loss functions:} combining the Volume Loss and Ball Loss achieved the best performance on UCSF-Test; on the smaller JHH-Test, the Volume Loss alone scored higher F1-Score and AUC. Training with the Volume Loss alone surpassed training with the Ball Loss alone, because the Volume Loss is easier to optimize. \textit{LLM errors (or report errors):} R-Super remained highly accurate when errors were injected into up to 10\% of the tumor reports in the training dataset. These errors can simulate either errors in the report or errors in the LLM analysis of the report. \textit{LLM size:} training with small LLMs (Qwen 3 4B or 0.6B instead of Llama 3.1 70B) substantially reduced R-Super performance. \textit{Organ mask quality:} organ masks generated by an nnU-Net trained on only 50 CT scans were sufficient to accurately train R-Super. Even when organ masks were fully removed, R-Super still outperformed the standard segmentation model (Tab.~\ref{tab:all_results}). \textit{Report information removed:} tumor diameters are important for R-Super, particularly for DSC.} {\PBrev Standard deviations and confidence intervals are available in Tab.~\ref{tab:cif_ablations} (Appendix~\ref{app:variability}). Results are reported at the operating point of maximum F1-Score for all models.}}

\label{tab:ablations}
\begin{tabular}{l*{19}{c}}
\toprule
 & \multicolumn{12}{c}{\footnotesize pancreas tumor} & \multicolumn{6}{c}{\footnotesize kidney tumor} \\
\cmidrule(lr){2-13}\cmidrule(lr){14-19}
 & \multicolumn{8}{c}{\scriptsize JHH-Test} & \multicolumn{4}{c}{\scriptsize UCSF-Test} & \multicolumn{6}{c}{\scriptsize UCSF-Test} \\
\cmidrule(lr){2-9}\cmidrule(lr){10-13}\cmidrule(lr){14-19}
\scriptsize train paradigm & mask & rep. & DSC & NSD & F1 & AUC & Se & Sp & F1 & AUC & Se & Sp & mask & rep. & F1 & AUC & Se & Sp \\
\midrule
\textbf{R-Super (ours)} & 344 & 2.2K & 59 & 69 & 91 & 92 & 94 & 88 & 83 & 90 & 83 & 89 & 1.7K & 2.7K & 75 & 78 & 80 & 70 \\
\midrule
\multicolumn{19}{l}{\textit{loss functions}} \\
Volume~Loss only & 344 & 2.2K & 59 & 61 & 94 & 97 & 90 & 98 & 76 & 89 & 80 & 82 & 1.7K & 2.7K & 75 & 74 & 86 & 63 \\
Ball Loss only & 344 & 2.2K & 59 & 59 & 83 & 92 & 78 & 90 & 71 & 82 & 64 & 90 & 1.7K & 2.7K & 74 & 75 & 87 & 58 \\
\midrule
\multicolumn{19}{l}{\textit{LLM size}} \\
small LLM (4B) & 344 & 2.2K & 63 & 74 & 82 & 84 & 76 & 92 & 70 & 80 & 74 & 79 & 1.7K & 2.7K & 72 & 76 & 89 & 47 \\
tiny LLM (0.6B) & 344 & 2.2K & 41 & 50 & 74 & 80 & 71 & 80 & 63 & 73 & 54 & 90 & 1.7K & 2.7K & 73 & 76 & 86 & 55 \\
\midrule
\multicolumn{19}{l}{\textit{report information removed}} \\
no pancreas sub-segment & 344 & 2.2K & 58 & 69 & 85 & 90 & 80 & 92 & 71 & 81 & 73 & 81 & 1.7K & 2.7K & 72 & 75 & 86 & 54 \\
no tumor size & 344 & 2.2K & 50 & 59 & 85 & 87 & 76 & 98 & 67 & 76 & 64 & 84 & 1.7K & 2.7K & 73 & 76 & 88 & 55 \\
\midrule
\multicolumn{19}{l}{\textit{organ mask importance}} \\
50 CT organ segmenter & 344 & 2.2K & 62 & 74 & 95 & 97 & 96 & 94 & 76 & 85 & 86 & 77 & 1.7K & 2.7K & 72 & 76 & 84 & 58 \\
no organ mask & 344 & 2.2K & 60 & 69 & 88 & 93 & 80 & 98 & 69 & 81 & 68 & 83 & 1.7K & 2.7K & 73 & 76 & 85 & 58 \\
\midrule
\multicolumn{19}{l}{\textit{injected LLM errors (or report errors)}} \\
+5\% errors & 344 & 2.2K & 62 & 74 & 93 & 94 & 90 & 96 & 72 & 85 & 73 & 84 & 1.7K & 2.7K & 74 & 77 & 91 & 51 \\
+10\% errors & 344 & 2.2K & 59 & 71 & 90 & 95 & 88 & 92 & 78 & 87 & 78 & 87 & 1.7K & 2.7K & 73 & 77 & 82 & 63 \\
+20\% errors & 344 & 2.2K & 61 & 73 & 89 & 90 & 86 & 94 & 71 & 82 & 76 & 79 & 1.7K & 2.7K & 72 & 76 & 86 & 53 \\
100\% errors & 344 & 2.2K & 53 & 61 & 70 & 74 & 59 & 90 & 56 & 68 & 72 & 51 & 1.7K & 2.7K & 66 & 70 & 77 & 50 \\
\bottomrule
\end{tabular}
\end{table*}

{\PBrev Tab.~\ref{tab:ablations} shows ablation studies performed with R-Super. Their aim was to understand the importance of each R-Super loss function, the resistance of R-Super to LLM errors or report errors in training, the importance of the LLM size, the importance of organ masks, the importance of the tumor information extracted from reports (diameters and sub-segments), and whether training R-Super to segment a single tumor type surpasses training a generalist R-Super to segment multiple tumor types (reported in the text only). Ablation studies were conducted with the same training and testing configuration presented in Sec.~\ref{sec:miccai_results}, with AbdomenAtlas 3.0 and UCSF-Train training sets, and UCSF-Test and JHH-Test test sets.}

{\PBrev \textbf{Training with the Volume Loss alone outperforms training with the Ball Loss alone, and training with both losses together provides the best performance on UCSF-Test.} The Ball Loss is more precise than the Volume Loss, by optimizing segmented tumors to match more details from the radiology report. However, the Ball Loss is harder to optimize when used alone. This explains why training with the Volume Loss alone surpassed training with the Ball Loss alone (Tab.~\ref{tab:ablations}). The Ball Loss identifies the most likely tumor locations based on both the radiology report and the segmentation model’s current output. It then encourages the model’s output to better match these tumor locations. Early in training, the model’s outputs are inaccurate, making the estimated tumor locations noisy and slowing training convergence. Importantly, this noise does not systematically direct the model toward incorrect segmentations. The Ball Loss may temporarily reinforce a false positive in an individual CT scan, but the same false positive is not reinforced consistently across patients. For example, similar false positives are penalized when appearing in no-tumor scans. In contrast, learning to segment real tumors (true positives) allows the segmentation model to consistently match the tumor locations, counts, and sizes described in the reports, consistently minimizing the Ball Loss. Thus, on the larger UCSF-Test, training with the Ball Loss and the Volume Loss together surpasses training with the Volume Loss alone; on the 100-scan JHH-Test, the two are within the confidence intervals of Tab.~\ref{tab:cif_ablations}. The less noisy Volume Loss stabilizes early training, while the Ball Loss refines the segmentation model’s outputs and becomes increasingly precise as the model’s predictions improve.}

{\PBrev \textbf{On the relative contribution of the Ball Loss.} We also analyze where training with the Ball Loss plus the Volume Loss (standard R-Super) has the greatest advantage over training with the Volume Loss only. To this end, we compare R-Super to the Volume-Loss-only ablation on several subsets of UCSF-Test: small tumors, multiple tumors, and complex tumors. Since UCSF-Test is larger than JHH-Test, it affords more statistical power for subset analyses. Adding the Ball Loss to the Volume Loss significantly improves overall detection for both organs: pancreatic tumors gain $+3$ AUC points (95\% CI $[+1, +6]$, $n{=}139$) and kidney tumors $+2$ ($[+1, +4]$, $n{=}169$). Two subsets gain well above this average. \textbf{Multifocal disease}: scans with $\geq 2$ kidney tumors gain $+3$ AUC points ($[+1, +4]$, $n{=}85$), above the $+2$ kidney average. For pancreas, multifocal lesions are rare ($n{=}18$), making this subset analysis not significant. \textbf{Anatomically complex pancreatic tumors}: tumors in the pancreatic tail and iso-attenuating tumors are rarer and harder to detect, and gain $+6$ AUC points ($[+1, +12]$, $n{=}27$), double the $+3$ pancreatic average. \textbf{Small tumors}: tumors $\leq 2$\,cm gain in line with the overall average, $+3$ AUC points for pancreas ($[-1, +6]$, $n{=}90$) and $+2$ for kidney ($[-1, +5]$, $n{=}77$). In short, the Ball Loss is not redundant with the Volume Loss; its improvements are largest precisely on the multifocal and complex tumors that are hard to detect.}

{\PBrev \textbf{R-Super tolerates LLM or report errors.} R-Super uses a large LLM (Llama 3.1 70B AWQ) to extract tumor information from reports accurately. A radiologist evaluated this LLM on 447 reports, and the LLM reached 96\% accuracy. To quantify how resistant R-Super is to LLM errors or report errors, we performed ablation studies injecting increasing numbers of synthetic errors into the information that the LLM extracted from reports. Tab.~\ref{tab:ablations} shows that adding synthetic errors to 5\% or 10\% of the tumor CT-Report pairs in the training dataset did not substantially change the R-Super performance. For example, it reduced tumor detection AUC by at most 5\%. With synthetic errors in 20\% of the tumor CT-Report pairs, AUC dropped by at most 8\%. Still, the performance of R-Super remained above the standard segmentation model (Tab.~\ref{tab:all_results}). With synthetic errors in 100\% of the pairs, R-Super performance dropped markedly (by up to 22\% AUC), falling below the standard segmentation model (trained on masks only) in tumor detection. This is expected, as a 100\% error rate turns report supervision essentially into training noise. The synthetic errors simulated, in equal proportions, four types of mistakes that LLMs could make when extracting tumor information from reports: tumor location errors were simulated by switching the pancreatic sub-segment for pancreatic tumors and switching the kidney laterality for kidney tumors; tumor size errors were simulated by replacing a tumor's size with the size of another tumor in the dataset; false positives were simulated by copying extracted tumor information from one report and adding it as an extra tumor to another report; and false negatives were simulated by deleting the information of an LLM-extracted tumor.}

{\PBrev \textbf{Large LLMs outperform small LLMs.} Besides injecting synthetic errors, we also trained R-Super with smaller LLMs, which should produce more real-world LLM errors. R-Super originally used a large LLM (Llama 3.1 70B AWQ). In our ablations, we replaced it with Qwen 3 4B FP8 or Qwen 3 0.6B FP8 \citep{yang2025qwen3technicalreport}. Training with these smaller LLMs substantially reduced R-Super performance. When trained with Qwen 3 4B, R-Super performed similarly on UCSF-Test to training with Llama 3.1 70B plus 20\% synthetic errors, but 6--7 points lower in F1-Score and AUC on JHH-Test. In most metrics, it slightly outperformed the standard segmentation model. With Qwen 3 0.6B, R-Super was often outperformed by the standard segmentation model (Tab.~\ref{tab:ablations}). These results suggest that current small LLMs may make too many errors in extracting tumor information from reports, making report supervision unreliable. We therefore strongly recommend large LLMs. Notably, the LLM only needs to run once per report, not representing an excessive computational cost (Llama 3.1 70B AWQ can analyze 5,000 reports in less than 1 day with a single NVIDIA RTX 6000 GPU).}

{\PBrev \textbf{R-Super can work without organ masks.} The R-Super loss functions are guided by organ masks. We dilate these masks to compensate for organ segmentation errors and for tumors growing beyond organ borders. Originally, the organ masks were generated by an nnU-Net trained on the full AbdomenAtlas 3.0 dataset of 9,262 CT scans. In our ablation, we replaced this nnU-Net with an nnU-Net trained on only 50 CT scans, randomly selected from AbdomenAtlas 3.0. This replacement did not substantially reduce R-Super's performance. For example, tumor detection AUC decreased by at most 5\% (Tab.~\ref{tab:ablations}). To evaluate the quality of these nnU-Nets in organ segmentation, we tested them on the JHH-Test dataset, which includes manually annotated pancreas and kidney masks (but not pancreas sub-segment masks). The nnU-Net trained on the full AbdomenAtlas 3.0 achieved a mean DSC of 81.8\% (SD: 12.6\%) for the pancreas and 98.0\% (SD: 1.0\%) for the kidneys. In comparison, the nnU-Net trained on only 50 CT scans achieved a mean DSC of 81.6\% (SD: 10.9\%) for the pancreas and 97.4\% (SD: 2.5\%) for the kidneys. In a more radical ablation, we fully removed the organ mask guidance from the R-Super loss functions. Without organ masks, the Volume Loss estimated the segmented tumor volume from the entire segmentation model output, and the Ball Loss searched the entire segmentation model output for the most likely tumor locations. Without organ masks, tumor detection AUC decreased by at most 9\% (Tab.~\ref{tab:ablations}), but R-Super still consistently outperformed the standard segmentation model (Tab.~\ref{tab:all_results}). In conclusion, organ masks make the R-Super loss functions more precise and improve performance. However, these organ masks can be created by organ segmentation models trained on few images, and even without organ masks, the R-Super loss functions can still improve performance.}

{\PBrev \textbf{Tumor diameters are a strong training signal.} To assess the contribution of different information in radiology reports to the R-Super performance, we independently removed tumor diameters or tumor sub-segments from all training reports. Removing tumor sub-segments decreased tumor detection AUC by up to 9\% and DSC by 1\%. In contrast, removing tumor diameters decreased AUC by up to 14\% and DSC by 9\% (Tab.~\ref{tab:ablations}). Thus, tumor diameters and sub-segments are useful, but tumor diameters provide stronger supervision to segmentation.}

{\PBrev \textbf{Training R-Super to segment a single tumor type performs slightly better than training it to segment two tumor types.} Training one R-Super model for pancreatic tumors only (rather than a single model for both pancreatic and kidney tumors) changed pancreatic tumor performance to DSC/NSD/Se/Sp/F1/AUC of 62/64/88/96/92/98 on JHH-Test (from 59/69/94/88/91/92 for the two-tumor model, Tab.~\ref{tab:ablations}) and Se/Sp/F1/AUC of 86/86/83/91 on UCSF-Test (from 83/89/83/90), but it requires training one model per tumor type, which scales poorly as more tumor types are added.}




\section{Conclusion}

R-Super transforms radiology reports into supervision for tumor segmentation, allowing AI performance to scale with the vast number of CT-Report pairs already stored in hospitals and increasingly available in public datasets \citep{hamamci2024ct2rep,bassi2025radgpt,blankemeier2024merlin,bassi2026merlin,li2026radthinking,li2026cancerverse}. This work demonstrates that reports substantially improve tumor detection and segmentation across multiple data scales, from adding over 6,000 CT-Report pairs to a small set of 200 CT-Mask pairs (100 with tumors), up to adding over 40,000 CT-Report pairs (decades of pancreatic cancer cases) to over 6,000 CT-Mask pairs (a dataset that took 8 radiologists 5 years to create). 

For reproducibility, we include not only experiments with private data, but also with public data. Furthermore, we release a new public CT-Report test set (Swiss-Test). Experiments at all data scales include validation on hospitals never seen during training, assessing generalization to out-of-distribution data. Notably, reports enable easier merging of data across hospitals, without the need to create masks for each hospital. This improves training data diversity, improving generalization, as shown by consistent gains on multiple external test sets (JHH-Test, JHH-Small, JHH-Large, and Swiss-Test). 

R-Super also surpassed alternative methods of learning from reports. For instance, R-Super outperformed CLIP-style pretraining and multi-task learning, trained on the same datasets. This is possible because R-Super uses reports to supervise tumor segmentation directly, rather than using reports to supervise an auxiliary task. To minimize the R-Super losses, the segmentation model learns to segment tumors that match tumor descriptions in reports. 

The scales reached here are already substantial. For instance, our largest training set (Sec.~\ref{sec:jhh_ucsf}) contains about 30 times more pancreatic tumor cases than the largest public pancreatic tumor segmentation dataset. Yet new public datasets and multi-institution collaborations can extend this scale much further. In other fields, such as natural language processing and computer vision, major performance gains came from learning at scale from readily available text. This study suggests that the same path is possible for tumor detection and segmentation.

\section*{Acknowledgments}

This work was supported by the Lustgarten Foundation for Pancreatic Cancer Research and the National Institutes of Health (NIH) under Award Numbers R01EB037669 and R01EB039836. We would like to thank the Johns Hopkins Research IT team in \href{https://researchit.jhu.edu/}{IT@JH} for their support and infrastructure resources where some of these analyses were conducted; especially \href{https://researchit.jhu.edu/research-hpc/}{DISCOVERY HPC}. 










\clearpage
\bibliographystyle{cas-model2-names}

\bibliography{cas-refs,zzhou}

\begin{thebibliography}{51}
\expandafter\ifx\csname natexlab\endcsname\relax\def\natexlab#1{#1}\fi
\providecommand{\url}[1]{\texttt{#1}}
\providecommand{\href}[2]{#2}
\providecommand{\path}[1]{#1}
\providecommand{\DOIprefix}{doi:}
\providecommand{\ArXivprefix}{arXiv:}
\providecommand{\URLprefix}{URL: }
\providecommand{\Pubmedprefix}{pmid:}
\providecommand{\doi}[1]{\href{http://dx.doi.org/#1}{\path{#1}}}
\providecommand{\Pubmed}[1]{\href{pmid:#1}{\path{#1}}}
\providecommand{\bibinfo}[2]{#2}
\ifx\xfnm\relax \def\xfnm[#1]{\unskip,\space#1}\fi
\bibitem[{Antonelli et~al.(2021)Antonelli, Reinke, Bakas, Farahani, Landman, Litjens, Menze, Ronneberger, Summers, van Ginneken et~al.}]{antonelli2021medical}
\bibinfo{author}{Antonelli, M.}, \bibinfo{author}{Reinke, A.}, \bibinfo{author}{Bakas, S.}, \bibinfo{author}{Farahani, K.}, \bibinfo{author}{Landman, B.A.}, \bibinfo{author}{Litjens, G.}, \bibinfo{author}{Menze, B.}, \bibinfo{author}{Ronneberger, O.}, \bibinfo{author}{Summers, R.M.}, \bibinfo{author}{van Ginneken, B.}, et~al., \bibinfo{year}{2021}.
\newblock \bibinfo{title}{The {Medical Segmentation Decathlon}}.
\newblock \bibinfo{journal}{arXiv preprint arXiv:2106.05735} .
\bibitem[{Bassi et~al.(2024a)Bassi, Dertkigil and Cavalli}]{bassi2024improving}
\bibinfo{author}{Bassi, P.R.}, \bibinfo{author}{Dertkigil, S.S.}, \bibinfo{author}{Cavalli, A.}, \bibinfo{year}{2024}a.
\newblock \bibinfo{title}{Improving deep neural network generalization and robustness to background bias via layer-wise relevance propagation optimization}.
\newblock \bibinfo{journal}{Nature Communications} \bibinfo{volume}{15}, \bibinfo{pages}{291}.
\bibitem[{Bassi et~al.(2025a)Bassi, Li, Chen, Zhu, Lin, Decherchi, Cavalli, Wang, Yang, Yuille and Zhou}]{bassi2025learning}
\bibinfo{author}{Bassi, P.R.}, \bibinfo{author}{Li, W.}, \bibinfo{author}{Chen, J.}, \bibinfo{author}{Zhu, Z.}, \bibinfo{author}{Lin, T.}, \bibinfo{author}{Decherchi, S.}, \bibinfo{author}{Cavalli, A.}, \bibinfo{author}{Wang, K.}, \bibinfo{author}{Yang, Y.}, \bibinfo{author}{Yuille, A.L.}, \bibinfo{author}{Zhou, Z.}, \bibinfo{year}{2025}a.
\newblock \bibinfo{title}{Learning segmentation from radiology reports}, in: \bibinfo{booktitle}{International Conference on Medical Image Computing and Computer-Assisted Intervention (MICCAI)}, \bibinfo{organization}{Springer}. pp. \bibinfo{pages}{305--315}.
\newblock \URLprefix \url{https://github.com/MrGiovanni/R-Super}.
\bibitem[{Bassi et~al.(2024b)Bassi, Li, Tang, Isensee, Wang, Chen, Chou, Kirchhoff, Rokuss, Huang, Ye, He, Wald, Ulrich, Baumgartner, Roy, Maier-Hein, Jaeger, Ye, Xie, Zhang, Chen, Xia, Xing, Zhu, Sadegheih, Bozorgpour, Kumari, Azad, Merhof, Shi, Ma, Du, Bai, Huang, Zhao, Wang, Li, Gu, Dong, Yang, Mazurowski, Gupta, Wu, Zhuang, Chen, Roth, Xu, Blaschko, Decherchi, Cavalli, Yuille and Zhou}]{bassi2024touchstone}
\bibinfo{author}{Bassi, P.R.}, \bibinfo{author}{Li, W.}, \bibinfo{author}{Tang, Y.}, \bibinfo{author}{Isensee, F.}, \bibinfo{author}{Wang, Z.}, \bibinfo{author}{Chen, J.}, \bibinfo{author}{Chou, Y.C.}, \bibinfo{author}{Kirchhoff, Y.}, \bibinfo{author}{Rokuss, M.}, \bibinfo{author}{Huang, Z.}, \bibinfo{author}{Ye, J.}, \bibinfo{author}{He, J.}, \bibinfo{author}{Wald, T.}, \bibinfo{author}{Ulrich, C.}, \bibinfo{author}{Baumgartner, M.}, \bibinfo{author}{Roy, S.}, \bibinfo{author}{Maier-Hein, K.H.}, \bibinfo{author}{Jaeger, P.}, \bibinfo{author}{Ye, Y.}, \bibinfo{author}{Xie, Y.}, \bibinfo{author}{Zhang, J.}, \bibinfo{author}{Chen, Z.}, \bibinfo{author}{Xia, Y.}, \bibinfo{author}{Xing, Z.}, \bibinfo{author}{Zhu, L.}, \bibinfo{author}{Sadegheih, Y.}, \bibinfo{author}{Bozorgpour, A.}, \bibinfo{author}{Kumari, P.}, \bibinfo{author}{Azad, R.}, \bibinfo{author}{Merhof, D.}, \bibinfo{author}{Shi, P.}, \bibinfo{author}{Ma, T.}, \bibinfo{author}{Du, Y.}, \bibinfo{author}{Bai, F.}, \bibinfo{author}{Huang, T.},
  \bibinfo{author}{Zhao, B.}, \bibinfo{author}{Wang, H.}, \bibinfo{author}{Li, X.}, \bibinfo{author}{Gu, H.}, \bibinfo{author}{Dong, H.}, \bibinfo{author}{Yang, J.}, \bibinfo{author}{Mazurowski, M.A.}, \bibinfo{author}{Gupta, S.}, \bibinfo{author}{Wu, L.}, \bibinfo{author}{Zhuang, J.}, \bibinfo{author}{Chen, H.}, \bibinfo{author}{Roth, H.}, \bibinfo{author}{Xu, D.}, \bibinfo{author}{Blaschko, M.B.}, \bibinfo{author}{Decherchi, S.}, \bibinfo{author}{Cavalli, A.}, \bibinfo{author}{Yuille, A.L.}, \bibinfo{author}{Zhou, Z.}, \bibinfo{year}{2024}b.
\newblock \bibinfo{title}{Touchstone benchmark: Are we on the right way for evaluating {AI} algorithms for medical segmentation?}
\newblock \bibinfo{journal}{Advances in Neural Information Processing Systems (NeurIPS) Datasets and Benchmarks Track} \bibinfo{volume}{37}, \bibinfo{pages}{15184--15201}.
\newblock \URLprefix \url{https://github.com/MrGiovanni/Touchstone}.
\bibitem[{Bassi et~al.(2025b)Bassi, Yavuz, Hamamci, Er, Chen, Li, Menze, Decherchi, Cavalli, Wang, Yang, Yuille and Zhou}]{bassi2025radgpt}
\bibinfo{author}{Bassi, P.R.}, \bibinfo{author}{Yavuz, M.C.}, \bibinfo{author}{Hamamci, I.E.}, \bibinfo{author}{Er, S.}, \bibinfo{author}{Chen, X.}, \bibinfo{author}{Li, W.}, \bibinfo{author}{Menze, B.}, \bibinfo{author}{Decherchi, S.}, \bibinfo{author}{Cavalli, A.}, \bibinfo{author}{Wang, K.}, \bibinfo{author}{Yang, Y.}, \bibinfo{author}{Yuille, A.}, \bibinfo{author}{Zhou, Z.}, \bibinfo{year}{2025}b.
\newblock \bibinfo{title}{{RadGPT}: Constructing {3D} image-text tumor datasets}, in: \bibinfo{booktitle}{Proceedings of the IEEE/CVF International Conference on Computer Vision}, pp. \bibinfo{pages}{23720--23730}.
\newblock \URLprefix \url{https://github.com/MrGiovanni/RadGPT}.
\bibitem[{Bassi et~al.(2025c)Bassi, Zhou, Li, P{\l}otka, Chen, Chen, Zhu, Prz{\k{a}}do, Hamamci, Er, Chen, Yavuz, Chou, Lin, Wang, Tang, Cwikla, Decherchi, Cavalli, Yang, Yuille and Zhou}]{bassi2025scaling}
\bibinfo{author}{Bassi, P.R.}, \bibinfo{author}{Zhou, X.}, \bibinfo{author}{Li, W.}, \bibinfo{author}{P{\l}otka, S.}, \bibinfo{author}{Chen, J.}, \bibinfo{author}{Chen, Q.}, \bibinfo{author}{Zhu, Z.}, \bibinfo{author}{Prz{\k{a}}do, J.}, \bibinfo{author}{Hamamci, I.E.}, \bibinfo{author}{Er, S.}, \bibinfo{author}{Chen, X.}, \bibinfo{author}{Yavuz, M.C.}, \bibinfo{author}{Chou, Y.C.}, \bibinfo{author}{Lin, T.}, \bibinfo{author}{Wang, K.}, \bibinfo{author}{Tang, Y.}, \bibinfo{author}{Cwikla, J.B.}, \bibinfo{author}{Decherchi, S.}, \bibinfo{author}{Cavalli, A.}, \bibinfo{author}{Yang, Y.}, \bibinfo{author}{Yuille, A.L.}, \bibinfo{author}{Zhou, Z.}, \bibinfo{year}{2025}c.
\newblock \bibinfo{title}{Scaling artificial intelligence for multi-tumor early detection with more reports, fewer masks}.
\newblock \bibinfo{journal}{arXiv preprint arXiv:2510.14803} \URLprefix \url{https://github.com/MrGiovanni/R-Super}.
\bibitem[{Bassi et~al.(2026a)Bassi, Li, Gu, Chen, Zhou, Zhu, Er, Hamamci, Menze, Akan, Wang, Yang, Yuille and Zhou}]{bassi2026rtsuper}
\bibinfo{author}{Bassi, P.R.A.S.}, \bibinfo{author}{Li, W.}, \bibinfo{author}{Gu, H.}, \bibinfo{author}{Chen, J.}, \bibinfo{author}{Zhou, X.}, \bibinfo{author}{Zhu, Z.}, \bibinfo{author}{Er, S.}, \bibinfo{author}{Hamamci, I.E.}, \bibinfo{author}{Menze, B.}, \bibinfo{author}{Akan, G.E.}, \bibinfo{author}{Wang, K.}, \bibinfo{author}{Yang, Y.}, \bibinfo{author}{Yuille, A.L.}, \bibinfo{author}{Zhou, Z.}, \bibinfo{year}{2026}a.
\newblock \bibinfo{title}{{RT-Super}: Learning tumor segmentation from future reports}, in: \bibinfo{booktitle}{International Conference on Medical Image Computing and Computer-Assisted Intervention (MICCAI)}.
\bibitem[{Bassi et~al.(2026b)Bassi, Li, P{\l}otka, Honjo, Prz{\k{a}}do, Zhou, Wang, Yang, Jensen, Chaudhari, Langlotz, Yuille and Zhou}]{bassi2026merlin}
\bibinfo{author}{Bassi, P.R.A.S.}, \bibinfo{author}{Li, W.}, \bibinfo{author}{P{\l}otka, S.}, \bibinfo{author}{Honjo, R.}, \bibinfo{author}{Prz{\k{a}}do, J.}, \bibinfo{author}{Zhou, X.}, \bibinfo{author}{Wang, K.}, \bibinfo{author}{Yang, Y.}, \bibinfo{author}{Jensen, M.}, \bibinfo{author}{Chaudhari, A.}, \bibinfo{author}{Langlotz, C.}, \bibinfo{author}{Yuille, A.L.}, \bibinfo{author}{Zhou, Z.}, \bibinfo{year}{2026}b.
\newblock \bibinfo{title}{{Merlin Plus}: A large-scale, multi-cancer, image-mask-report dataset}, in: \bibinfo{booktitle}{International Conference on Medical Image Computing and Computer-Assisted Intervention (MICCAI)}.
\bibitem[{Blankemeier et~al.(2026)Blankemeier, Kumar, Cohen, Van~Veen, Gardezi, Paschali, Chen, Delbrouck, Reis, Truyts et~al.}]{blankemeier2024merlin}
\bibinfo{author}{Blankemeier, L.}, \bibinfo{author}{Kumar, A.}, \bibinfo{author}{Cohen, J.P.}, \bibinfo{author}{Van~Veen, D.}, \bibinfo{author}{Gardezi, S.J.S.}, \bibinfo{author}{Paschali, M.}, \bibinfo{author}{Chen, Z.}, \bibinfo{author}{Delbrouck, J.B.}, \bibinfo{author}{Reis, E.}, \bibinfo{author}{Truyts, C.}, et~al., \bibinfo{year}{2026}.
\newblock \bibinfo{title}{Merlin: a computed tomography vision--language foundation model and dataset}.
\newblock \bibinfo{journal}{Nature} \bibinfo{volume}{652}, \bibinfo{pages}{1318--1328}.
\newblock \DOIprefix\doi{10.1038/s41586-026-10181-8}.
\bibitem[{Bosma et~al.(2023)Bosma, Saha, Hosseinzadeh, Slootweg, de~Rooij and Huisman}]{bosma2023semisupervised}
\bibinfo{author}{Bosma, J.S.}, \bibinfo{author}{Saha, A.}, \bibinfo{author}{Hosseinzadeh, M.}, \bibinfo{author}{Slootweg, I.}, \bibinfo{author}{de~Rooij, M.}, \bibinfo{author}{Huisman, H.}, \bibinfo{year}{2023}.
\newblock \bibinfo{title}{Semisupervised learning with report-guided pseudo labels for deep learning--based prostate cancer detection using biparametric mri}.
\newblock \bibinfo{journal}{Radiology: Artificial Intelligence} \bibinfo{volume}{5}, \bibinfo{pages}{e230031}.
\bibitem[{Can et~al.(2018)Can, Chaitanya, Mustafa, Koch, Konukoglu and Baumgartner}]{can_scrible}
\bibinfo{author}{Can, Y.B.}, \bibinfo{author}{Chaitanya, K.}, \bibinfo{author}{Mustafa, B.}, \bibinfo{author}{Koch, L.M.}, \bibinfo{author}{Konukoglu, E.}, \bibinfo{author}{Baumgartner, C.F.}, \bibinfo{year}{2018}.
\newblock \bibinfo{title}{Learning to segment medical images with scribble-supervision alone}, in: \bibinfo{editor}{Stoyanov, D.}, \bibinfo{editor}{Taylor, Z.}, \bibinfo{editor}{Carneiro, G.}, \bibinfo{editor}{Syeda-Mahmood, T.}, \bibinfo{editor}{Martel, A.}, \bibinfo{editor}{Maier-Hein, L.}, \bibinfo{editor}{Tavares, J.M.R.}, \bibinfo{editor}{Bradley, A.}, \bibinfo{editor}{Papa, J.P.}, \bibinfo{editor}{Belagiannis, V.}, \bibinfo{editor}{Nascimento, J.C.}, \bibinfo{editor}{Lu, Z.}, \bibinfo{editor}{Conjeti, S.}, \bibinfo{editor}{Moradi, M.}, \bibinfo{editor}{Greenspan, H.}, \bibinfo{editor}{Madabhushi, A.} (Eds.), \bibinfo{booktitle}{Deep Learning in Medical Image Analysis and Multimodal Learning for Clinical Decision Support}, \bibinfo{publisher}{Springer International Publishing}, \bibinfo{address}{Cham}. pp. \bibinfo{pages}{236--244}.
\bibitem[{Cao et~al.(2023)Cao, Xia, Yao, Han, Lambert, Zhang, Tang, Jin, Jiang, Fang et~al.}]{cao2023large}
\bibinfo{author}{Cao, K.}, \bibinfo{author}{Xia, Y.}, \bibinfo{author}{Yao, J.}, \bibinfo{author}{Han, X.}, \bibinfo{author}{Lambert, L.}, \bibinfo{author}{Zhang, T.}, \bibinfo{author}{Tang, W.}, \bibinfo{author}{Jin, G.}, \bibinfo{author}{Jiang, H.}, \bibinfo{author}{Fang, X.}, et~al., \bibinfo{year}{2023}.
\newblock \bibinfo{title}{Large-scale pancreatic cancer detection via non-contrast {CT} and deep learning}.
\newblock \bibinfo{journal}{Nature medicine} \bibinfo{volume}{29}, \bibinfo{pages}{3033--3043}.
\bibitem[{Chaitanya et~al.(2020)Chaitanya, Erdil, Karani and Konukoglu}]{Chaitanya_Contrastive}
\bibinfo{author}{Chaitanya, K.}, \bibinfo{author}{Erdil, E.}, \bibinfo{author}{Karani, N.}, \bibinfo{author}{Konukoglu, E.}, \bibinfo{year}{2020}.
\newblock \bibinfo{title}{Contrastive learning of global and local features for medical image segmentation with limited annotations}, in: \bibinfo{editor}{Larochelle, H.}, \bibinfo{editor}{Ranzato, M.}, \bibinfo{editor}{Hadsell, R.}, \bibinfo{editor}{Balcan, M.}, \bibinfo{editor}{Lin, H.} (Eds.), \bibinfo{booktitle}{Advances in Neural Information Processing Systems}, \bibinfo{publisher}{Curran Associates, Inc.}. pp. \bibinfo{pages}{12546--12558}.
\newblock \URLprefix \url{https://proceedings.neurips.cc/paper_files/paper/2020/file/949686ecef4ee20a62d16b4a2d7ccca3-Paper.pdf}.
\bibitem[{Chen et~al.(2019)Chen, Dong, Li, Jiang, Rong and Wu}]{chen2019lesion}
\bibinfo{author}{Chen, E.Z.}, \bibinfo{author}{Dong, X.}, \bibinfo{author}{Li, X.}, \bibinfo{author}{Jiang, H.}, \bibinfo{author}{Rong, R.}, \bibinfo{author}{Wu, J.}, \bibinfo{year}{2019}.
\newblock \bibinfo{title}{Lesion attributes segmentation for melanoma detection with multi-task {U-Net}}, in: \bibinfo{booktitle}{2019 IEEE 16th International Symposium on Biomedical Imaging (ISBI 2019)}, \bibinfo{organization}{IEEE}. pp. \bibinfo{pages}{485--488}.
\bibitem[{Chen et~al.(2025)Chen, Zhou, Liu, Chen, Li, Jiang, Huang, Zhao, Yu, He, Zheng, Shao, Yuille and Zhou}]{chen2025scaling}
\bibinfo{author}{Chen, Q.}, \bibinfo{author}{Zhou, X.}, \bibinfo{author}{Liu, C.}, \bibinfo{author}{Chen, H.}, \bibinfo{author}{Li, W.}, \bibinfo{author}{Jiang, Z.}, \bibinfo{author}{Huang, Z.}, \bibinfo{author}{Zhao, Y.}, \bibinfo{author}{Yu, D.}, \bibinfo{author}{He, J.}, \bibinfo{author}{Zheng, Y.}, \bibinfo{author}{Shao, L.}, \bibinfo{author}{Yuille, A.}, \bibinfo{author}{Zhou, Z.}, \bibinfo{year}{2025}.
\newblock \bibinfo{title}{Scaling tumor segmentation: Best lessons from real and synthetic data}, in: \bibinfo{booktitle}{Proceedings of the IEEE International Conference on Computer Vision (ICCV)}, pp. \bibinfo{pages}{24001--24013}.
\newblock \URLprefix \url{https://github.com/BodyMaps/AbdomenAtlas2.0}.
\bibitem[{Chou et~al.(2024)Chou, Li, Fan, Yuille and Zhou}]{chou2024acquiring}
\bibinfo{author}{Chou, Y.C.}, \bibinfo{author}{Li, B.}, \bibinfo{author}{Fan, D.P.}, \bibinfo{author}{Yuille, A.}, \bibinfo{author}{Zhou, Z.}, \bibinfo{year}{2024}.
\newblock \bibinfo{title}{Acquiring weak annotations for tumor localization in temporal and volumetric data}.
\newblock \bibinfo{journal}{Machine Intelligence Research} , \bibinfo{pages}{1--13}\URLprefix \url{https://github.com/johnson111788/Drag-Drop}.
\bibitem[{Dubey et~al.(2024)Dubey, Jauhri, Pandey, Kadian, Al-Dahle, Letman, Mathur, Schelten, Yang, Fan et~al.}]{dubey2024llama}
\bibinfo{author}{Dubey, A.}, \bibinfo{author}{Jauhri, A.}, \bibinfo{author}{Pandey, A.}, \bibinfo{author}{Kadian, A.}, \bibinfo{author}{Al-Dahle, A.}, \bibinfo{author}{Letman, A.}, \bibinfo{author}{Mathur, A.}, \bibinfo{author}{Schelten, A.}, \bibinfo{author}{Yang, A.}, \bibinfo{author}{Fan, A.}, et~al., \bibinfo{year}{2024}.
\newblock \bibinfo{title}{The {Llama} 3 herd of models}.
\newblock \bibinfo{journal}{arXiv preprint arXiv:2407.21783} .
\bibitem[{Eisenhauer et~al.(2009)Eisenhauer, Therasse, Bogaerts, Schwartz, Sargent, Ford, Dancey, Arbuck, Gwyther, Mooney et~al.}]{eisenhauer2009new}
\bibinfo{author}{Eisenhauer, E.A.}, \bibinfo{author}{Therasse, P.}, \bibinfo{author}{Bogaerts, J.}, \bibinfo{author}{Schwartz, L.H.}, \bibinfo{author}{Sargent, D.}, \bibinfo{author}{Ford, R.}, \bibinfo{author}{Dancey, J.}, \bibinfo{author}{Arbuck, S.}, \bibinfo{author}{Gwyther, S.}, \bibinfo{author}{Mooney, M.}, et~al., \bibinfo{year}{2009}.
\newblock \bibinfo{title}{New response evaluation criteria in solid tumours: revised {RECIST} guideline (version 1.1)}.
\newblock \bibinfo{journal}{European journal of cancer} \bibinfo{volume}{45}, \bibinfo{pages}{228--247}.
\bibitem[{Gao et~al.(2022)Gao, Zhou, Liu, Yan, Zhang and Metaxas}]{gao2022data}
\bibinfo{author}{Gao, Y.}, \bibinfo{author}{Zhou, M.}, \bibinfo{author}{Liu, D.}, \bibinfo{author}{Yan, Z.}, \bibinfo{author}{Zhang, S.}, \bibinfo{author}{Metaxas, D.N.}, \bibinfo{year}{2022}.
\newblock \bibinfo{title}{A data-scalable transformer for medical image segmentation: architecture, model efficiency, and benchmark}.
\newblock \bibinfo{journal}{arXiv preprint arXiv:2203.00131} .
\bibitem[{Gobara et~al.(2019)Gobara, Yoshizako, Yoshida, Nakamura, Shiina and Kitagaki}]{gobara2019t1a}
\bibinfo{author}{Gobara, A.}, \bibinfo{author}{Yoshizako, T.}, \bibinfo{author}{Yoshida, R.}, \bibinfo{author}{Nakamura, M.}, \bibinfo{author}{Shiina, H.}, \bibinfo{author}{Kitagaki, H.}, \bibinfo{year}{2019}.
\newblock \bibinfo{title}{{T1a} renal cell carcinoma on unenhanced {CT}: Analysis of detectability and imaging features}.
\newblock \bibinfo{journal}{Acta Radiologica Open} \bibinfo{volume}{8}, \bibinfo{pages}{2058460119849706}.
\newblock \DOIprefix\doi{10.1177/2058460119849706}.
\bibitem[{Haghighi et~al.(2020)Haghighi, Hosseinzadeh~Taher, Zhou, Gotway and Liang}]{haghighi2020learning}
\bibinfo{author}{Haghighi, F.}, \bibinfo{author}{Hosseinzadeh~Taher, M.R.}, \bibinfo{author}{Zhou, Z.}, \bibinfo{author}{Gotway, M.B.}, \bibinfo{author}{Liang, J.}, \bibinfo{year}{2020}.
\newblock \bibinfo{title}{Learning semantics-enriched representation via self-discovery, self-classification, and self-restoration}, in: \bibinfo{booktitle}{International Conference on Medical Image Computing and Computer-Assisted Intervention}, \bibinfo{organization}{Springer}. pp. \bibinfo{pages}{137--147}.
\newblock \URLprefix \url{https://github.com/fhaghighi/SemanticGenesis}.
\bibitem[{Hamamci et~al.(2024)Hamamci, Er and Menze}]{hamamci2024ct2rep}
\bibinfo{author}{Hamamci, I.E.}, \bibinfo{author}{Er, S.}, \bibinfo{author}{Menze, B.}, \bibinfo{year}{2024}.
\newblock \bibinfo{title}{{CT2Rep}: Automated radiology report generation for {3D} medical imaging}, in: \bibinfo{booktitle}{International Conference on Medical Image Computing and Computer-Assisted Intervention}, \bibinfo{organization}{Springer}. pp. \bibinfo{pages}{476--486}.
\bibitem[{Hoogenboom et~al.(2022)Hoogenboom, Engels, Chuprin, van Hooft, LeGout, Wallace and Bolan}]{hoogenboom2022prevalence}
\bibinfo{author}{Hoogenboom, S.A.}, \bibinfo{author}{Engels, M.M.}, \bibinfo{author}{Chuprin, A.V.}, \bibinfo{author}{van Hooft, J.E.}, \bibinfo{author}{LeGout, J.D.}, \bibinfo{author}{Wallace, M.B.}, \bibinfo{author}{Bolan, C.W.}, \bibinfo{year}{2022}.
\newblock \bibinfo{title}{Prevalence, features, and explanations of missed and misinterpreted pancreatic cancer on imaging: a matched case--control study}.
\newblock \bibinfo{journal}{Abdominal Radiology} \bibinfo{volume}{47}, \bibinfo{pages}{4160--4172}.
\bibitem[{Hooper et~al.(2023)Hooper, Chen, Saab, Bhatia, Langlotz and R{\'e}}]{hooper2023case}
\bibinfo{author}{Hooper, S.}, \bibinfo{author}{Chen, M.}, \bibinfo{author}{Saab, K.}, \bibinfo{author}{Bhatia, K.}, \bibinfo{author}{Langlotz, C.}, \bibinfo{author}{R{\'e}, C.}, \bibinfo{year}{2023}.
\newblock \bibinfo{title}{A case for reframing automated medical image classification as segmentation}.
\newblock \bibinfo{journal}{Advances in Neural Information Processing Systems} \bibinfo{volume}{36}, \bibinfo{pages}{55415--55441}.
\bibitem[{Isensee et~al.(2021)Isensee, Jaeger, Kohl, Petersen and Maier-Hein}]{isensee2021nnu}
\bibinfo{author}{Isensee, F.}, \bibinfo{author}{Jaeger, P.F.}, \bibinfo{author}{Kohl, S.A.}, \bibinfo{author}{Petersen, J.}, \bibinfo{author}{Maier-Hein, K.H.}, \bibinfo{year}{2021}.
\newblock \bibinfo{title}{{nnU-Net}: a self-configuring method for deep learning-based biomedical image segmentation}.
\newblock \bibinfo{journal}{Nature Methods} \bibinfo{volume}{18}, \bibinfo{pages}{203--211}.
\bibitem[{Kolesnikov and Lampert(2016)}]{kolesnikov2016seed}
\bibinfo{author}{Kolesnikov, A.}, \bibinfo{author}{Lampert, C.H.}, \bibinfo{year}{2016}.
\newblock \bibinfo{title}{Seed, expand and constrain: Three principles for weakly-supervised image segmentation}, in: \bibinfo{booktitle}{European conference on computer vision}, \bibinfo{organization}{Springer}. pp. \bibinfo{pages}{695--711}.
\bibitem[{Landman et~al.(2017)Landman, Xu, Eugenio~Igelsias, Styner, Robin~Langerak and Klein}]{landman2017multiatlas}
\bibinfo{author}{Landman, B.}, \bibinfo{author}{Xu, Z.}, \bibinfo{author}{Eugenio~Igelsias, J.}, \bibinfo{author}{Styner, M.}, \bibinfo{author}{Robin~Langerak, T.}, \bibinfo{author}{Klein, A.}, \bibinfo{year}{2017}.
\newblock \bibinfo{title}{Multi-atlas labeling beyond the cranial vault-workshop and challenge}, in: \bibinfo{booktitle}{MICCAI Multi-Atlas Labeling Beyond the Cranial Vault---Workshop and Challenge}.
\bibitem[{Li et~al.(2025a)Li, Bassi, Lin, Chou, Zhou, Tang, Isensee, Wang, Chen, Xu, Ye, Zhu, Decherchi, Cavalli, Yuille and Zhou}]{li2025scalemai}
\bibinfo{author}{Li, W.}, \bibinfo{author}{Bassi, P.R.}, \bibinfo{author}{Lin, T.}, \bibinfo{author}{Chou, Y.C.}, \bibinfo{author}{Zhou, X.}, \bibinfo{author}{Tang, Y.}, \bibinfo{author}{Isensee, F.}, \bibinfo{author}{Wang, K.}, \bibinfo{author}{Chen, Q.}, \bibinfo{author}{Xu, X.}, \bibinfo{author}{Ye, J.}, \bibinfo{author}{Zhu, Z.}, \bibinfo{author}{Decherchi, S.}, \bibinfo{author}{Cavalli, A.}, \bibinfo{author}{Yuille, A.L.}, \bibinfo{author}{Zhou, Z.}, \bibinfo{year}{2025}a.
\newblock \bibinfo{title}{{ScaleMAI}: Accelerating the development of trusted datasets and {AI} models}.
\newblock \bibinfo{journal}{arXiv preprint arXiv:2501.03410} \URLprefix \url{https://github.com/MrGiovanni/ScaleMAI}.
\bibitem[{Li et~al.(2026a)Li, Bassi, Zhou, Wasserthal, Yuille and Zhou}]{li2026radthinking}
\bibinfo{author}{Li, W.}, \bibinfo{author}{Bassi, P.R.}, \bibinfo{author}{Zhou, X.}, \bibinfo{author}{Wasserthal, J.}, \bibinfo{author}{Yuille, A.L.}, \bibinfo{author}{Zhou, Z.}, \bibinfo{year}{2026}a.
\newblock \bibinfo{title}{{RadThinking}: A dataset for longitudinal clinical reasoning in radiology}.
\newblock \bibinfo{journal}{arXiv preprint arXiv:2605.10761} .
\bibitem[{Li et~al.(2026b)Li, Bassi, Wu, Zhou, Zhao, Chen, Plotka, Lin, Zhu, Martin, Caskey, Jiang, Chen, \'{C}wik{\l}a, Sankowski, Wu, Decherchi, Cavalli, Lall, Tomasetti, Guo, Yu, Cai, Qiao, Bao, Hu, Wang, Sitek, Ding, Li, Wang, Yu, Zhang, Yang, Wang, Yuille and Zhou}]{li2026early}
\bibinfo{author}{Li, W.}, \bibinfo{author}{Bassi, P.R.A.S.}, \bibinfo{author}{Wu, L.}, \bibinfo{author}{Zhou, X.}, \bibinfo{author}{Zhao, Y.}, \bibinfo{author}{Chen, Q.}, \bibinfo{author}{Plotka, S.}, \bibinfo{author}{Lin, T.}, \bibinfo{author}{Zhu, Z.}, \bibinfo{author}{Martin, M.}, \bibinfo{author}{Caskey, J.}, \bibinfo{author}{Jiang, S.}, \bibinfo{author}{Chen, X.}, \bibinfo{author}{\'{C}wik{\l}a, J.B.}, \bibinfo{author}{Sankowski, A.}, \bibinfo{author}{Wu, Y.}, \bibinfo{author}{Decherchi, S.}, \bibinfo{author}{Cavalli, A.}, \bibinfo{author}{Lall, C.}, \bibinfo{author}{Tomasetti, C.}, \bibinfo{author}{Guo, Y.}, \bibinfo{author}{Yu, X.}, \bibinfo{author}{Cai, Y.}, \bibinfo{author}{Qiao, H.}, \bibinfo{author}{Bao, J.}, \bibinfo{author}{Hu, C.}, \bibinfo{author}{Wang, X.}, \bibinfo{author}{Sitek, A.}, \bibinfo{author}{Ding, K.}, \bibinfo{author}{Li, H.}, \bibinfo{author}{Wang, M.}, \bibinfo{author}{Yu, D.}, \bibinfo{author}{Zhang, G.}, \bibinfo{author}{Yang, Y.}, \bibinfo{author}{Wang, K.},
  \bibinfo{author}{Yuille, A.L.}, \bibinfo{author}{Zhou, Z.}, \bibinfo{year}{2026}b.
\newblock \bibinfo{title}{Early and prediagnostic detection of pancreatic cancer from computed tomography}.
\newblock \bibinfo{journal}{arXiv preprint arXiv:2601.22134} \URLprefix \url{https://github.com/BodyMaps/ePAI}.
\bibitem[{Li et~al.(2026c)Li, Bassi, Zhou, Wasserthal, Hamamci, Er, Menze, Akan, P{\l}otka, Tang, Xu, Wang, Yang, Yuille and Zhou}]{li2026cancerverse}
\bibinfo{author}{Li, W.}, \bibinfo{author}{Bassi, P.R.A.S.}, \bibinfo{author}{Zhou, X.}, \bibinfo{author}{Wasserthal, J.}, \bibinfo{author}{Hamamci, I.}, \bibinfo{author}{Er, S.}, \bibinfo{author}{Menze, B.}, \bibinfo{author}{Akan, G.E.}, \bibinfo{author}{P{\l}otka, S.}, \bibinfo{author}{Tang, Y.}, \bibinfo{author}{Xu, D.}, \bibinfo{author}{Wang, K.}, \bibinfo{author}{Yang, Y.}, \bibinfo{author}{Yuille, A.L.}, \bibinfo{author}{Zhou, Z.}, \bibinfo{year}{2026}c.
\newblock \bibinfo{title}{{CancerVerse}: A fully open longitudinal and multimodal dataset for multi-cancer screening}, in: \bibinfo{booktitle}{International Conference on Medical Image Computing and Computer-Assisted Intervention (MICCAI)}.
\bibitem[{Li et~al.(2024)Li, Qu, Chen, Bassi, Shi, Lai, Yu, Xue, Chen, Lin, Tang, Cao, Han, Zhang, Liu, Zhang, Ma, Wang, Zhang, Yuille and Zhou}]{li2024abdomenatlas}
\bibinfo{author}{Li, W.}, \bibinfo{author}{Qu, C.}, \bibinfo{author}{Chen, X.}, \bibinfo{author}{Bassi, P.R.}, \bibinfo{author}{Shi, Y.}, \bibinfo{author}{Lai, Y.}, \bibinfo{author}{Yu, Q.}, \bibinfo{author}{Xue, H.}, \bibinfo{author}{Chen, Y.}, \bibinfo{author}{Lin, X.}, \bibinfo{author}{Tang, Y.}, \bibinfo{author}{Cao, Y.}, \bibinfo{author}{Han, H.}, \bibinfo{author}{Zhang, Z.}, \bibinfo{author}{Liu, J.}, \bibinfo{author}{Zhang, T.}, \bibinfo{author}{Ma, Y.}, \bibinfo{author}{Wang, J.}, \bibinfo{author}{Zhang, G.}, \bibinfo{author}{Yuille, A.}, \bibinfo{author}{Zhou, Z.}, \bibinfo{year}{2024}.
\newblock \bibinfo{title}{{AbdomenAtlas}: A large-scale, detailed-annotated, \& multi-center dataset for efficient transfer learning and open algorithmic benchmarking}.
\newblock \bibinfo{journal}{Medical Image Analysis} \bibinfo{volume}{97}, \bibinfo{pages}{103285}.
\newblock \URLprefix \url{https://github.com/MrGiovanni/AbdomenAtlas}.
\bibitem[{Li et~al.(2025b)Li, Zhou, Chen, Lin, Bassi, Chen, Ye, Zhu, Ding, Li, Wang, Yang, Tang, Xu, Yuille and Zhou}]{li2025pants}
\bibinfo{author}{Li, W.}, \bibinfo{author}{Zhou, X.}, \bibinfo{author}{Chen, Q.}, \bibinfo{author}{Lin, T.}, \bibinfo{author}{Bassi, P.R.}, \bibinfo{author}{Chen, X.}, \bibinfo{author}{Ye, C.}, \bibinfo{author}{Zhu, Z.}, \bibinfo{author}{Ding, K.}, \bibinfo{author}{Li, H.}, \bibinfo{author}{Wang, K.}, \bibinfo{author}{Yang, Y.}, \bibinfo{author}{Tang, Y.}, \bibinfo{author}{Xu, D.}, \bibinfo{author}{Yuille, A.L.}, \bibinfo{author}{Zhou, Z.}, \bibinfo{year}{2025}b.
\newblock \bibinfo{title}{{PanTS}: The pancreatic tumor segmentation dataset}, in: \bibinfo{booktitle}{Conference on Neural Information Processing Systems (NeurIPS) Datasets and Benchmarks Track}.
\newblock \URLprefix \url{https://github.com/MrGiovanni/PanTS}.
\bibitem[{Liu et~al.(2024)Liu, Zhang, Wang, Yavuz, Chen, Yuan, Li, Yang, Yuille, Tang and Zhou}]{liu2024universal}
\bibinfo{author}{Liu, J.}, \bibinfo{author}{Zhang, Y.}, \bibinfo{author}{Wang, K.}, \bibinfo{author}{Yavuz, M.C.}, \bibinfo{author}{Chen, X.}, \bibinfo{author}{Yuan, Y.}, \bibinfo{author}{Li, H.}, \bibinfo{author}{Yang, Y.}, \bibinfo{author}{Yuille, A.}, \bibinfo{author}{Tang, Y.}, \bibinfo{author}{Zhou, Z.}, \bibinfo{year}{2024}.
\newblock \bibinfo{title}{Universal and extensible language-vision models for organ segmentation and tumor detection from abdominal computed tomography}.
\newblock \bibinfo{journal}{Medical Image Analysis} \bibinfo{volume}{97}, \bibinfo{pages}{103226}.
\newblock \URLprefix \url{https://github.com/ljwztc/CLIP-Driven-Universal-Model}.
\bibitem[{Ma et~al.(2024)Ma, Kim, Li, Baharoon, Asakereh, Lyu and Wang}]{ma2024segment}
\bibinfo{author}{Ma, J.}, \bibinfo{author}{Kim, S.}, \bibinfo{author}{Li, F.}, \bibinfo{author}{Baharoon, M.}, \bibinfo{author}{Asakereh, R.}, \bibinfo{author}{Lyu, H.}, \bibinfo{author}{Wang, B.}, \bibinfo{year}{2024}.
\newblock \bibinfo{title}{{Segment Anything} in medical images and videos: Benchmark and deployment}.
\newblock \bibinfo{journal}{arXiv preprint arXiv:2408.03322} .
\bibitem[{Miller et~al.(1981)Miller, Hoogstraten, Staquet and Winkler}]{miller1981reporting}
\bibinfo{author}{Miller, A.}, \bibinfo{author}{Hoogstraten, B.}, \bibinfo{author}{Staquet, M.}, \bibinfo{author}{Winkler, A.}, \bibinfo{year}{1981}.
\newblock \bibinfo{title}{Reporting results of cancer treatment}.
\newblock \bibinfo{journal}{Cancer} \bibinfo{volume}{47}, \bibinfo{pages}{207--214}.
\bibitem[{Park et~al.(2020)Park, Chu, Fishman, Yuille, Vogelstein, Kinzler, Horton, Hruban, Zinreich, Fouladi et~al.}]{park2020annotated}
\bibinfo{author}{Park, S.}, \bibinfo{author}{Chu, L.}, \bibinfo{author}{Fishman, E.}, \bibinfo{author}{Yuille, A.}, \bibinfo{author}{Vogelstein, B.}, \bibinfo{author}{Kinzler, K.}, \bibinfo{author}{Horton, K.}, \bibinfo{author}{Hruban, R.}, \bibinfo{author}{Zinreich, E.}, \bibinfo{author}{Fouladi, D.F.}, et~al., \bibinfo{year}{2020}.
\newblock \bibinfo{title}{Annotated normal {CT} data of the abdomen for deep learning: Challenges and strategies for implementation}.
\newblock \bibinfo{journal}{Diagnostic and interventional imaging} \bibinfo{volume}{101}, \bibinfo{pages}{35--44}.
\bibitem[{Qu et~al.(2023)Qu, Zhang, Qiao, Liu, Tang, Yuille and Zhou}]{qu2023annotating}
\bibinfo{author}{Qu, C.}, \bibinfo{author}{Zhang, T.}, \bibinfo{author}{Qiao, H.}, \bibinfo{author}{Liu, J.}, \bibinfo{author}{Tang, Y.}, \bibinfo{author}{Yuille, A.}, \bibinfo{author}{Zhou, Z.}, \bibinfo{year}{2023}.
\newblock \bibinfo{title}{{AbdomenAtlas-8K}: Annotating 8,000 abdominal {CT} volumes for multi-organ segmentation in three weeks}, in: \bibinfo{booktitle}{Conference on Neural Information Processing Systems (NeurIPS) Datasets and Benchmarks Track}.
\newblock \URLprefix \url{https://github.com/MrGiovanni/AbdomenAtlas}.
\bibitem[{Radford et~al.(2021)Radford, Kim, Hallacy, Ramesh, Goh, Agarwal, Sastry, Askell, Mishkin, Clark et~al.}]{radford2021learning}
\bibinfo{author}{Radford, A.}, \bibinfo{author}{Kim, J.W.}, \bibinfo{author}{Hallacy, C.}, \bibinfo{author}{Ramesh, A.}, \bibinfo{author}{Goh, G.}, \bibinfo{author}{Agarwal, S.}, \bibinfo{author}{Sastry, G.}, \bibinfo{author}{Askell, A.}, \bibinfo{author}{Mishkin, P.}, \bibinfo{author}{Clark, J.}, et~al., \bibinfo{year}{2021}.
\newblock \bibinfo{title}{Learning transferable visual models from natural language supervision}, in: \bibinfo{booktitle}{International conference on machine learning}, \bibinfo{organization}{PMLR}. pp. \bibinfo{pages}{8748--8763}.
\bibitem[{Rajchl et~al.(2017)Rajchl, Lee, Oktay, Kamnitsas, Passerat-Palmbach, Bai, Damodaram, Rutherford, Hajnal, Kainz and Rueckert}]{deepcut}
\bibinfo{author}{Rajchl, M.}, \bibinfo{author}{Lee, M.C.H.}, \bibinfo{author}{Oktay, O.}, \bibinfo{author}{Kamnitsas, K.}, \bibinfo{author}{Passerat-Palmbach, J.}, \bibinfo{author}{Bai, W.}, \bibinfo{author}{Damodaram, M.}, \bibinfo{author}{Rutherford, M.A.}, \bibinfo{author}{Hajnal, J.V.}, \bibinfo{author}{Kainz, B.}, \bibinfo{author}{Rueckert, D.}, \bibinfo{year}{2017}.
\newblock \bibinfo{title}{{DeepCut}: Object segmentation from bounding box annotations using convolutional neural networks}.
\newblock \bibinfo{journal}{IEEE Transactions on Medical Imaging} \bibinfo{volume}{36}, \bibinfo{pages}{674--683}.
\newblock \DOIprefix\doi{10.1109/TMI.2016.2621185}.
\bibitem[{Sellergren et~al.(2025)Sellergren, Kazemzadeh, Jaroensri, Kiraly, Traverse, Kohlberger, Xu, Jamil, Hughes, Lau et~al.}]{sellergren2025medgemma}
\bibinfo{author}{Sellergren, A.}, \bibinfo{author}{Kazemzadeh, S.}, \bibinfo{author}{Jaroensri, T.}, \bibinfo{author}{Kiraly, A.}, \bibinfo{author}{Traverse, M.}, \bibinfo{author}{Kohlberger, T.}, \bibinfo{author}{Xu, S.}, \bibinfo{author}{Jamil, F.}, \bibinfo{author}{Hughes, C.}, \bibinfo{author}{Lau, C.}, et~al., \bibinfo{year}{2025}.
\newblock \bibinfo{title}{{MedGemma} technical report}.
\newblock \bibinfo{journal}{arXiv preprint arXiv:2507.05201} .
\bibitem[{Tang et~al.(2022)Tang, Yang, Li, Roth, Landman, Xu, Nath and Hatamizadeh}]{tang2022self}
\bibinfo{author}{Tang, Y.}, \bibinfo{author}{Yang, D.}, \bibinfo{author}{Li, W.}, \bibinfo{author}{Roth, H.R.}, \bibinfo{author}{Landman, B.}, \bibinfo{author}{Xu, D.}, \bibinfo{author}{Nath, V.}, \bibinfo{author}{Hatamizadeh, A.}, \bibinfo{year}{2022}.
\newblock \bibinfo{title}{Self-supervised pre-training of {Swin} transformers for {3D} medical image analysis}, in: \bibinfo{booktitle}{Proceedings of the IEEE/CVF Conference on Computer Vision and Pattern Recognition}, pp. \bibinfo{pages}{20730--20740}.
\bibitem[{Wald et~al.(2024)Wald, Ulrich, Lukyanenko, Goncharov, Paderno, Miller, Maerkisch, J{\"a}ger and Maier-Hein}]{wald2024revisiting}
\bibinfo{author}{Wald, T.}, \bibinfo{author}{Ulrich, C.}, \bibinfo{author}{Lukyanenko, S.}, \bibinfo{author}{Goncharov, A.}, \bibinfo{author}{Paderno, A.}, \bibinfo{author}{Miller, M.}, \bibinfo{author}{Maerkisch, L.}, \bibinfo{author}{J{\"a}ger, P.F.}, \bibinfo{author}{Maier-Hein, K.}, \bibinfo{year}{2024}.
\newblock \bibinfo{title}{Revisiting {MAE} pre-training for {3D} medical image segmentation}.
\newblock \bibinfo{journal}{arXiv preprint arXiv:2410.23132} .
\bibitem[{Xia et~al.(2022)Xia, Yu, Chu, Kawamoto, Park, Liu, Chen, Zhu, Li, Zhou, Lu, Wang, Shen, Xie, Zhou, Wolfgang, Javed, Fouladi, Shayesteh, Graves, Blanco, Zinreich, Kinny-K{\"o}ster, Kinzler, Hruban, Vogelstein, Yuille and Fishman}]{xia2022felix}
\bibinfo{author}{Xia, Y.}, \bibinfo{author}{Yu, Q.}, \bibinfo{author}{Chu, L.}, \bibinfo{author}{Kawamoto, S.}, \bibinfo{author}{Park, S.}, \bibinfo{author}{Liu, F.}, \bibinfo{author}{Chen, J.}, \bibinfo{author}{Zhu, Z.}, \bibinfo{author}{Li, B.}, \bibinfo{author}{Zhou, Z.}, \bibinfo{author}{Lu, Y.}, \bibinfo{author}{Wang, Y.}, \bibinfo{author}{Shen, W.}, \bibinfo{author}{Xie, L.}, \bibinfo{author}{Zhou, Y.}, \bibinfo{author}{Wolfgang, C.}, \bibinfo{author}{Javed, A.}, \bibinfo{author}{Fouladi, D.F.}, \bibinfo{author}{Shayesteh, S.}, \bibinfo{author}{Graves, J.}, \bibinfo{author}{Blanco, A.}, \bibinfo{author}{Zinreich, E.S.}, \bibinfo{author}{Kinny-K{\"o}ster, B.}, \bibinfo{author}{Kinzler, K.}, \bibinfo{author}{Hruban, R.H.}, \bibinfo{author}{Vogelstein, B.}, \bibinfo{author}{Yuille, A.L.}, \bibinfo{author}{Fishman, E.K.}, \bibinfo{year}{2022}.
\newblock \bibinfo{title}{The {FELIX} project: Deep networks to detect pancreatic neoplasms}.
\newblock \bibinfo{journal}{medRxiv} .
\bibitem[{Yang et~al.(2025)Yang, Li, Yang, Zhang, Hui, Zheng, Yu, Gao, Huang, Lv, Zheng, Liu, Zhou, Huang, Hu, Ge, Wei, Lin, Tang, Yang, Tu, Zhang, Yang, Yang, Zhou, Zhou, Lin, Dang, Bao, Yang, Yu, Deng, Li, Xue, Li, Zhang, Wang, Zhu, Men, Gao, Liu, Luo, Li, Tang, Yin, Ren, Wang, Zhang, Ren, Fan, Su, Zhang, Zhang, Wan, Liu, Wang, Cui, Zhang, Zhou and Qiu}]{yang2025qwen3technicalreport}
\bibinfo{author}{Yang, A.}, \bibinfo{author}{Li, A.}, \bibinfo{author}{Yang, B.}, \bibinfo{author}{Zhang, B.}, \bibinfo{author}{Hui, B.}, \bibinfo{author}{Zheng, B.}, \bibinfo{author}{Yu, B.}, \bibinfo{author}{Gao, C.}, \bibinfo{author}{Huang, C.}, \bibinfo{author}{Lv, C.}, \bibinfo{author}{Zheng, C.}, \bibinfo{author}{Liu, D.}, \bibinfo{author}{Zhou, F.}, \bibinfo{author}{Huang, F.}, \bibinfo{author}{Hu, F.}, \bibinfo{author}{Ge, H.}, \bibinfo{author}{Wei, H.}, \bibinfo{author}{Lin, H.}, \bibinfo{author}{Tang, J.}, \bibinfo{author}{Yang, J.}, \bibinfo{author}{Tu, J.}, \bibinfo{author}{Zhang, J.}, \bibinfo{author}{Yang, J.}, \bibinfo{author}{Yang, J.}, \bibinfo{author}{Zhou, J.}, \bibinfo{author}{Zhou, J.}, \bibinfo{author}{Lin, J.}, \bibinfo{author}{Dang, K.}, \bibinfo{author}{Bao, K.}, \bibinfo{author}{Yang, K.}, \bibinfo{author}{Yu, L.}, \bibinfo{author}{Deng, L.}, \bibinfo{author}{Li, M.}, \bibinfo{author}{Xue, M.}, \bibinfo{author}{Li, M.}, \bibinfo{author}{Zhang, P.}, \bibinfo{author}{Wang, P.},
  \bibinfo{author}{Zhu, Q.}, \bibinfo{author}{Men, R.}, \bibinfo{author}{Gao, R.}, \bibinfo{author}{Liu, S.}, \bibinfo{author}{Luo, S.}, \bibinfo{author}{Li, T.}, \bibinfo{author}{Tang, T.}, \bibinfo{author}{Yin, W.}, \bibinfo{author}{Ren, X.}, \bibinfo{author}{Wang, X.}, \bibinfo{author}{Zhang, X.}, \bibinfo{author}{Ren, X.}, \bibinfo{author}{Fan, Y.}, \bibinfo{author}{Su, Y.}, \bibinfo{author}{Zhang, Y.}, \bibinfo{author}{Zhang, Y.}, \bibinfo{author}{Wan, Y.}, \bibinfo{author}{Liu, Y.}, \bibinfo{author}{Wang, Z.}, \bibinfo{author}{Cui, Z.}, \bibinfo{author}{Zhang, Z.}, \bibinfo{author}{Zhou, Z.}, \bibinfo{author}{Qiu, Z.}, \bibinfo{year}{2025}.
\newblock \bibinfo{title}{Qwen3 technical report}.
\newblock \URLprefix \url{https://arxiv.org/abs/2505.09388}, \href{http://arxiv.org/abs/2505.09388}{\tt arXiv:2505.09388}.
\bibitem[{Zeman et~al.(1996)Zeman, Zeiberg, Hayes, Silverman, Cooper and Garra}]{zeman1996helical}
\bibinfo{author}{Zeman, R.K.}, \bibinfo{author}{Zeiberg, A.}, \bibinfo{author}{Hayes, W.S.}, \bibinfo{author}{Silverman, P.M.}, \bibinfo{author}{Cooper, C.}, \bibinfo{author}{Garra, B.S.}, \bibinfo{year}{1996}.
\newblock \bibinfo{title}{Helical {CT} of renal masses: The value of delayed scans}.
\newblock \bibinfo{journal}{American Journal of Roentgenology} \bibinfo{volume}{167}, \bibinfo{pages}{771--776}.
\newblock \DOIprefix\doi{10.2214/AJR.167.3.8751698}.
\bibitem[{Zhang et~al.(2021)Zhang, Li, Du, Qin, Wang, Chen, Liu, Gao, Ma and Lei}]{zhang20213d}
\bibinfo{author}{Zhang, Y.}, \bibinfo{author}{Li, H.}, \bibinfo{author}{Du, J.}, \bibinfo{author}{Qin, J.}, \bibinfo{author}{Wang, T.}, \bibinfo{author}{Chen, Y.}, \bibinfo{author}{Liu, B.}, \bibinfo{author}{Gao, W.}, \bibinfo{author}{Ma, G.}, \bibinfo{author}{Lei, B.}, \bibinfo{year}{2021}.
\newblock \bibinfo{title}{3d multi-attention guided multi-task learning network for automatic gastric tumor segmentation and lymph node classification}.
\newblock \bibinfo{journal}{IEEE transactions on medical imaging} \bibinfo{volume}{40}, \bibinfo{pages}{1618--1631}.
\bibitem[{Zhou et~al.(2025)Zhou, Zhao, Zhuang, Yu, Yuille and Zhou}]{zhou2025efficient}
\bibinfo{author}{Zhou, X.}, \bibinfo{author}{Zhao, Y.}, \bibinfo{author}{Zhuang, C.}, \bibinfo{author}{Yu, D.}, \bibinfo{author}{Yuille, A.L.}, \bibinfo{author}{Zhou, Z.}, \bibinfo{year}{2025}.
\newblock \bibinfo{title}{Efficient human-in-the-loop pancreatic tumors annotation via large-scale pre-trained model with adaptive post-processing}, in: \bibinfo{booktitle}{IEEE International Symposium on Biomedical Imaging (ISBI)}.
\bibitem[{Zhou et~al.(2021)Zhou, Sodha, Pang, Gotway and Liang}]{zhou2021models}
\bibinfo{author}{Zhou, Z.}, \bibinfo{author}{Sodha, V.}, \bibinfo{author}{Pang, J.}, \bibinfo{author}{Gotway, M.B.}, \bibinfo{author}{Liang, J.}, \bibinfo{year}{2021}.
\newblock \bibinfo{title}{Models {Genesis}}.
\newblock \bibinfo{journal}{Medical Image Analysis} \bibinfo{volume}{67}, \bibinfo{pages}{101840}.
\newblock \URLprefix \url{https://github.com/MrGiovanni/ModelsGenesis}.
\bibitem[{Zhou et~al.(2019)Zhou, Sodha, Siddiquee, Feng, Tajbakhsh, Gotway and Liang}]{zhou2019models}
\bibinfo{author}{Zhou, Z.}, \bibinfo{author}{Sodha, V.}, \bibinfo{author}{Siddiquee, M.M.R.}, \bibinfo{author}{Feng, R.}, \bibinfo{author}{Tajbakhsh, N.}, \bibinfo{author}{Gotway, M.B.}, \bibinfo{author}{Liang, J.}, \bibinfo{year}{2019}.
\newblock \bibinfo{title}{Models {Genesis}: Generic autodidactic models for {3D} medical image analysis}, in: \bibinfo{booktitle}{International Conference on Medical Image Computing and Computer-Assisted Intervention}, \bibinfo{organization}{Springer}. pp. \bibinfo{pages}{384--393}.
\newblock \URLprefix \url{https://github.com/MrGiovanni/ModelsGenesis}.
\bibitem[{Zhu et~al.(2020)Zhu, Li, Hu, Ma, Zhou and Zheng}]{zhu2020rubik}
\bibinfo{author}{Zhu, J.}, \bibinfo{author}{Li, Y.}, \bibinfo{author}{Hu, Y.}, \bibinfo{author}{Ma, K.}, \bibinfo{author}{Zhou, S.K.}, \bibinfo{author}{Zheng, Y.}, \bibinfo{year}{2020}.
\newblock \bibinfo{title}{{Rubik's Cube+}: A self-supervised feature learning framework for {3D} medical image analysis}.
\newblock \bibinfo{journal}{Medical Image Analysis} \bibinfo{volume}{64}, \bibinfo{pages}{101746}.

\end{thebibliography}


\clearpage
\appendix

\section{Details on the Loss Functions}
\label{app:loss_details}
\subsection{Weighted Cross-Entropy and Gaussian Ball Kernel}

\paragraph{Weighted cross-entropy.}
We replace the cross-entropy in Eq.~\ref{eq:ce_dice} with a rank-weighted variant that down-weights voxels the model is less confident about. Tumor borders typically receive intermediate predicted probabilities, so this rank-weighting reduces the gradient contribution of border voxels and concentrates supervision on confidently predicted tumor voxels.

The variant operates on the binary mask $\mathbf{M}$ marking the most probable tumor voxels in organ $o$ (the union of the per-tumor masks $\mathbf{M}_i$, Eq.~\ref{eq:union_masks}) and on the tumor segmentation output $\mathbf{T}^{o}$. At each optimization step, we sort the voxels in $\mathbf{M}$ by predicted tumor probability in descending order, assigning each voxel a rank $r(x,y,z) \in \{0, \dots, n-1\}$, where $r=0$ is the highest-probability voxel and $n$ is the total number of voxels in $\mathbf{M}$. The weight $w$ then decays geometrically with rank:

\begin{gather}
\label{eq:weighted_ce}
\begin{aligned}
\mathrm{CE}\!\big((1-\mathbf{M}')\!\odot\!\mathbf{T}^{o},\,\mathbf{M}\big)
= \\-\frac{1}{|\Omega|}\!\!\sum_{(x,y,z)\in\Omega}\!\!
w(x,y,z)\,\Big[
\mathbf{M}\log\!\big((1-\mathbf{M}')\mathbf{T}^{o}\big)\\
+(1-\mathbf{M})\log\!\big(1-(1-\mathbf{M}')\mathbf{T}^{o}\big)
\Big]\end{aligned}\\
\label{eq:weighted_ce_weights}
w(x,y,z)=
\begin{cases}
\dfrac{n\,(1-c)^{r(x,y,z)/n}}
       {\sum_{r'=0}^{n-1}(1-c)^{r'/n}}, & \text{if }\mathbf{M}(x,y,z)=1\\[6pt]
1, & \text{otherwise}
\end{cases}
\end{gather}

The rank-weighting applies only to voxels in $\mathbf{M}$; all other voxels keep weight $1$, so the loss outside $\mathbf{M}$ is unchanged from standard cross-entropy. The multiplication of $\mathbf{T}^{o}$ by $(1-\mathbf{M}')$ zeros the gradient at the margin voxels, as in Eq.~\ref{eq:ce_dice}. $\Omega$ is the set of all voxels in $\mathbf{T}^{o}$. The hyper-parameter $c \in (0,1)$ controls how sharply the weight concentrates on the top-ranked voxels: we set $c=0.5$, so the weight decays smoothly across the ranking and the least confident voxel in $\mathbf{M}$ receives half the weight of the most confident one (the most confident half of the voxels then carries about 59\% of the total tumor-voxel weight). The normalization in Eq.~\ref{eq:weighted_ce_weights} ensures $\sum_{(x,y,z) \in \mathbf{M}} w(x,y,z) = n$, matching the scale of the unweighted cross-entropy; setting $c \to 0$ recovers the standard cross-entropy. This weighting scheme is based on Global Weighted Rank Pooling~\citep{kolesnikov2016seed}, adding a hard cutoff at $n$.

\paragraph{Gaussian-decay ball kernel.}
In the Ball Loss, Ball Convolutions locate each tumor $i$ (Eq.~\ref{eq:ball_max}). The kernel of the Ball Convolution is a ball with the same diameter as tumor $i$, according to the report (Eq.~\ref{eq:ball_kernel}). The Ball Convolution is applied at the output of the segmentation model (per-voxel tumor probabilities), excluding tumor probabilities outside organ $o$ ($\mathbf{T}^{o}\odot\mathbf{O}$, where $\mathbf{O}$ is the organ segmentation mask). At each position of this kernel, it sums all the per-voxel tumor probabilities inside the ball. With a uniform binary kernel, every voxel inside the ball contributes equally to this sum, so the convolution maximum is insensitive to where, within the ball, predicted tumor probability is highest. We want the argmax to favor positions where probability is concentrated near the ball center, since predicted tumor probabilities tend to peak at tumor centers.

We achieve this by replacing the binary kernel $\mathbf{k}_i$ with a Gaussian-tapered version that is largest at the kernel center, smaller near the ball boundary, and zero outside the ball:

\begin{gather}
\label{eq:ball_kernel_gaussian}
\mathbf{k}_{i}(u,v,w) =
\frac{1}{Z_i}\,\mathbb{1}\!\Big[\sqrt{u^{2}+v^{2}+w^{2}}\le\tfrac{d_i}{2}\Big]\,
e^{-\tfrac{u^{2}+v^{2}+w^{2}}{2\sigma_i^{2}}}\\
\text{where }
\sigma_i = \alpha\,\frac{d_i}{2},
\qquad \alpha = 3
\end{gather}

Here $(u,v,w)$ are the kernel coordinates measured from the kernel center, and $d_i$ is the diameter reported for tumor $i$. The indicator $\mathbb{1}[\cdot]$ preserves the hard cutoff at the ball boundary from Eq.~\ref{eq:ball_kernel}, so $\mathbf{k}_i$ is zero outside the ball. The Gaussian standard deviation $\sigma_i$ is scaled to the ball radius $d_i/2$ via $\alpha$, so the decay shape is identical across reported diameters. The normalization $Z_i$ is chosen so $\sum_{u,v,w} \mathbf{k}_i(u,v,w) = 1$, keeping Ball Convolution outputs comparable across kernel sizes; because $Z_i$ is constant for a given tumor, it does not change the location of the maximum in Eq.~\ref{eq:ball_max}.

We set $\alpha = 3$, which gives a boundary-to-center kernel ratio of $\exp(-1/18) \approx 0.95$. The decay is therefore very mild: enough to break argmax ties in favor of positions where predicted probability is concentrated near the kernel center, but small enough that the output of the Ball Convolution is close to a (scaled) sum of the tumor probabilities inside each ball position.

\subsection{Volume Loss for Reports without Tumor Size or Count}
\label{supp:volume_loss_missing_info}
The standard Volume Loss penalizes the segmented tumor volume $V_{s,o}$ for deviating from the reported volume $V_{r,o}$, which is computed from per-tumor diameters. When a report states that an organ contains tumors but does not provide diameters or counts, $V_{r,o}$ cannot be computed, so the standard Volume Loss cannot be applied. In this case, we replace the precise $V_{r,o}$ target with a tolerated range $[V_{min}, V_{max}]$. Any segmented volume inside this range incurs zero loss and zero gradient, while volumes outside the range are penalized as before.

We calibrate $V_{min}$ and $V_{max}$ from the diameter distribution in our reports: fewer than 5.5\% of images have all tumors below 5 mm and fewer than 1.66\% have tumors above 120 mm. We therefore set $V_{min} = 65$ mm$^3$ and $V_{max} = 904{,}779$ mm$^3$, the volumes of spheres at these two diameters. The high-tolerance loss (Eq.~\ref{eq:vol_high_tolerance}) implements the tolerated range by replacing $V_{r,o}$ in the standard $L_{\text{forg},o}$ (Eq.~\ref{eq:loss_formula}) with a target $\widehat{V}_{r,o}$ that equals $V_{s,o}$ whenever $V_{s,o}$ lies in $[V_{min}, V_{max}]$. With $\widehat{V}_{r,o} = V_{s,o}$, the loss and its gradient vanish; outside the range, $\widehat{V}_{r,o}$ pulls the segmented tumors' volume back toward the range.

\begin{gather}
\label{eq:vol_high_tolerance}
L_{\text{vol},o}
= L_{\text{forg},o}\!\left(V_{s,o},\,\widehat{V}_{r,o}\right)
+ L_{\text{bkg},o}\!\left(\mathbf{T}^{o}\right)\\
\text{where}\quad \widehat{V}_{r,o} =
\begin{cases}
V_{min} & \text{if } V_{s,o} < V_{min},\\
V_{s,o} & \text{if } V_{min} \le V_{s,o} \le V_{max},\\
V_{max} & \text{if } V_{s,o} > V_{max}.
\end{cases}
\end{gather}

\subsection{Ball Loss for Reports without Tumor Size or Count}
\label{supp:ball_loss_missing_info}

The standard Ball Loss requires reports to provide per-tumor diameters because diameters are used to define the kernel of the Ball Convolution, and because the loss derives the voxel count $N_i$ from the reported diameters (Sec.~\ref{sec:ball_loss}). The high-tolerance variant of the Ball Loss accepts reports with missing diameters or missing tumor counts by enforcing only a lower bound on the segmentation: the model must segment at least one small tumor in organ $o$, but it is not penalized for segmenting a larger tumor or multiple tumors. 

First, we locate the most probable voxels for one small tumor by running the Ball Convolution with a small diameter (we use 5 mm, the same lower cutoff used in the high-tolerance Volume Loss above). This produces a small highest-probability ball, which may actually sit near the center of a large tumor. Inside this ball, we select the $N_i$ most probable voxels (with $N_i$ set to the voxel count of a 5 mm sphere, reduced by 20\%). We then create a mask $\mathbf{M}$ that is 1 in these $N_i$ most probable voxels and 0 outside, as in the standard Ball Loss (Eq.~\ref{eq:mask}).

The mask $\mathbf{M}'$, originally defined as a narrow tolerance margin around $\mathbf{M}$ (Eq.~\ref{eq:margin}), is redefined here. In the high-tolerance Ball Loss, we set $\mathbf{M}'$ to 1 at every voxel inside organ $o$ except the $N_i$ voxels in $\mathbf{M}$ (Eq.~\ref{eq:m_lin_big}). When the Dice and cross-entropy losses are applied as in the standard Ball Loss (Eq.~\ref{eq:ce_dice}), the multiplication by $(1-\mathbf{M}')$ now zeros the gradient at every non-located voxel inside the organ. The model is therefore supervised only at the $N_i$ most probable voxels (where $\mathbf{M}=1$, gradient pushes probabilities toward 1) and outside the organ (where both masks are 0, gradient pushes probabilities toward 0); everywhere else, the model is free to segment larger or additional tumors without penalty. This is shown in Eq.~\ref{eq:m_lin_big}, where $\mathbf{O}$ is the pre-saved organ segmentation mask for organ $o$.

\begin{equation}
\label{eq:m_lin_big}
\mathbf{M}' = \mathbf{O} \odot (\mathbf{1} - \mathbf{M})
\end{equation}

We then apply the same Dice and cross-entropy losses as in the standard Ball Loss (Eqs.~\ref{eq:ce_dice} to \ref{eq:ball_all_organs}, copied below), using $\mathbf{M}$ as the target. The multiplication by $(1-\mathbf{M}')$ zeros the loss gradient at every voxel in $\mathbf{M}'$, leaving those voxels unpenalized.

\begin{gather*}
\begin{aligned}
L'_{\text{ball},o}=\mathrm{CE}((1-\mathbf{M}')\odot\mathbf{T}^{o},\mathbf{M})
    +\\
    \mathrm{Dice}((1-\mathbf{M}')\odot\mathbf{T}^{o},\mathbf{M})\end{aligned}\\
L_{\text{ball},o}
=
\begin{cases}
    L'_{\text{ball},o}, & \text{if tumors in } o \\
\text{CE}(\mathbf{T}^{o},\mathbf{0}), & \text{otherwise}
\end{cases}\\
L_{\text{ball}} = \sum_{o\in\mathcal{O}}L_{\text{ball},o}
\end{gather*}

The result is a loss that maximizes tumor probabilities in a small number of voxels (the most likely voxels for a 5 mm tumor), minimizes probabilities outside the organ $o$ as usual, and applies no loss or gradient to all other voxels inside the organ $o$. The model is therefore encouraged to segment at least one small tumor, but it is free to segment a larger tumor or multiple tumors, without penalty.

\section{Training Details}\label{supp:training}

We trained all segmentation models on 3D CT patches of $128\times128\times128$ voxels at an isotropic spacing of 1 mm. For CT-Report pairs, each patch was centered on a target organ (selected with 80\% probability among the organs reported as containing tumors) and was sized to fully contain that organ. Without this constraint, a tumor mentioned in the report could fall outside the patch, and the report-based losses would push the model toward a tumor that is not visible in the input.

Optimization followed the MedFormer defaults \citep{gao2022data}: AdamW with weight decay $5\times10^{-2}$, gradient norm clipping at 1, batch size 2, and 100 epochs of 1,000 batches each. The learning rate started at $1\times10^{-4}$ with a 5-epoch warmup followed by polynomial decay. CT intensities were clipped to $[-991, 500]$ HU and then normalized. Data augmentation comprised rotation, brightness, gamma, contrast, Gaussian blur, and Gaussian noise, applied per sample as in MedFormer. The only R-Super-specific hyper-parameters that we tuned are the loss weights, fixed across all experiments at 1 for the segmentation losses (Dice and cross-entropy on CT-Mask pairs) and 0.1 for the Volume and Ball Losses. These weights were tuned once on a validation set (10\% of UCSF-Train, held out at random) and reused throughout, so no per-experiment tuning is required. When CT-Mask and CT-Report pairs were combined, CT-Mask pairs were oversampled to 50\% of each epoch's training samples.

All baselines compared to R-Super (standard segmentation, MTL, CLIP, Models Genesis, report-guided pseudo-labels) used the same MedFormer backbone and hyper-parameters as R-Super, so that differences in performance reflect the training methodology rather than the architecture. MTL additionally has a classification head attached to the MedFormer encoder output. Standard segmentation was trained on the CT-Mask pairs; MTL was trained on both CT-Mask and CT-Report pairs; CLIP was first pre-trained on the CT-Report pairs with a CLIP loss and then fine-tuned on the CT-Mask pairs; Models Genesis was first pre-trained on our CT scans with a masked-autoencoder loss and then fine-tuned on the CT-Mask pairs; report-guided pseudo-labels used the standard segmentation model to create pseudo-masks, masks were refined following \citet{bosma2023semisupervised}, and a new standard segmentation model (here, also MedFormer) was trained on these pseudo-masks plus the original tumor masks (following the same hyper-parameters as the other models).

nnU-Net \citep{isensee2021nnu} is the only method with a different architecture and training scheme. As a self-configuring framework intended to be used out of the box, it was trained on CT-Mask pairs using its standard procedure, automatic hyper-parameters, and ResEncL architecture. The only non-default setting was the isotropic voxel spacing of 1 mm, matching the spacing used by all other models.

\section{Result Variability}
\label{app:variability}

{\PBrev Tabs.~\ref{tab:cif_all_results} to \ref{tab:cif_ablations} report the variability associated with the results in the corresponding tables of the main paper. For the segmentation metrics (DSC and NSD), we report the standard deviation. These standard deviations are large because segmentation performance varies substantially across patients. Tumors in different patients vary considerably in size, shape, and texture, leading to differences in segmentation difficulty and, consequently, varying DSC and NSD scores. Although organs exhibit less variation than tumors, organ segmentation also yields large standard deviations in DSC and NSD \citep{bassi2024touchstone}. For the tumor detection metrics (sensitivity, specificity, F1-Score, and AUC), we report 95\% confidence intervals, calculated using 1,000 bootstrap samples. Wide confidence intervals or standard deviations for individual results do not necessarily mean that the differences between methods are not statistically significant. Metrics that are significantly higher than standard segmentation ($p<0.05$) are highlighted in orange in Tabs.~\ref{tab:cif_all_results} to \ref{tab:cif_jhh_ucsf}.}


\begin{table*}[!t]
\centering
\scriptsize
\caption{{\PBrev 95\% confidence intervals (lower--upper, percentage points) for the detection metrics and standard deviations for DSC/NSD in Tab.~\ref{tab:all_results}. The 95\% confidence intervals were estimated by nonparametric bootstrap over the test cases (1{,}000 resamples; 2.5th and 97.5th percentiles). Statistical tests were performed to compare F1-Score and AUC (paired permutation test for F1-Score and DeLong's test for AUC). Orange highlights statistically significant gains over standard segmentation ($p<0.05$).}}
\label{tab:cif_all_results}
\setlength{\tabcolsep}{2.65pt}
\begin{tabular}{l*{19}{c}}
\toprule
 & \multicolumn{12}{c}{\footnotesize pancreas tumor} & \multicolumn{6}{c}{\footnotesize kidney tumor} \\
\cmidrule(lr){2-13}\cmidrule(lr){14-19}
 & \multicolumn{8}{c}{\scriptsize JHH-Test} & \multicolumn{4}{c}{\scriptsize UCSF-Test} & \multicolumn{6}{c}{\scriptsize UCSF-Test} \\
\cmidrule(lr){2-9}\cmidrule(lr){10-13}\cmidrule(lr){14-19}
\scriptsize train paradigm & mask & rep. & DSC & NSD & F1 & AUC & Se & Sp & F1 & AUC & Se & Sp & mask & rep. & F1 & AUC & Se & Sp \\
\midrule
\multicolumn{19}{l}{\textit{few training masks (50)}} \\
\href{https://arxiv.org/abs/2203.00131}{standard segmentation} & 50 & 0 & 47 & 46 & 50--73 & 51--73 & 48--75 & 50--76 & 43--58 & 57--70 & 39--55 & 72--82 & 50 & 0 & 58--69 & 63--73 & 61--75 & 56--69 \\
\textbf{R-Super (ours)} & 50 & 2.2K & 47 & 47 & 55--77 & 65--83 & 56--80 & 51--78 & 55--67 & 65--75 & 66--80 & 52--65 & 50 & 2.7K & 60--70 & 60--71 & 66--80 & 48--62 \\
\midrule
\multicolumn{19}{l}{\textit{medium / many training masks (344 / 1.7K)}} \\
\href{https://www.nature.com/articles/s41586-026-10181-8}{CLIP-Like} & 344 & 2.2K & 46 & 46 & 75--90 & 77--92 & 82--98 & 58--84 & 53--66 & 68--79 & 52--67 & 70--81 & 1.7K & 2.7K & 59--70 & 66--76 & 68--81 & 40--55 \\
\href{https://ieeexplore.ieee.org/document/8759483/}{Multi-task learning} & 344 & 2.2K & 44 & 44 & 62--81 & 70--88 & 66--89 & 46--74 & 46--59 & 56--68 & 57--73 & 43--56 & 1.7K & 2.7K & 58--70 & 65--76 & 61--76 & 56--70 \\
\href{https://pubs.rsna.org/doi/full/10.1148/ryai.230031}{Report-G Pseudo-labels} & 344 & 2.2K & 44 & 43 & 72--88 & 68--87 & 69--90 & 65--88 & 62--75 & 78--87 & 57--72 & 81--90 & 1.7K & 2.7K & 66--75 & 69--79 & 74--86 & 53--66 \\
\href{https://doi.org/10.1016/j.media.2020.101840}{Models Genesis} & 344 & 0 & 46 & 45 & 68--87 & 71--89 & 65--90 & 62--86 & 52--65 & 67--78 & 58--74 & 57--69 & 1.7K & 0 & 61--72 & 68--78 & 71--83 & 47--60 \\
\href{https://www.nature.com/articles/s41592-020-01008-z}{nnU-Net} & 344 & 0 & 47 & 45 & 64--83 & 68--86 & 63--86 & 61--85 & 62--74 & 74--84 & 67--81 & 68--80 & 1.7K & 0 & 67--77 & 70--80 & 82--92 & 45--60 \\
\href{https://arxiv.org/abs/2203.00131}{standard segmentation} & 344 & 0 & 45 & 42 & 74--90 & 75--92 & 66--89 & 79--96 & 60--73 & 72--83 & 54--69 & 81--90 & 1.7K & 0 & 66--75 & 68--78 & 58--72 & 61--74 \\
\textbf{R-Super (ours)} & 344 & 2.2K & 41 & 38 & \cellcolor{lightorange}85--96 & \cellcolor{lightorange}85--97 & 87--100 & 78--96 & \cellcolor{lightorange}78--87 & \cellcolor{lightorange}87--94 & 76--89 & 85--93 & 1.7K & 2.7K & \cellcolor{lightorange}69--79 & \cellcolor{lightorange}73--82 & 74--85 & 64--77 \\
\bottomrule
\end{tabular}
\end{table*}

\begin{table*}[t]
\centering
\scriptsize
\caption{{\PBrev 95\% confidence intervals (lower--upper, percentage points) for the detection metrics and standard deviations for DSC/NSD in Tab.~\ref{tab:results_by_size} (size-stratified). The 95\% confidence intervals were estimated by nonparametric bootstrap over the test cases (1{,}000 resamples; 2.5th and 97.5th percentiles). Statistical tests were performed to compare F1-Score and AUC (paired permutation test for F1-Score and DeLong's test for AUC). Orange highlights statistically significant gains over standard segmentation ($p<0.05$).}}
\label{tab:cif_results_by_size}
\setlength{\tabcolsep}{2.3pt}
\begin{tabular}{l*{19}{c}}
\toprule
 & \multicolumn{12}{c}{\footnotesize pancreas tumor} & \multicolumn{6}{c}{\footnotesize kidney tumor} \\
\cmidrule(lr){2-13}\cmidrule(lr){14-19}
 & \multicolumn{8}{c}{\scriptsize JHH-Test} & \multicolumn{4}{c}{\scriptsize UCSF-Test} & \multicolumn{6}{c}{\scriptsize UCSF-Test} \\
\cmidrule(lr){2-9}\cmidrule(lr){10-13}\cmidrule(lr){14-19}
\scriptsize train paradigm & mask & rep. & DSC & NSD & F1 & AUC & Se & Sp & F1 & AUC & Se & Sp & mask & rep. & F1 & AUC & Se & Sp \\
\midrule
\multicolumn{19}{c}{\textbf{small tumors (diameter $\leq$ 2 cm)}} \\
\midrule
\addlinespace[2pt]
\multicolumn{19}{l}{\textit{few training masks (50)}} \\
\href{https://arxiv.org/abs/2203.00131}{standard segmentation} & 50 & 0 & 13 & 29 & 18--56 & 43--78 & 27--78 & 48--76 & 32--50 & 51--66 & 31--51 & 72--83 & 50 & 0 & 34--51 & 52--67 & 42--64 & 56--70 \\
\textbf{R-Super (ours)} & 50 & 2.2K & 24 & 32 & 15--52 & 51--83 & 21--73 & 50--77 & 46--61 & 65--78 & 66--84 & 52--65 & 50 & 2.7K & 38--54 & 53--69 & 54--75 & 48--62 \\
\midrule
\multicolumn{19}{l}{\textit{medium / many training masks (344 / 1.7K)}} \\
\href{https://www.nature.com/articles/s41586-026-10181-8}{CLIP-Like} & 344 & 2.2K & 19 & 32 & 52--82 & 82--97 & 100--100 & 58--83 & 41--58 & 67--79 & 45--64 & 69--81 & 1.7K & 2.7K & 32--48 & 58--73 & 46--68 & 41--54 \\
\href{https://ieeexplore.ieee.org/document/8759483/}{Multi-task learning} & 344 & 2.2K & 20 & 31 & 35--70 & 69--94 & 65--100 & 46--74 & 34--50 & 54--68 & 49--71 & 43--56 & 1.7K & 2.7K & 37--54 & 57--73 & 46--68 & 56--70 \\
\href{https://pubs.rsna.org/doi/full/10.1148/ryai.230031}{Report-G Pseudo-labels} & 344 & 2.2K & 21 & 36 & 40--77 & 62--91 & 50--94 & 68--90 & 54--71 & 76--87 & 53--72 & 81--90 & 1.7K & 2.7K & 41--57 & 64--77 & 56--76 & 53--66 \\
\href{https://doi.org/10.1016/j.media.2020.101840}{Models Genesis} & 344 & 0 & 15 & 27 & 40--78 & 74--94 & 58--100 & 63--87 & 40--55 & 63--76 & 50--70 & 57--70 & 1.7K & 0 & 34--52 & 58--72 & 48--71 & 47--61 \\
\href{https://www.nature.com/articles/s41592-020-01008-z}{nnU-Net} & 344 & 0 & 26 & 30 & 36--74 & 61--91 & 50--94 & 62--86 & 53--68 & 72--84 & 62--80 & 68--79 & 1.7K & 0 & 45--61 & 64--78 & 72--90 & 46--59 \\
\href{https://arxiv.org/abs/2203.00131}{standard segmentation} & 344 & 0 & 20 & 31 & 42--81 & 69--95 & 40--92 & 78--96 & 50--67 & 70--83 & 48--67 & 80--89 & 1.7K & 0 & 29--48 & 62--76 & 33--56 & 60--74 \\
\textbf{R-Super (ours)} & 344 & 2.2K & 26 & 31 & 62--93 & 80--96 & 79--100 & 78--96 & \cellcolor{lightorange}68--82 & \cellcolor{lightorange}86--94 & 68--85 & 85--93 & 1.7K & 2.7K & 48--64 & \cellcolor{lightorange}72--84 & 59--79 & 62--76 \\
\midrule[1pt]
\multicolumn{19}{c}{\textbf{large tumors (diameter > 2 cm)}} \\
\midrule
\addlinespace[2pt]
\multicolumn{19}{l}{\textit{few training masks (50)}} \\
\href{https://arxiv.org/abs/2203.00131}{standard segmentation} & 50 & 0 & 6 & 10 & 45--71 & 48--73 & 49--80 & 49--75 & 23--48 & 67--87 & 42--78 & 72--83 & 50 & 0 & 47--65 & 68--82 & 71--91 & 56--69 \\
\textbf{R-Super (ours)} & 50 & 2.2K & 15 & 24 & 52--78 & 68--87 & 62--90 & 51--77 & 21--41 & 65--85 & 62--93 & 52--65 & 50 & 2.7K & 45--61 & 65--78 & 77--93 & 48--62 \\
\midrule
\multicolumn{19}{l}{\textit{medium / many training masks (344 / 1.7K)}} \\
\href{https://www.nature.com/articles/s41586-026-10181-8}{CLIP-Like} & 344 & 2.2K & 19 & 28 & 64--85 & 73--91 & 72--97 & 58--84 & 30--54 & 73--93 & 67--96 & 69--80 & 1.7K & 2.7K & 46--62 & 71--83 & 89--100 & 41--54 \\
\href{https://ieeexplore.ieee.org/document/8759483/}{Multi-task learning} & 344 & 2.2K & 22 & 32 & 49--75 & 66--88 & 58--88 & 47--73 & 19--37 & 59--80 & 68--96 & 44--56 & 1.7K & 2.7K & 46--62 & 70--83 & 68--87 & 56--70 \\
\href{https://pubs.rsna.org/doi/full/10.1148/ryai.230031}{Report-G Pseudo-labels} & 344 & 2.2K & 22 & 34 & 67--87 & 67--89 & 70--94 & 67--90 & 42--67 & 81--97 & 68--95 & 81--90 & 1.7K & 2.7K & 54--69 & 69--81 & 92--100 & 53--67 \\
\href{https://doi.org/10.1016/j.media.2020.101840}{Models Genesis} & 344 & 0 & 18 & 26 & 58--82 & 68--87 & 62--90 & 63--87 & 26--46 & 72--92 & 71--97 & 57--70 & 1.7K & 0 & 50--65 & \cellcolor{lightorange}74--86 & 92--100 & 47--61 \\
\href{https://www.nature.com/articles/s41592-020-01008-z}{nnU-Net} & 344 & 0 & 26 & 30 & 58--81 & 68--88 & 61--90 & 62--85 & 33--56 & 84--94 & 76--100 & 68--79 & 1.7K & 0 & 48--63 & 70--82 & 87--99 & 46--59 \\
\href{https://arxiv.org/abs/2203.00131}{standard segmentation} & 344 & 0 & 22 & 31 & 71--91 & 73--93 & 68--94 & 78--96 & 42--67 & 77--96 & 71--97 & 80--90 & 1.7K & 0 & 55--71 & 68--81 & 84--97 & 61--74 \\
\textbf{R-Super (ours)} & 344 & 2.2K & 27 & 34 & 79--96 & \cellcolor{lightorange}86--98 & 86--100 & 78--96 & \cellcolor{lightorange}56--79 & \cellcolor{lightorange}89--97 & 89--100 & 85--93 & 1.7K & 2.7K & \cellcolor{lightorange}59--74 & 69--81 & 92--100 & 62--75 \\
\bottomrule
\end{tabular}
\end{table*}

\begin{table*}[!h]
\centering
\scriptsize
\caption{{\PBrev 95\% confidence intervals (lower--upper, percentage points) for the detection metrics and standard deviations for DSC/NSD in Tab.~\ref{tab:pants}. The 95\% confidence intervals were estimated by nonparametric bootstrap over the test cases (1{,}000 resamples; 2.5th and 97.5th percentiles). Statistical tests were performed to compare R-Super and standard segmentation for F1-Score and AUC (paired permutation test for F1-Score and DeLong's test for AUC); orange highlights statistically significant gains ($p<0.05$).}}
\label{tab:cif_pants}
\setlength{\tabcolsep}{2.1pt}
\begin{tabular}{l*{18}{c}}
\toprule
  & \multicolumn{2}{c}{\footnotesize training} & \multicolumn{6}{c}{\footnotesize JHH-Small} & \multicolumn{6}{c}{\footnotesize JHH-Large} & \multicolumn{4}{c}{\footnotesize Merlin-Test} \\
\cmidrule(lr){2-3}\cmidrule(lr){4-9}\cmidrule(lr){10-15}\cmidrule(lr){16-19}
\scriptsize train paradigm & mask & rep. & DSC & NSD & F1 & AUC & Se & Sp & DSC & NSD & F1 & AUC & Se & Sp & F1 & AUC & Se & Sp \\
\midrule
\href{https://www.nature.com/articles/s41592-020-01008-z}{nnU-Net} & 926 & 0 & 26 & 30 & 79--88 & 87--94 & 74--87 & 82--93 & 26 & 30 & 76--92 & 90--99 & 78--97 & 90--98 & 64--75 & 71--80 & 61--75 & 67--79 \\
\href{https://ieeexplore.ieee.org/document/8759483/}{Multi-task learning} & 926 & 1{,}848 & 25 & 32 & 87--95 & 91--98 & 85--97 & 86--97 & 24 & 24 & 93--100 & 96--100 & 93--100 & 90--100 & 64--75 & 71--79 & 56--70 & 77--88 \\
\href{https://arxiv.org/abs/2203.00131}{standard segmentation} & 926 & 0 & 24 & 31 & 86--95 & 90--97 & 78--93 & 94--100 & 22 & 21 & 96--100 & 99--100 & 93--100 & 100--100 & 61--73 & 69--78 & 52--67 & 77--87 \\
\textbf{R-Super (ours)} & 926 & 1{,}848 & 24 & 29 & \cellcolor{lightorange}92--98 & 94--99 & 87--98 & 94--100 & 23 & 22 & 95--100 & 100--100 & 100--100 & 90--100 & \cellcolor{lightorange}73--82 & \cellcolor{lightorange}79--87 & 70--83 & 74--85 \\
\bottomrule
\end{tabular}
\end{table*}

\begin{table}[!h]
\centering
\scriptsize
\caption{{\PBrev 95\% confidence intervals (lower--upper, percentage points) for the detection metrics in Tab.~\ref{tab:jhh_ucsf}. The 95\% confidence intervals were estimated by nonparametric bootstrap over the test cases (1{,}000 resamples; 2.5th and 97.5th percentiles). Statistical tests compare R-Super and standard segmentation (paired permutation test for F1-Score, DeLong's test for AUC); orange highlights statistically significant gains ($p<0.05$).}}
\label{tab:cif_jhh_ucsf}
\setlength{\tabcolsep}{3.7pt}
\begin{tabular}{l*{6}{c}}
\toprule
  & \multicolumn{2}{c}{\footnotesize training} & \multicolumn{4}{c}{\footnotesize pancreas tumor} \\
\cmidrule(lr){2-3}\cmidrule(lr){4-7}
\scriptsize train paradigm & mask & rep. & F1 & AUC & Se & Sp \\
\midrule
\multicolumn{7}{c}{\footnotesize\textit{{\PBrev Swiss-Test}}} \\
\midrule
\href{https://www.nature.com/articles/s41592-020-01008-z}{nnU-Net} & 3,488 & 0 & 74--81 & 70--78 & 70--78 & 65--77 \\
\href{https://ieeexplore.ieee.org/document/8759483/}{Multi-task learning} & 3,488 & 28{,}295 & 70--77 & 72--79 & 59--68 & 80--89 \\
\href{https://arxiv.org/abs/2203.00131}{standard segmentation} & 3,488 & 0 & 76--84 & 79--85 & 67--77 & 80--91 \\
\textbf{R-Super (ours)} & 3,488 & 28{,}295 & \cellcolor{lightorange}83--88 & \cellcolor{lightorange}82--88 & 78--85 & 81--90 \\
\midrule
\multicolumn{7}{c}{\footnotesize\textit{{\PBrev UCSF-Test-Non-Contrast}}} \\
\midrule
\href{https://www.nature.com/articles/s41592-020-01008-z}{nnU-Net} & {\PBrev 3,488} & {\PBrev 0} & {\PBrev 40--62} & {\PBrev 58--76} & {\PBrev 52--80} & {\PBrev 66--80} \\
\href{https://ieeexplore.ieee.org/document/8759483/}{Multi-task learning} & {\PBrev 3,488} & {\PBrev 28{,}295} & {\PBrev 5--33} & {\PBrev 50--60} & {\PBrev 2--20} & {\PBrev 98--100} \\
\href{https://arxiv.org/abs/2203.00131}{standard segmentation} & {\PBrev 3,488} & {\PBrev 0} & {\PBrev 47--74} & {\PBrev 67--82} & {\PBrev 36--65} & {\PBrev 93--99} \\
\textbf{R-Super (ours)} & {\PBrev 3,488} & {\PBrev 28{,}295} & \cellcolor{lightorange}{\PBrev 63--84} & \cellcolor{lightorange}{\PBrev 88--96} & {\PBrev 55--80} & {\PBrev 93--99} \\
\bottomrule
\end{tabular}
\end{table}

\begin{table*}[t]
\centering
\scriptsize
\caption{{\PBrev Uncertainty for Tab.~\ref{tab:ablations}: 95\% confidence intervals (lower--upper, percentage points) for Se/Sp/F1/AUC and standard deviations for DSC/NSD. The 95\% confidence intervals were estimated by nonparametric bootstrap over the test cases (1{,}000 resamples; 2.5th and 97.5th percentiles).}}
\label{tab:cif_ablations}
\setlength{\tabcolsep}{2.35pt}
\begin{tabular}{l*{19}{c}}
\toprule
 & \multicolumn{12}{c}{\footnotesize pancreas tumor} & \multicolumn{6}{c}{\footnotesize kidney tumor} \\
\cmidrule(lr){2-13}\cmidrule(lr){14-19}
 & \multicolumn{8}{c}{\scriptsize JHH-Test} & \multicolumn{4}{c}{\scriptsize UCSF-Test} & \multicolumn{6}{c}{\scriptsize UCSF-Test} \\
\cmidrule(lr){2-9}\cmidrule(lr){10-13}\cmidrule(lr){14-19}
\scriptsize train paradigm & mask & rep. & DSC & NSD & F1 & AUC & Se & Sp & F1 & AUC & Se & Sp & mask & rep. & F1 & AUC & Se & Sp \\
\midrule
\textbf{R-Super (ours)} & 344 & 2.2K & 41 & 38 & 85--96 & 85--97 & 87--100 & 78--96 & 78--87 & 87--94 & 76--89 & 85--93 & 1.7K & 2.7K & 69--79 & 73--82 & 74--85 & 64--77 \\
\midrule
\multicolumn{19}{l}{\textit{loss functions}} \\
Volume~Loss only & 344 & 2.2K & 41 & 40 & 88--98 & 94--100 & 81--98 & 94--100 & 70--81 & 85--93 & 73--86 & 77--86 & 1.7K & 2.7K & 70--80 & 69--79 & 81--91 & 56--70 \\
Ball Loss only & 344 & 2.2K & 43 & 42 & 74--91 & 86--96 & 66--90 & 81--98 & 64--77 & 77--87 & 56--72 & 86--94 & 1.7K & 2.7K & 69--78 & 69--80 & 82--92 & 51--65 \\
\midrule
\multicolumn{19}{l}{\textit{LLM size}} \\
small LLM (4B) & 344 & 2.2K & 40 & 34 & 73--90 & 75--92 & 63--86 & 84--98 & 64--76 & 74--85 & 66--81 & 74--84 & 1.7K & 2.7K & 66--76 & 71--80 & 85--94 & 41--55 \\
tiny LLM (0.6B) & 344 & 2.2K & 44 & 42 & 63--83 & 72--88 & 57--83 & 68--90 & 55--70 & 68--78 & 45--63 & 86--94 & 1.7K & 2.7K & 68--77 & 71--81 & 81--91 & 48--62 \\
\midrule
\multicolumn{19}{l}{\textit{report information removed}} \\
no pancreas sub-segment & 344 & 2.2K & 43 & 37 & 76--92 & 83--96 & 68--90 & 84--98 & 65--77 & 76--85 & 65--81 & 76--86 & 1.7K & 2.7K & 67--77 & 70--80 & 81--92 & 47--60 \\
no tumor size & 344 & 2.2K & 45 & 42 & 76--93 & 81--94 & 63--88 & 93--100 & 60--73 & 72--81 & 56--72 & 79--88 & 1.7K & 2.7K & 68--78 & 71--81 & 82--93 & 48--62 \\
\midrule
\multicolumn{19}{l}{\textit{organ mask importance}} \\
50 CT organ segmenter & 344 & 2.2K & 39 & 31 & 89--99 & 93--100 & 89--100 & 86--100 & 70--81 & 81--89 & 80--91 & 71--82 & 1.7K & 2.7K & 67--77 & 70--80 & 78--89 & 50--64 \\
no organ mask & 344 & 2.2K & 42 & 37 & 80--94 & 87--97 & 67--91 & 93--100 & 63--75 & 76--85 & 60--76 & 78--88 & 1.7K & 2.7K & 67--78 & 71--81 & 79--90 & 51--65 \\
\midrule
\multicolumn{19}{l}{\textit{injected LLM errors (or report errors)}} \\
+5\% errors & 344 & 2.2K & 39 & 32 & 86--97 & 88--98 & 81--98 & 90--100 & 66--78 & 80--89 & 65--80 & 78--88 & 1.7K & 2.7K & 69--78 & 72--81 & 86--94 & 44--58 \\
+10\% errors & 344 & 2.2K & 41 & 34 & 82--95 & 90--98 & 78--96 & 83--98 & 72--83 & 83--91 & 71--85 & 83--91 & 1.7K & 2.7K & 68--78 & 72--81 & 77--88 & 56--70 \\
+20\% errors & 344 & 2.2K & 41 & 34 & 82--95 & 83--96 & 76--94 & 87--100 & 65--77 & 77--86 & 68--83 & 73--84 & 1.7K & 2.7K & 67--77 & 71--81 & 81--92 & 46--60 \\
100\% errors & 344 & 2.2K & 45 & 42 & 57--80 & 64--84 & 45--73 & 80--98 & 50--62 & 62--73 & 65--80 & 44--57 & 1.7K & 2.7K & 60--71 & 64--75 & 70--83 & 43--57 \\
\bottomrule
\end{tabular}
\end{table*}

\section{LLM Prompt}
\label{app:prompts}

We used the prompt below to extract tumor information from radiology reports. The prompt was iteratively created in a collaboration between computer scientists and radiologists (line breaks as in the original prompt; \%(organ)s, \%(example\_report)s, and \%(example\_answer)s are placeholders filled per organ with an in-context example):

\begin{quote}\small\itshape\sloppy\emergencystretch=3em\setlength{\parskip}{1pt}\obeylines
Instructions: The radiology report below possibly mentions one or more focal lesions (e.g., tumor, mass, nodule, cyst). 
  Read it carefully, paying special attention to the findings, clinical history, and impressions sections (if available). 
  Your task is to list the types, certainty of lesion type, sizes, organ, locations and attenuation of all lesions in the report. 
  Fill out the template below, using one line per lesion (you may add or remove lines from the template): 
  lesion 1: type = \_; certainty = \_; size = \_; organ = \_; location = \_; attenuation = \_; 
  lesion 2: type = \_; certainty = \_; size = \_; organ = \_; location = \_; attenuation = \_;
  ... 
  If you are absolutely sure the report mentions no lesion, do not use the template. Instead, reply with: `No lesions mentioned.' and justify why you are sure the report mentions no lesion. 
  Consider the following instructions: 
   A - What is a Lesion:
  A lesion is any focal abnormality, including masses, cysts, or areas of altered density (hyperdense, hypodense, or isodense).
  You must list both benign and malignant lesions. Include lesions that are confirmed as well as those that are only suspicious (in this case, use `certainty = low' in the filled template).
  Common terms that indicate a lesion: metastasis, nodule, soft tissue, nodular thickening, tumor, lesion, mass, cyst, pseudo-cyst, neoplasm, cancer, index lesion, oncologic finding, adenoma, carcinoma, growth, abnormal thickening, focus, LI-RADS lesion, hyperdensity, hypodensity, and isodensity.
  Terms that are not a focal lesion: diverticulum (unless it is suspicious for malignancy); renal/gallbladder stones (e.g., cholelithiasis); hyper/hypoenhancing fluid collection not caused by a cyst or mass (e.g., caused by abscess, or postoperative changes); if the patient has cancer history but the tumor was surgically removed (e.g., whipple procedure for pancreatic tumors) and there is no current evidence of the tumor, do not list it;
   B - Size:
  1- When to use numbers:
  Reports can write the size of the lesion in 1D, 2D, or 3D measurements, and you should use the same standards used in the report. 
  Write 1D measurements as: 15 mm; 2D measurements as: 15 x 10 mm; and 3D measurements as: 40 x 30 x 30 mm. You may use either cm or mm, but you MUST WRITE in each line of the filled template the unit you are using (cm or mm). If a report does not specify the unit, assume it is mm. 
  Say size = U if the report does not specify the size of a lesion. 
  2- When to use `size = tiny' or `size = massive':
  If a reports describes a lesion without numeric diameters, but describes the lesion with adjectives like tiny, small, large, or massive, you **must** include one entry with `size = tiny' or `size = massive' for that lesion. However, always prefer using the lesion diameter, if provided. 
  If the report does not provide the diameter nor any size-related adjective, say size = U. 
  3- When to use `size = multiple':
  If the report mentions multiple lesions in an organ **but does not give an exact count**, you **must** include one entry with `size = multiple' for that organ.
  Only use `size = multiple' when you cannot determine the number; otherwise, create individual entries for each described lesion.
  Handling reports with unknown lesion counts AND some described lesions:
  If the report says there are multiple/innumerable/many lesions in an organ **and** then describes a few of them (e.g., gives their size), your output should include:
  One row describing each **explicitly described** lesion.
  One additional row with `size = multiple' to represent the unspecified lesions.
  Example:
  Report: A 2 cm metastasis in liver sub-segment 3, and multiple other small metastases in the liver.
  Your output:
  lesion 1: type = metastasis; certainty = high; size = 2 cm; organ = liver; location = segment 3; attenuation = U;
  lesion 2: type = metastasis; certainty = high; size = multiple; organ = liver; location = U; attenuation = U;
  Key rules:
  1. Always include individual entries for every lesion that is specifically described in the report.
  2. Whenever the report indicates multiple unspecified lesions in an organ, also include one entry with `size = multiple' for the organ.
   C - Organ:
  Organ is the organ where the lesion is located. Use standard organ names, like: liver, pancreas, kidney, spleen, colon, pelvis, adrenal gland, bladder, gallbladder, breast, stomach, lung, esophagus, uterus, bone, prostate, and duodenum. 
  Do not consider the `GI-Tract' as an organ. Instead, try to localize the tumor in one of these GI-tract organs: esophagus, stomach, duodenum, small intestines or colon. You can consider `organ = colon' for rectal lesions. You can consider `organ = esophagus' for lesions in the esophagus-gastric junction. 
   D - Location:
  For location, check if the report specifies the sub-segment or part of the organ where the lesion is. If it specifies: for liver, choose location as segment 1/2/.../8. For the pancreas choose the pancreas head/body/neck/tail/uncinate process. For the kidney: left kidney/right kidney. For other organs, just check if the report mentions some type of organ region or sub-segment. 
  Say location = U if the report does not specify the intra-organ location of a lesion. 
  If a single lesion is in more than one location, you can say both. E.g., location = liver segment 4/5. 
   E - Type:
  If the report provides lesion type, inform it. 
  Otherwise, say type = U if the report does not specify the type of the lesion. 
  Follow these rules:
  1- If the lesion type is not specified in findings, you may deduce it from the clinical history or impressions sections. 
  2- Cyst, or cystic lesion, is a common type of lesion. For any cysts, you say type = cyst. 
  3- Assign type = metastasis if the lesion is described as a metastasis (or implant) originating from a cancer in another organ, unless the report explicitly states otherwise. Determine certainty based on the level of confidence expressed in the report. 
  4- If the report does not mention the lesion type, but you read that the patient has cancer in the organ where the lesion is, or has a history of malignant lesions in the organ, or the lesion is growing, or it is `suspicious for malignancy', say type = malignant. However, do not say type = malignant if a more specific lesion type is given (like PDAC and PNET in pancreas, RCC in kidney, HCC in liver,...). 
  4- If the report does not mention the lesion type, but it explicitly indicates that the lesion is likely benign (or mentions it may be one of several benign lesion types), say type = benign. However, do not say type = benign if a more specific lesion type is given (like cyst or polyp). 
  5- Try reporting types using their most standard name, followed by their acronym. For example, if the report mentions `adenocarcinoma in the pancreas' or `PDAC', say type = Pancreatic Ductal Adenocarcinoma (PDAC). 
   F - Certainty:
  Certainty of the lesion type, according to the report. If a report mentions a lesion type in the findings, history, or impressions, without demonstrating uncertainty, say certainty = certain. 
  If the report expresses strong confidence in lesion type, say certainty = high. 
  If the report mentions a lesion type but expresses significant uncertainty about it, say certainty = low. 
  If the report does not mention the lesion type, say certainty = U. 
   G - Attenuation:
  For each lesion, inform the attenuation if the report mentions it. You should choose one of these options: hyperenhancing, hypoenchanging, isoenhancing, hererogeneously enhancing, or U (unknown). The report may use synonyms like hypermetabolic and hypoattenuating, but you must only answer me hyperenhancing, hypoenchanging, isoenhancing, hererogeneously enhancing or U (unknown). 
   H - Justification:
  Besides filling the template, justify your answer, carefully mentioning each section of the report if present: history, findings, and impressions. 
  Explain from which sentences you got each size, location, and type.
  Some reports may refer to past measurements (using words like previously, before, or giving dates). Ignore previous measurements. 
  Provide me a syntactic analysis of the report sentences mentioning lesion sizes. In this analysis, explain which measurement refers to which lesion, if the measurement is current or past, and if the corresponding lesion is malignant or benign.
  I will provide an example of a report and a correct answer for a \%(organ)s lesion:
  Example report: 
  \%(example\_report)s 
  Example answer: 
  \%(example\_answer)s 
  End of the example.
\end{quote}

\end{document}